%% file: main.tex
\documentclass[runningheads]{ks_press}
\input{preamble}
\begin{document}
\setpagewiselinenumbers



%
\journal{XXXXX}
\volume{XXXXX}
\publisher{XXXXX}
\myDOI{XXXXX}


\title{A Quadratic Programming Approach to Manipulation\\in Real-Time Using Modular
Robots}
\titlerunning{QP Approach to Manipulation}

\author{Chao Liu \and Mark Yim}
\authorrunning{C. Liu et al.}
\institute{University of Pennsylvania at 3401 Grays Ferry Avenue, Philadelphia, US\\
\email{\{chaoliu,yim\}@seas.upenn.edu}
}
\maketitle

\begin{history}
\received{(xx/xx/xx)}
\revised{(xx/xx/xx)}
\accepted{(xx/xx/xx)}
\end{history}

\begin{abstract}
  Motion planning in high-dimensional space is a challenging task. In
  order to perform dexterous manipulation in an unstructured
  environment, a robot with many degrees of freedom is usually
  necessary, which also complicates its motion planning
  problem. Real-time control brings about more difficulties in which
  robots have to maintain the stability while moving towards the
  target. Redundant systems are common in modular robots that consist
  of multiple modules and are able to transform into different
  configurations with respect to different needs. Different from
  robots with fixed geometry or configurations, the kinematics model
  of a modular robotic system can alter as the robot reconfigures
  itself, and developing a generic control and motion planning
  approach for such systems is difficult, especially when multiple
  motion goals are coupled. A new manipulation planning framework is
  developed in this paper. The problem is formulated as a sequential
  linearly constrained quadratic program (QP) that can be solved
  efficiently. Some constraints can be incorporated into this QP,
  including a novel way to approximate environment obstacles. This
  solution can be used directly for real-time applications or as an
  off-line planning tool, and it is validated and demonstrated on the
  CKBot and the SMORES-EP modular robot platforms.
\end{abstract}


\section{Introduction}
\label{S:1}

Manipulation tasks are common in robotics applications. In
unstructured, cluttered environments, these tasks are usually executed
by redundant robots to reach larger workspaces while avoiding
obstacles and other constraints. This results in motion planning in
high-dimensional space.

The motion planning problem is usually solved by some well developed
framework (e.g. MoveIt!~\cite{Coleman-moveit-2014}) containing three
components: a \textit{path planner}, a \textit{trajectory generator},
and a \textit{tracking controller}. The path planner is responsible
for generating collision-free paths. The trajectory generator smooths
the computed paths and generates trajectories that can be
parameterized by time while satisfying motion constraints, such as
maximum velocities and accelerations. The tracking controller
guarantees the motion of the robot when executing the generated
trajectories. This type of framework has shown successful applications
in many scenarios but rarely achieves real-time performance for all
three components in high dimensions. Some approaches combined path
planning with trajectory optimization that can directly construct
trajectories resulting from optimization over a variety of
criteria. These approaches are related to optimal control of robotic
systems.

\begin{figure}[b]
  \centering
  \includegraphics[width=0.6\textwidth]{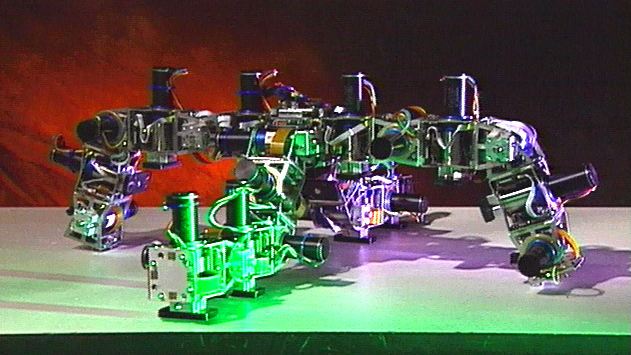}
  \caption{A modular robot configuration built by PolyBot modules is
    composed of multiple chains~\cite{Yim-polybot-icra-2000}.}
  \label{fig:polybot}
\end{figure}

Modular self-reconfigurable robotic systems are usually composed of a
small set of building blocks with uniform docking interfaces that
allow the transfer of mechanical forces and moments, electrical power,
and communication throughout the
robot~\cite{Yim-review-ram-2007}. These platforms are designed to be
versatile and adaptable with respect to different tasks, environments,
functions or activities. A single module in a modular robotic system
usually has one or more degrees of freedom (DoFs). Combining many
modules to form versatile systems results in robots requiring
representations with high dimensions. This dimensionality makes
control and motion planning difficult. That the system is not a single
structure but can take a very large number of configurations
(typically exponential in the number of modules) requires an approach
that can be applied to arbitrary configurations. For example, a
modular robot configuration built with
PolyBot~\cite{Yim-polybot-icra-2000} modules is shown in
Fig.~\ref{fig:polybot} which has multiple serial kinematic
chains. This is different from common multi-limb systems with a single
base. These systems can be modeled such that chains are decoupled.

In a modular robotic system, modules usually are approximated by
simple shapes such as a cube or a sphere. This is often useful for
reconfiguration but can make manipulation more complex. Rather than
two long fingers in a parallel jaw gripper, to obtain similar
geometry, a modular robot may require many modules to form those long
fingers. This results in grasping type of motions in which two or
more chains can behave as a multi-arm system to clamp an
object~\cite{Seo-whole-arm-grasping-ijrr-2016}. Hence, in order to
grasp some object, motion planning for multiple (potentially high DoF)
chains are necessary. In addition their motions are strongly
coupled. This fact leads to a more complicated control and planning
problem.

In this paper, we present a new approach for real-time manipulation
planning and control as well as a novel way to approximate environment
obstacles, and apply it to two modular robotic systems. In order to
solve the problem in general, an universal kinematics model is
required for arbitrary configurations. This approach concomitantly can
be easily extended, such as a dual-arm system or a modular robot
configuration that has multiple chains. We propose a quadratic
programming approach that can be solved efficiently for real-time
applications. This requires that system control stability, hardware
motion constraints (joint limits and actuation limits), and collision
avoidance can be incorporated into this quadratic program (QP) as
linear constraints. One advantage of this approach is that the large
variety of configurations and kinematic structures found from modular
robotic systems can be represented easily as linear constraints,
including situations where multiple portions of the robot may have
different simultaneous goals. Our approach can also be used as an
off-line trajectory planner by simple Euler integration. The framework
is tested and evaluated on CKBot~\cite{Yim-ckbot-2009} and
SMORES-EP~\cite{Liu-smores-reconfig-ral-2019} in the end.

The paper is organized as follows. Sec.~\ref{sec:related} reviews
relevant and previous works. Sec.~\ref{sec:kinematics} introduces the
details to derive the kinematics model required to describe the motion
of any modular robotic configuration. Sec.~\ref{sec:control-planning}
discusses the approach to control and motion planning for given
tasks. Some experiments are validated in Sec.~\ref{sec:experiment}
with some analysis. Sec.~\ref{sec:conclusion} includes the conclusions
and future work.

\section{Related Work}
\label{sec:related}

High-dimensional motion planning and control have been studied over
several decades. In this section, we review several types of
approaches from previous work and some special approaches for modular
robots.

\subsection{Motion Planning for Manipulation}

Artificial potential field manipulation planning methods can avoid
searching in high-dimensional configuration space, planning in
operational space
directly~\cite{Khatib-potential-field-ijrr-1986}. Robots can avoid
collision in real time, but may get stuck at local minima. Analytical
navigation functions that have a unique minimum at the goal
configuration avoiding local minima are shown
in~\cite{Rimon-navigation-function-icra-1988}. However, it is usually
computationally expensive to build such a navigation function in
general. A Monte Carlo technique was applied to escape local minima of
the potential by executing Brownian
motions~\cite{Barraquand-potential-field-ijrr-1991}.

Sampling-based approaches have been used widely for high-dimensional
motion planning problems. The probabilistic roadmap (PRM) has been
demonstrated on planar articulated robots with many
DoFs~\cite{Amato-prm-icra-1996,Kavraki-prm-tro-1996}.  Expansive
configuration space was proposed to resolve the narrow passage issue
which is a common problem for sampling-based
planners~\cite{Hsu-narrow-passage-icra-1997}. Rapidly-exploring Random
Trees (RRT) approach was later presented
in~\cite{Lavalle-rrt-icra-1999} to deal with nonholonomic
constraints. An optimal sampling-based planner (RRT$^{\ast}$) is
introduced in~\cite{Karaman-rrt-star-ijrr-2011} with less
efficiency. These approaches require post-processing to generate
smooth trajectories in order to be executable for real tasks.

Search-based planners rely on the discretization of the space. However
these approaches are generally not suitable for high-dimensional
problems. For example, na\"ive A$^{\ast}$ can rarely scale to large
complicated planning problems. In order to increase the efficiency of
these approaches, a number of suboptimal heuristic searches have been
proposed~\cite{Furcy-heuristic-search-thesis,Likhachev-anytime-a-star-nips-2004,Likhachev-r-star-aaai-2008}. These
methods are promising but are currently computationally inefficient
when solving motion planning problems in high-dimensional space.

Once a feasible path is found, a trajectory generator is needed to
smooth and shorten the computed path with time
parameterization. Trajectories are modeled as elastic bands that need
to maintain equilibrium states under internal contraction forces and
external repulsion
forces~\cite{Quinlan-elastic-band-icra-1993,Brock-elastic-strips-ijrr-2002}. Obstacles
in the workspace are considered directly which is also beneficial for
real-time trajectory modification while a precomputed path is
necessary.

Another class of planners are related to optimal control. Rather than
separate the planning process into two phases (path planning and
trajectory planning), trajectories are constructed directly by these
frameworks which optimize over a variety of criteria. A global
time-optimal trajectory generator is introduced
in~\cite{Shiller-optimal-traj-manipulation-tro-1991}. It combines a
grid search with a local optimization to obtain the global optimal
solution. This approach requires the representations of obstacle
regions in configuration space which is difficult to derive for
high-dimensional
problems. CHOMP~\cite{Ratliff-chomp-icra-2009,Zucker-chomp-ijrr-2013}
formulates the cost to be the combination of trajectory smoothness and
obstacle avoidance, and gradients for these two terms are needed. This
approach uses pre-computed signed distance fields for collision
checking. A similar idea is used in ITOMP that can also deal with
dynamic environments~\cite{Park-itomp-icaps-2012}. In contrast,
STOMP~\cite{Kalakrishnan-stomp-icra-2011} can also handle more general
objective functions for which gradients are not available by using
trajectory samples, but can be difficult to determine the number of
samples. A sequential convex optimization approach is presented
in~\cite{Schulman-arm-planning-ijrr-2014} which adds new constraints
and costs during the motion so as to tackle a larger range of
motion. Collision is detected by checking the intersection of the
swept-out volume of the robot in an interval and obstacles, and a
collision-avoidance penalty gradient can be incorporated into the
optimization problem to ensure safety. These works mainly focus on
single-high-DoF-arm manipulation tasks. Given the trajectories of
end-effectors, an optimal control framework is formulated to solve
whole-body manipulation tasks~\cite{Shankar-qp-planning-wafr-2015}. A
repulsive velocity can be applied to any rigid body whenever it
collides with any obstacle based on a physical simulator. Reachable
sets are used for safe and real-time trajectory design
in~\cite{Holmes-manipulation-traj-rss-2020}, but the reachability
analysis has to be offline. Some optimal controllers handle the
obstacles by mixed integer programming which is known to be an NP-hard
problem~\cite{Schouwenaars-mip-vehicle-planning-ecc-2001}.

Our approach is also related to optimal control and differs from these
previous works in two ways: (a) the way in which the motion planning
problem is formulated and (b) the simple model that approximates the
environment obstacles. We incorporate multiple motion goals into the
objective function in the form of feedback controllers to guarantee
the trajectory tracking performance or efficient search for
navigation, and the output can be applied on the system directly to
achieve real-time performance. During the motion, both the objective
function and constraints may be updated according to the current
scenario which allows our approach to tackle a wider range of
tasks. For collision avoidance, we present a new way to simplify
obstacles dynamically during the motion, and the collision avoidance
constraint can be modeled as linear constraints in order to solve the
optimization problem efficiently. A collision-avoidance penalty is
added to the objective function when any rigid body is near any
obstacle in the form of the projected motion from the rigid body to
this obstacle. Our approach is well suited for real-time applications
since its output can be applied on robotic systems directly in
real-time, or can be used as an off-line trajectory planner by
integrating the output over time.

\subsection{Modular Robot Control and Planning}

Modular robots are inherently systems with many DoFs. They are usually
composed of a large number of modules with each module has one or more
DoFs. This paper addresses the manipulation tasks of modular robots
that form configurations in tree topologies. That is they are
constructed from multiple serial chain configurations without forming
loops. Work related to manipulation of modular robot systems includes
inverse kinematics for highly redundant chains using
PolyBot~\cite{Yim-polybot-icra-2000,Yim-joint-solution-redundant-icra-2001},
and constrained optimization techniques with nonlinear
constraints~\cite{Fromherz-modular-robot-control-2001}. Due to
complicated constraints in these approaches, real-time applications
for large systems cannot be guaranteed and numerical issues have to be
addressed when solving the optimization problem in the presence of
obstacles. Another set of related work includes controller design for
modular robots, such as an adaptive control approach using a neural
network architecture~\cite{Melek-modular-robot-control-2003}, a
virtual decomposition control
approach~\cite{Zhu-modular-robot-control-vdc-tro-2013}, a distributed
control method with torque
sensing~\cite{Liu-distributed-control-2008}, and a centralized
controller~\cite{Giusti-modularrobot-control-iros-2015}. These
approaches consider the control problem in a free environment and
require extra motion planning to fully control the system in a complex
environment.

This paper is built upon the work from our conference
paper~\cite{Liu-control-planning-irc2020}. In the conference paper, we
introduced a quadratic programming approach to address real-time
control and planning, as well as a general solution to build
kinematics models of modular robotic systems. In this work, we present
a sequential optimization approach in which the objective function and
constraints are updated according to the current situation to ensure
safety. In order to handle more complex environments, a novel model is
introduced to approximate obstacles by their significance to the
current robot motion.

\section{Kinematics For Modular Robots}
\label{sec:kinematics}

In this section, we derive a general kinematics model for modular
robots. For other manipulators, a similar technique can be applied to
derive necessary models in order to utilize our planning framework.

\subsection{Kinematics Graph}
\label{sec:configuration}

The representation of a modular robot configuration is discussed
in~\cite{Liu-config-recognition-isrr-2017} which is an undirected
graph $G = (V, E)$. Each vertex $v\in V$ represents a module and each
edge $e\in E$ represents the connection between two modules.

\begin{figure}[t]
  \centering
  \begin{subfloat}[]{\includegraphics[height=0.4\textwidth]{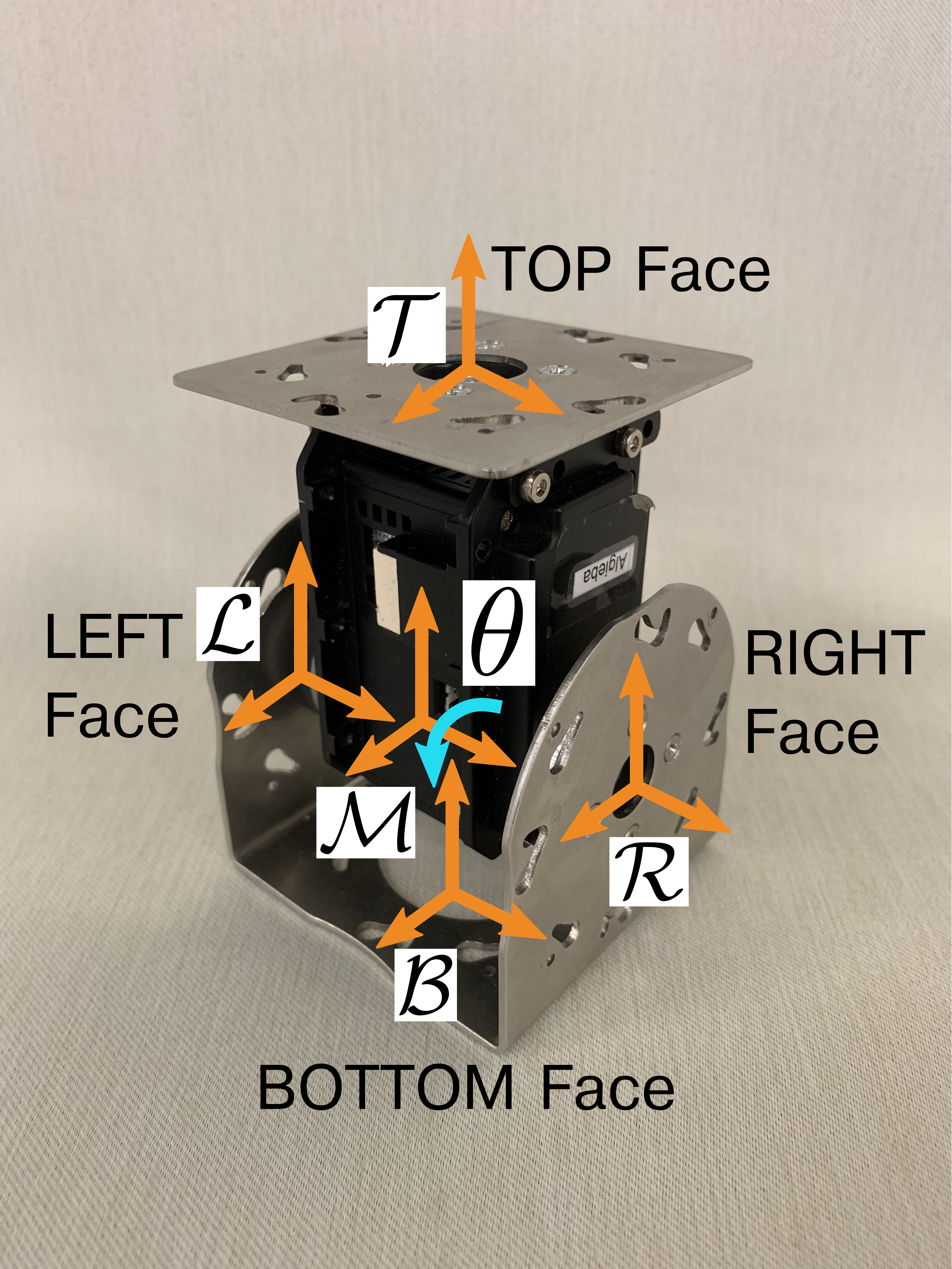}\label{fig:ckbot-ub}}
  \end{subfloat}
  \hfil
  \begin{subfloat}[]{\includegraphics[height=0.4\textwidth]{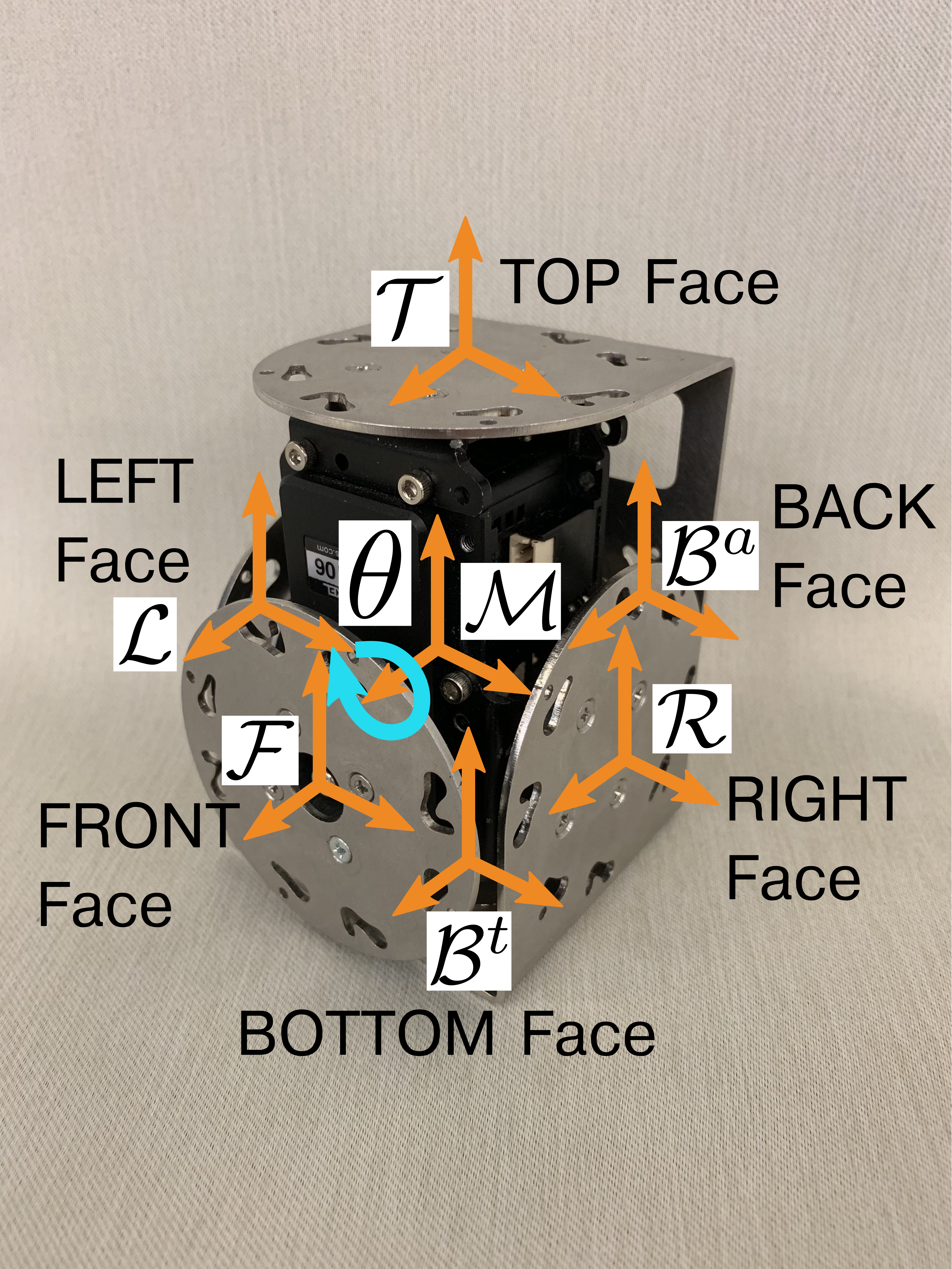}\label{fig:ckbot-cr}}
  \end{subfloat}
  \caption{(a) A CKBot UBar module has one DoF and four
    connectors. (b) A CKBot CR module has one DoF and six connectors.}
  \label{fig:ckbot-module-graph}
\end{figure}

\begin{figure}[t]
  \centering
  \subfloat[]{\includegraphics[width=0.3\textwidth]{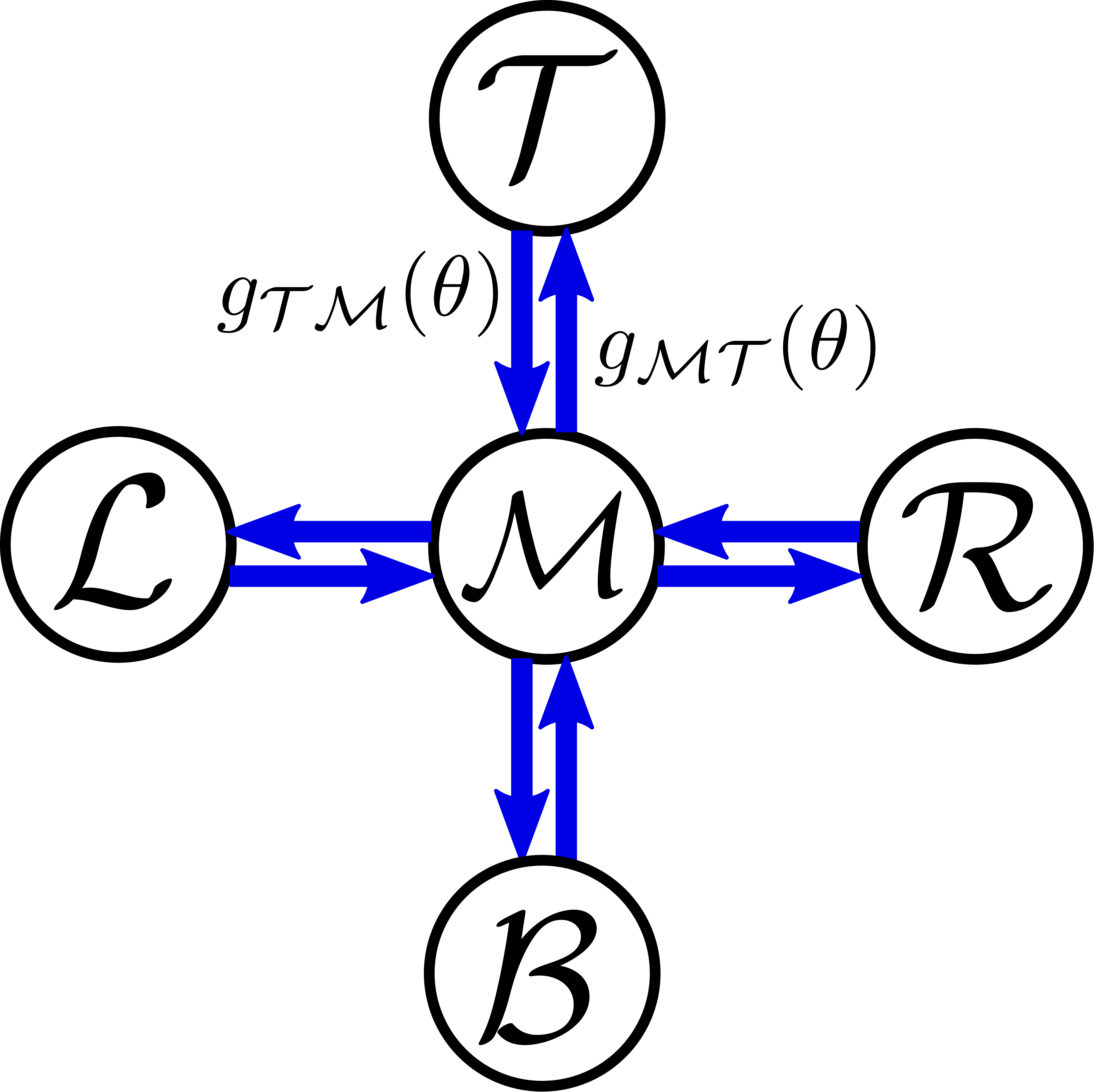}\label{fig:ckbot-ub-graph}}
  \hfil
  \subfloat[]{\includegraphics[width=0.3\textwidth]{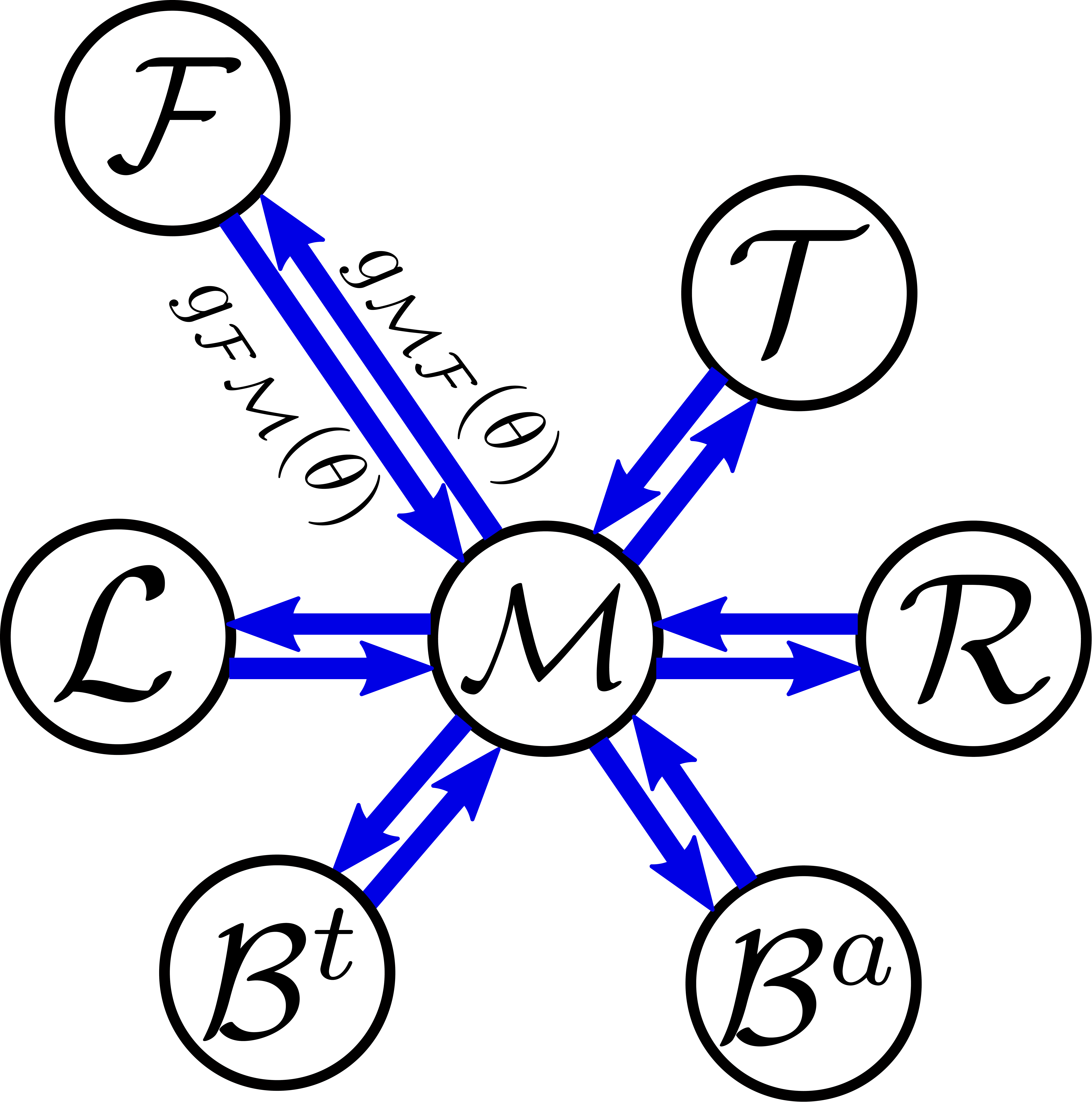}\label{fig:ckbot-cr-graph}}
  \caption{(a) The module graph of a CKBot UBar module in which
    $g_{\mathcal{M}\mathcal{B}}$, $g_{\mathcal{B}\mathcal{M}}$,
    $g_{\mathcal{M}\mathcal{L}}$, $g_{\mathcal{L}\mathcal{M}}$,
    $g_{\mathcal{M}\mathcal{R}}$, and $g_{\mathcal{R}\mathcal{M}}$ are
    invariant of $\theta$. (b) The module graph of a CKBot CR module in
    which $g_{\mathcal{M}\mathcal{B}^a}$,
    $g_{\mathcal{B}^a\mathcal{M}}$, $g_{\mathcal{M}\mathcal{B}^t}$,
    $g_{\mathcal{B}^t\mathcal{M}}$, $g_{\mathcal{M}\mathcal{T}}$,
    $g_{\mathcal{T}\mathcal{M}}$, $g_{\mathcal{M}\mathcal{L}}$,
    $g_{\mathcal{L}\mathcal{M}}$, $g_{\mathcal{M}\mathcal{R}}$, and
    $g_{\mathcal{R}\mathcal{M}}$ are invariant of $\theta$.}
\end{figure}

We use a \textit{module graph} to model a module's topology which
includes all connectors and joints. A \textbf{module graph} is a
directed graph $G_m = (V_m, E_m)$: each vertex is a rigid body in the
module which is either a connector or the module body, and each edge
represents how two adjacent rigid bodies are connected. The
transformations among all rigid bodies are determined by the joint set
and the geometry. For example, a CKBot UBar module in
Fig.~\ref{fig:ckbot-ub} is a single-DoF module as well as four
connectors (\textrm{TOP Face} or $\mathcal{T}$, \textrm{BOTTOM Face}
or $\mathcal{B}$, \textrm{LEFT Face} or $\mathcal{L}$, and
\textrm{RIGHT Face} or $\mathcal{R}$). For simplicity, when the module
joints are in its zero position, all rigid bodies are in the same
orientation and the translation offsets among them are determined by
the module geometry. Let $\mathcal{B}$ be fixed in $\mathcal{M}$, then
the homogeneous transformations among $\mathcal{M}$, $\mathcal{L}$,
and $\mathcal{R}$ are invariant of the joint parameter $\theta$ because
they are rigidly connected. Only the homogeneous transformation
between $\mathcal{M}$ and $\mathcal{T}$ is not invariant to $\theta$.  This
relationship can be fully represented in a directed graph shown in
Fig.~\ref{fig:ckbot-ub-graph}. The edge direction denotes the
direction of the corresponding forward kinematics. $\mathbf{G}_m$ is
the set of unique module graphs $G_m$ for a modular robotic system
since some systems have more than one type of modules (e.g., CKBot in
Fig.~\ref{fig:ckbot-module-graph}).


\begin{figure}[t]
  \centering
  \begin{subfloat}[]{\includegraphics[height=0.23\textwidth]{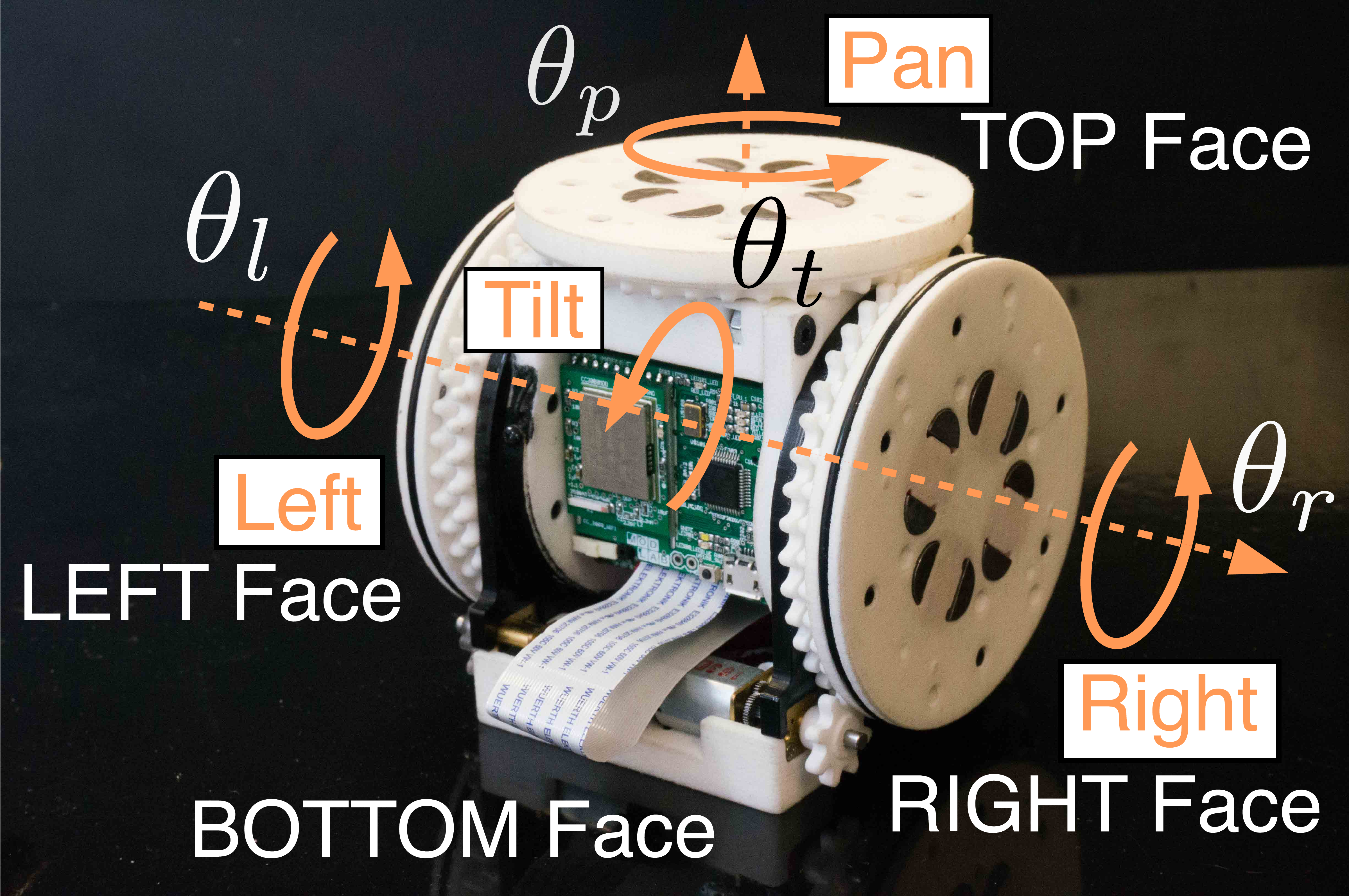}\label{fig:smores}}
  \end{subfloat}
  \begin{subfloat}[]{\includegraphics[height=0.23\textwidth]{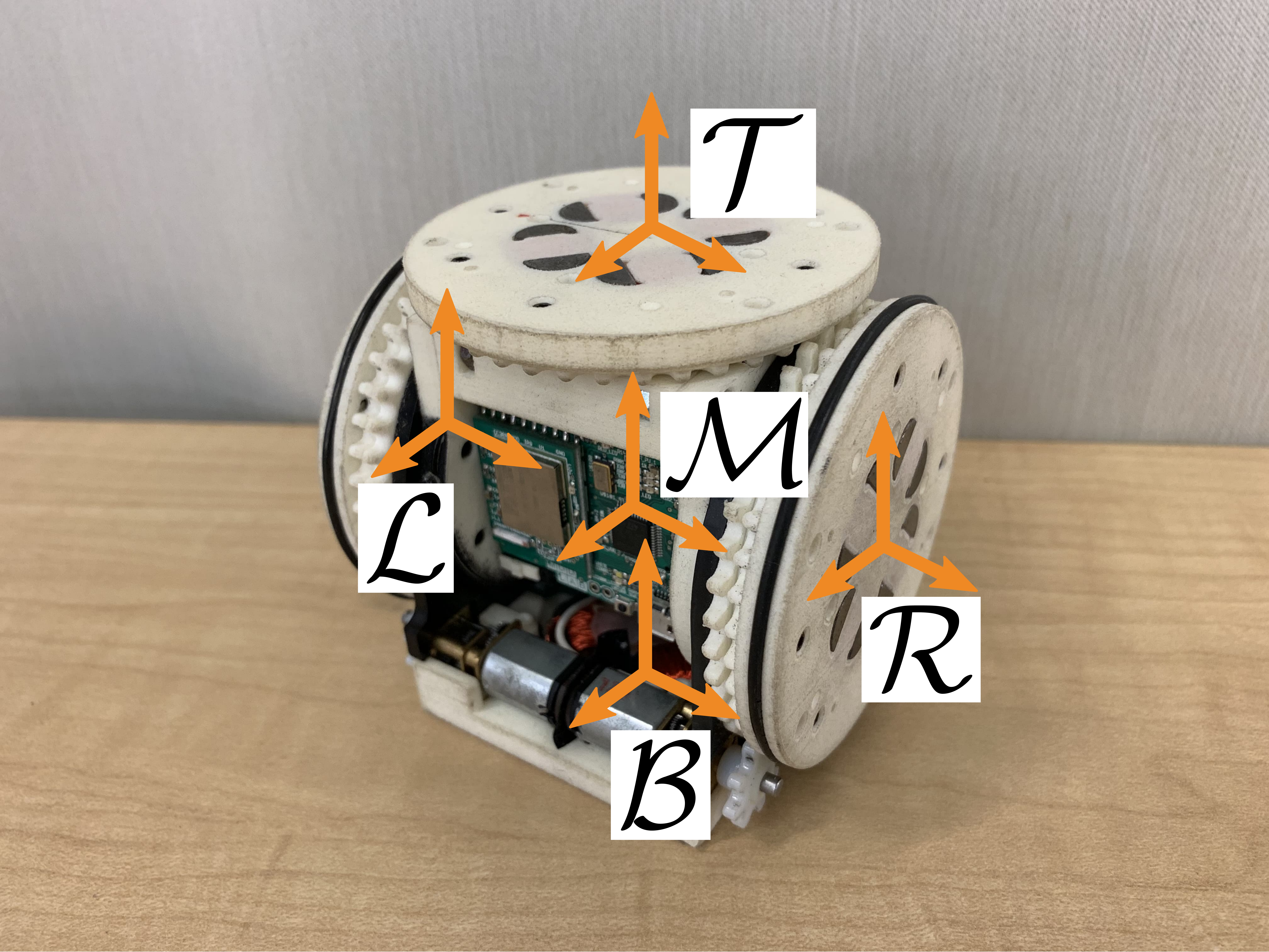}\label{fig:smores-frame}}
  \end{subfloat}
  \begin{subfloat}[]{\raisebox{0ex}{\includegraphics[height=0.23\textwidth]{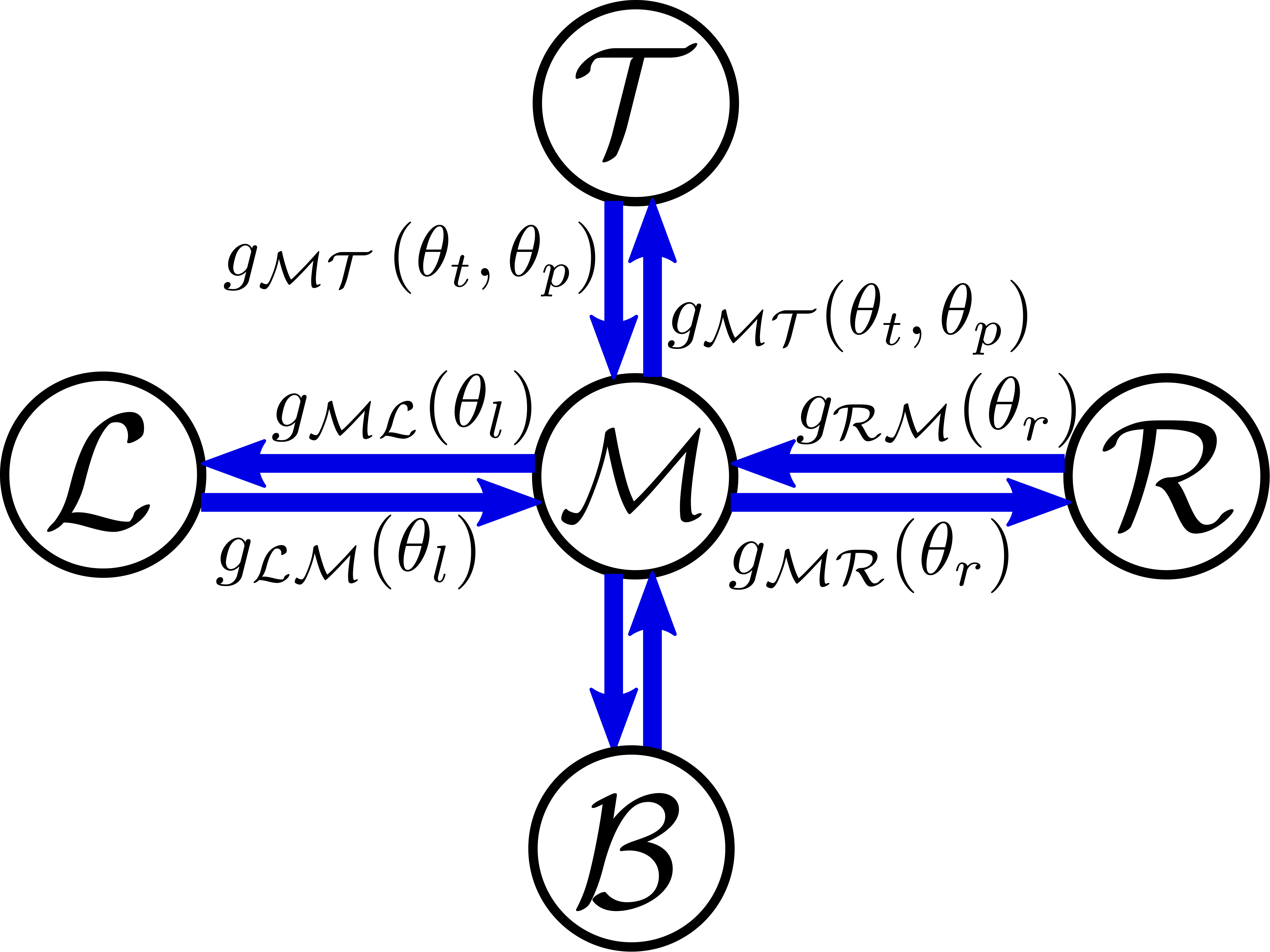}\label{fig:smres-graph}}}
  \end{subfloat}
  \caption{(a) A SMORES-EP module has four DoFs and four
    connectors. (b) The frames of all rigid bodies are shown and
    $\mathcal{B}$ is fixed in $\mathcal{M}$. (c) The module graph of a
    SMORES-EP module in which $g_{\mathcal{M}\mathcal{B}}$ and
    $g_{\mathcal{B}\mathcal{M}}$ are invariant of
    $\Theta = (\theta_l, \theta_r, \theta_p, \theta_t)$.}
  \label{fig:smores-model}
\end{figure}

\begin{figure}[t]
  \centering
  \begin{tabular}{cc}
    \multirow{2}[1]{*}[80pt]{\subfloat[]{\includegraphics[height=0.35\textwidth]{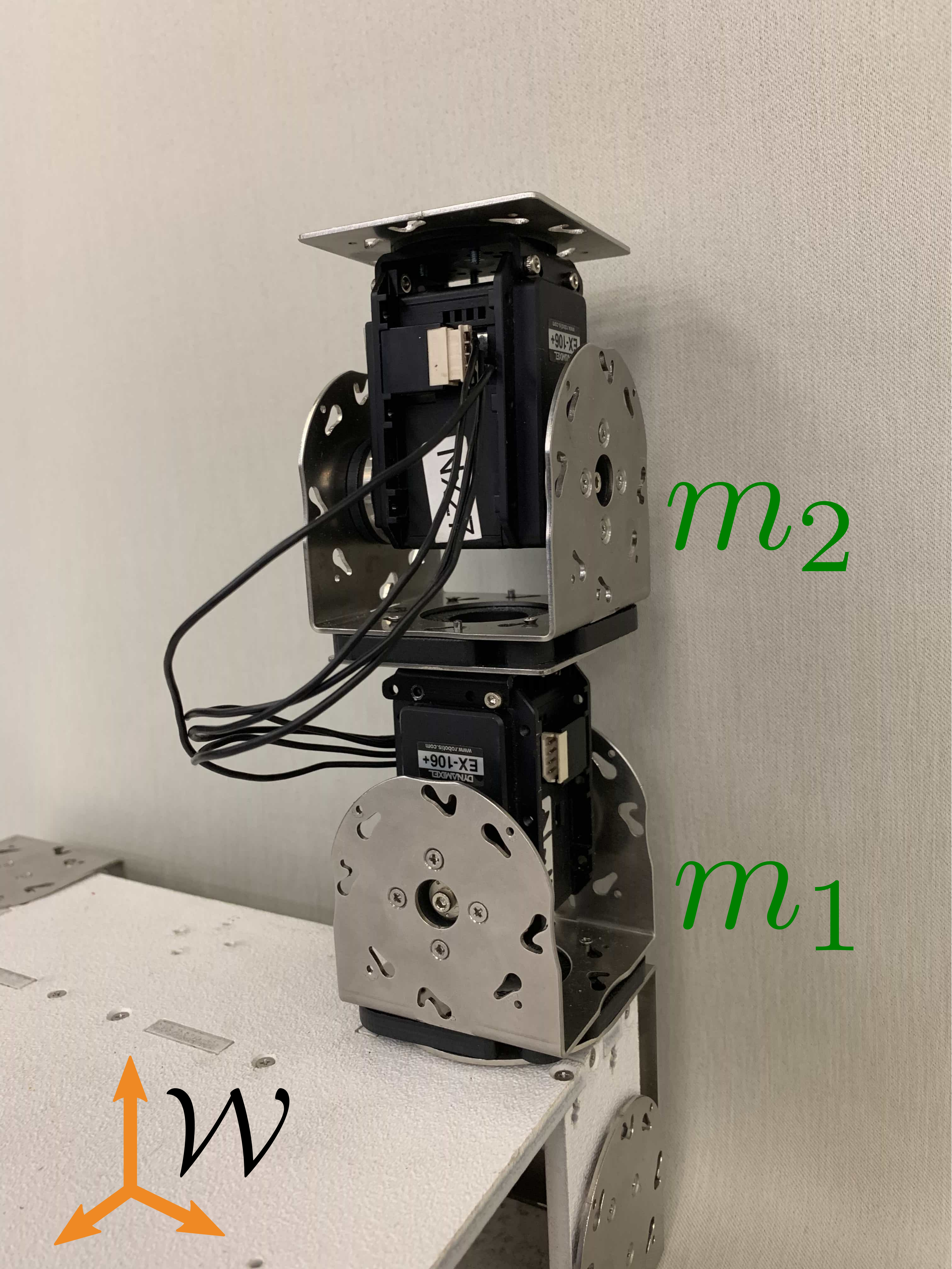}\label{fig:hardware-config}}}
    &\subfloat[]{\includegraphics[width=0.7\textwidth]{two-ckbots-graph}\label{fig:kinematics-tree-model}}\\
    &\subfloat[]{\includegraphics[width=0.7\textwidth]{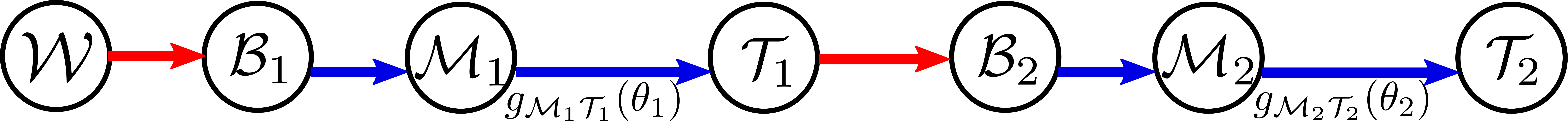}\label{fig:kinematics-chain-model}}
  \end{tabular}
  \caption{(a) A configuration by two CKBot UBar modules. (b) The
    kinematics graph model of the configurations. (c) The kinematic
    chain from $\mathcal{W}$ to $\mathcal{T}_2$.}
  \label{fig:config-kinematics}
\end{figure}

In general, given a module $m$ with connector set $C$ and joint set
$\Theta$, a frame $\mathcal{C}$ is attached to each connector $c\in C$ and
frame $\mathcal{M}$ is attached to the module body. Let mapping
$g_{\mathcal{F}_1\mathcal{F}_2}: Q \to SE(3)$ describe the forward
kinematics from $\mathcal{F}_1$ to $\mathcal{F}_2$ in joint space $Q$,
then $\forall c\in C$, $g_{\mathcal{M}\mathcal{C}}$ and
$g_{\mathcal{C}\mathcal{M}}$ can be defined with respect to $\Theta$. The
results for CKBot CR modules and SMORES-EP modules are shown in
Fig.~\ref{fig:ckbot-cr-graph} and Fig.~\ref{fig:smores-model}. With a
module graph model, we can easily obtain the \textbf{kinematics graph}
$G_K = (V_K, E_K)$ for a modular robot configuration which is
constructed by composing the modules by connecting connectors. A
directed edge is used to denote each connection and the transformation
between the two mating connectors is fixed since they are rigidly
connected. Using this kinematics graph, a \textit{kinematic chain}
from frame $\mathcal{F}_1$ to frame $\mathcal{F}_2$ can be derived by
following the shortest path
$G_K: \mathcal{F}_1\rightsquigarrow \mathcal{F}_2$. This creates a
graph with no loops. A simple configuration built by two CKBot UBar
modules is shown in Fig.~\ref{fig:hardware-config}. Frame
$\mathcal{W}$ is the world frame and module $m_1$ is fixed to it via
its \textrm{BOTTOM Face}. The kinematics graph for this configuration
is shown in Fig.~\ref{fig:kinematics-tree-model} and the kinematic
chain from $\mathcal{W}$ to $\mathcal{T}_2$ shown in
Fig.~\ref{fig:kinematics-chain-model}. All the edges have fixed
homogeneous transformations except for edge
$(\mathcal{M}_1,\mathcal{T}_1)$ and edge
$(\mathcal{M}_2, \mathcal{T}_2)$, and we can conclude that the forward
kinematics mapping is
$g_{\mathcal{W}\mathcal{T}_2}: \mathbb{T}^2\to SE(3)$ where
$\mathbb{T}^p$ represents the $p$-torus. However, we can also see that
all the edges in the shortest path from $\mathcal{W}$ to
$\mathcal{L}_1$ have fixed homogeneous transformations, so
$\mathcal{L}_1$ is fixed in $\mathcal{W}$.


Similar to the configuration discovery algorithm
in~\cite{Liu-config-recognition-isrr-2017}, the kinematics graph can
be built by visiting modules in breadth-first-search order. The given
configuration is traversed from the module fixed to the world frame
$\mathcal{W}$. When visiting a new module $m$, denoting its parent via
its connector $c$ as $\tilde{m}$ and the mating connector of
$\tilde{m}$ as $\tilde{c}$, record the fixed homogeneous
transformation $g_{\mathcal{C}\tilde{\mathcal{C}}}$ in which frame
$\mathcal{C}$ and frame $\widetilde{\mathcal{C}}$ are attached to $c$
and $\tilde{c}$ respectively. Not until all modules are visited, is
the $G_K = (V_K, E_K)$ of the given configuration constructed. With
this structure, there is no need for case-by-case derivation of the
kinematics as long as the kinematics for each type of module and
connection are defined.

\subsection{Kinematics for Modules}
\label{sec:mod-kinematics}

Recall that given a module $m$ with connector set $C$ and joint set
$\Theta$, a frame $\mathcal{C}$ is attached to each connector $c\in C$ and
frame $\mathcal{M}$ is attached to the module body. For a joint
$\theta\in \Theta$, a twist $\hat{\xi}_{\theta}\in se(3)$ can be defined with respect to
$\mathcal{M}$ in which
$\xi_\theta= (v_\theta, \omega_\theta)\in \mathbb{R}^6$ is the twist coordinates for
$\hat{\xi}_{\theta}$\footnote{Refer to Chapter 2 in~\cite{robotics-math} for
  background.}, and $\bm{\xi}$ is the set of the twist associated with
each joint. For homogeneous transformation $g_{\mathcal{M C}}$, it is
straightforward to have
\begin{equation}
  \label{eq:module-forward}
  g_{\mathcal{M C}} = g_{\mathcal{M C}}(\Theta^{\mathcal{C}}) = \prod_i
  \exp(\hat{\xi}_{\Theta_i^{\mathcal{C}}}\Theta_i^{\mathcal{C}})\ g_{\mathcal{M C}}(0)
\end{equation}
in which $\Theta^{\mathcal{C}}$ denotes the parameter vector in the joint
space of the kinematic chain from $\mathcal{M}$ to $\mathcal{C}$. If
no joints are involved in the kinematic chain from $\mathcal{M}$ to
$\mathcal{C}$, then $\mathcal{C}$ is fixed in $\mathcal{M}$ and
$g_{\mathcal{M C}}$ is a constant determined by the geometry of the
module. $g_{\mathcal{C M}}$ is just the inverse of
$g_{\mathcal{M C}}$.

\begin{figure}[t]
  \centering
  \begin{tabular}{cccc}
    \multirow{2}{*}[50pt]{\subfloat[]{\includegraphics[width=0.35\textwidth]{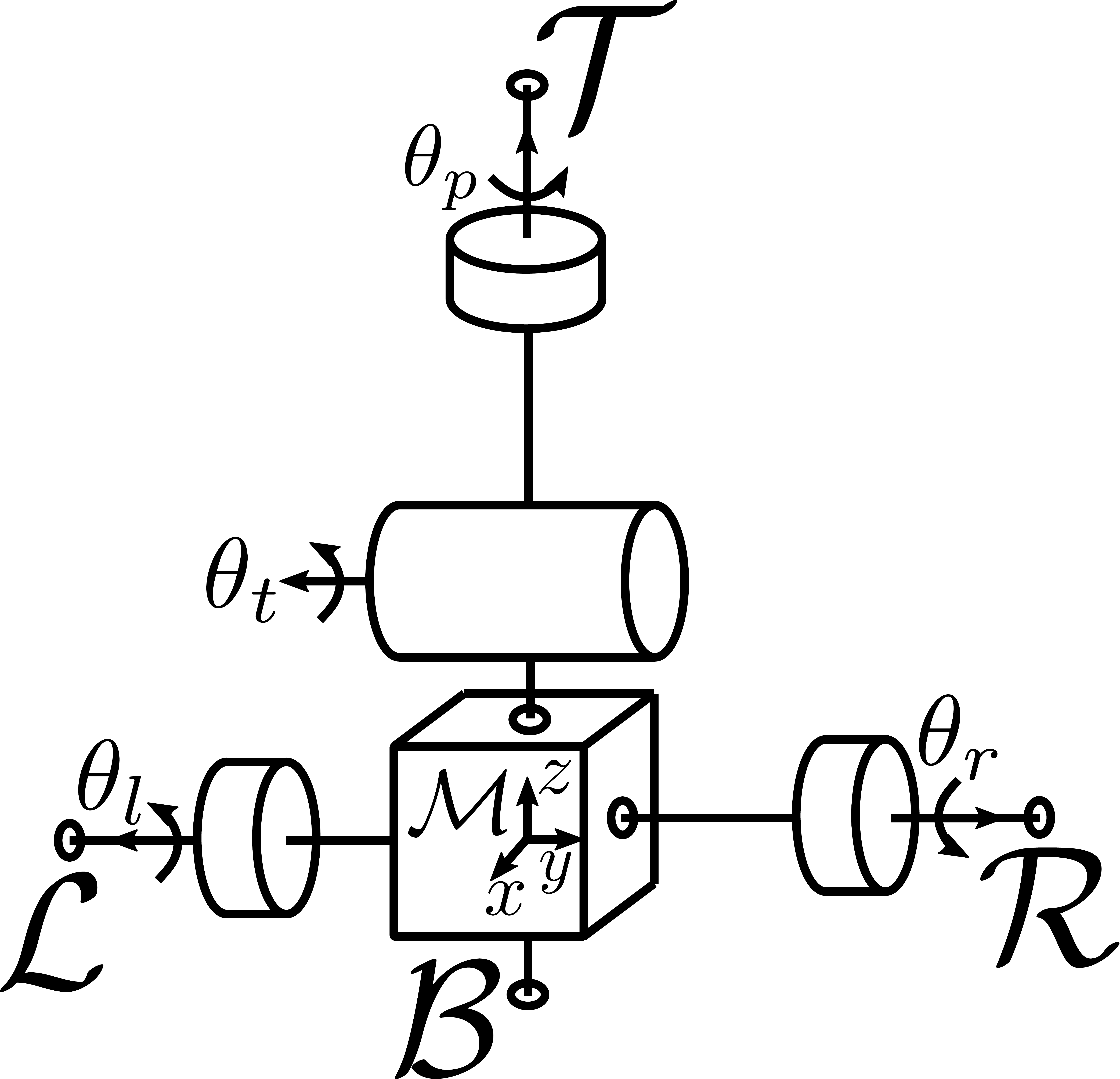}\label{fig:smores-kinematics}}}
    &\quad&\subfloat[]{\includegraphics[width=0.2\textwidth]{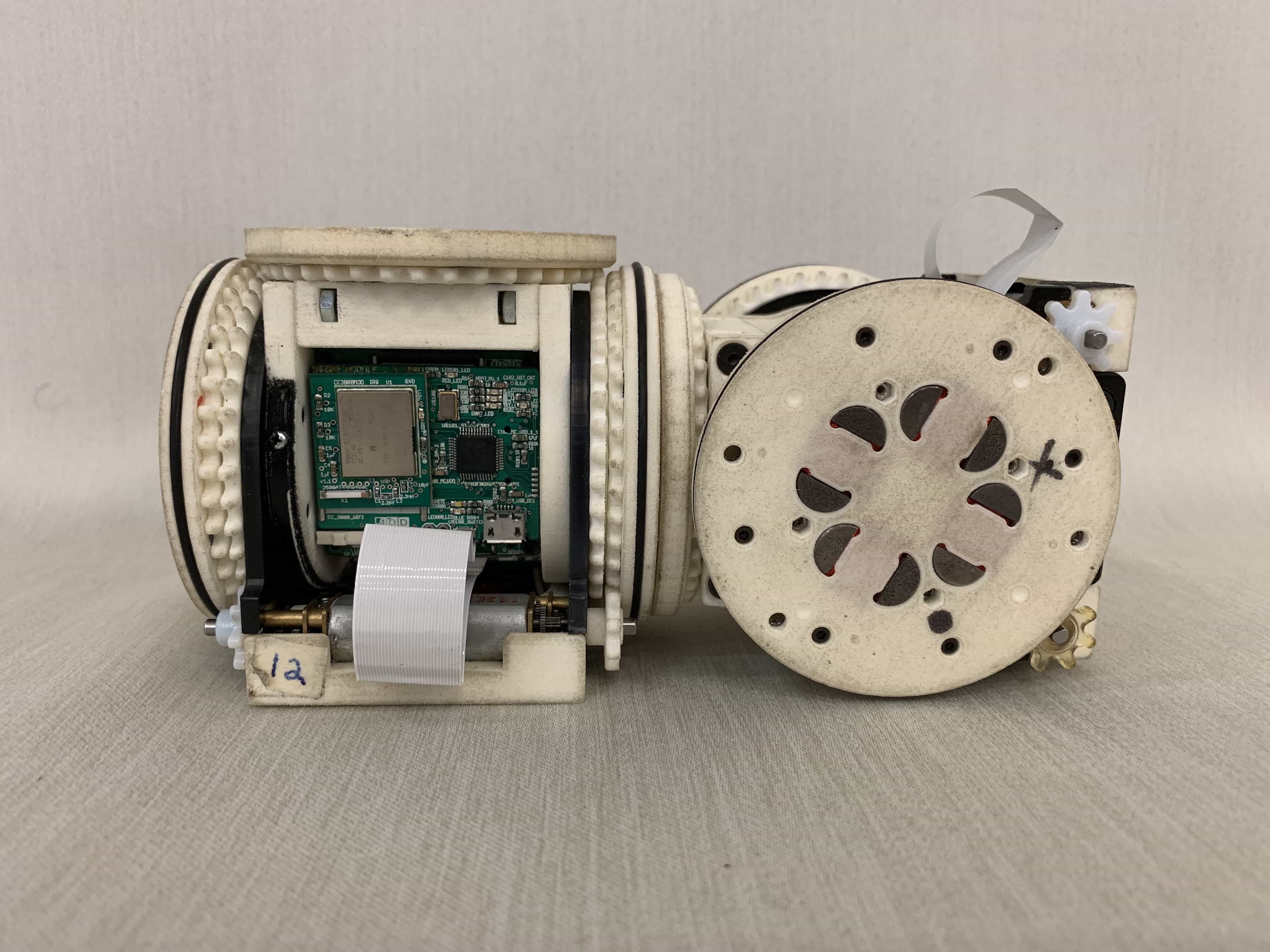}\label{fig:smores-con-1}}
    &\subfloat[]{\includegraphics[width=0.2\textwidth]{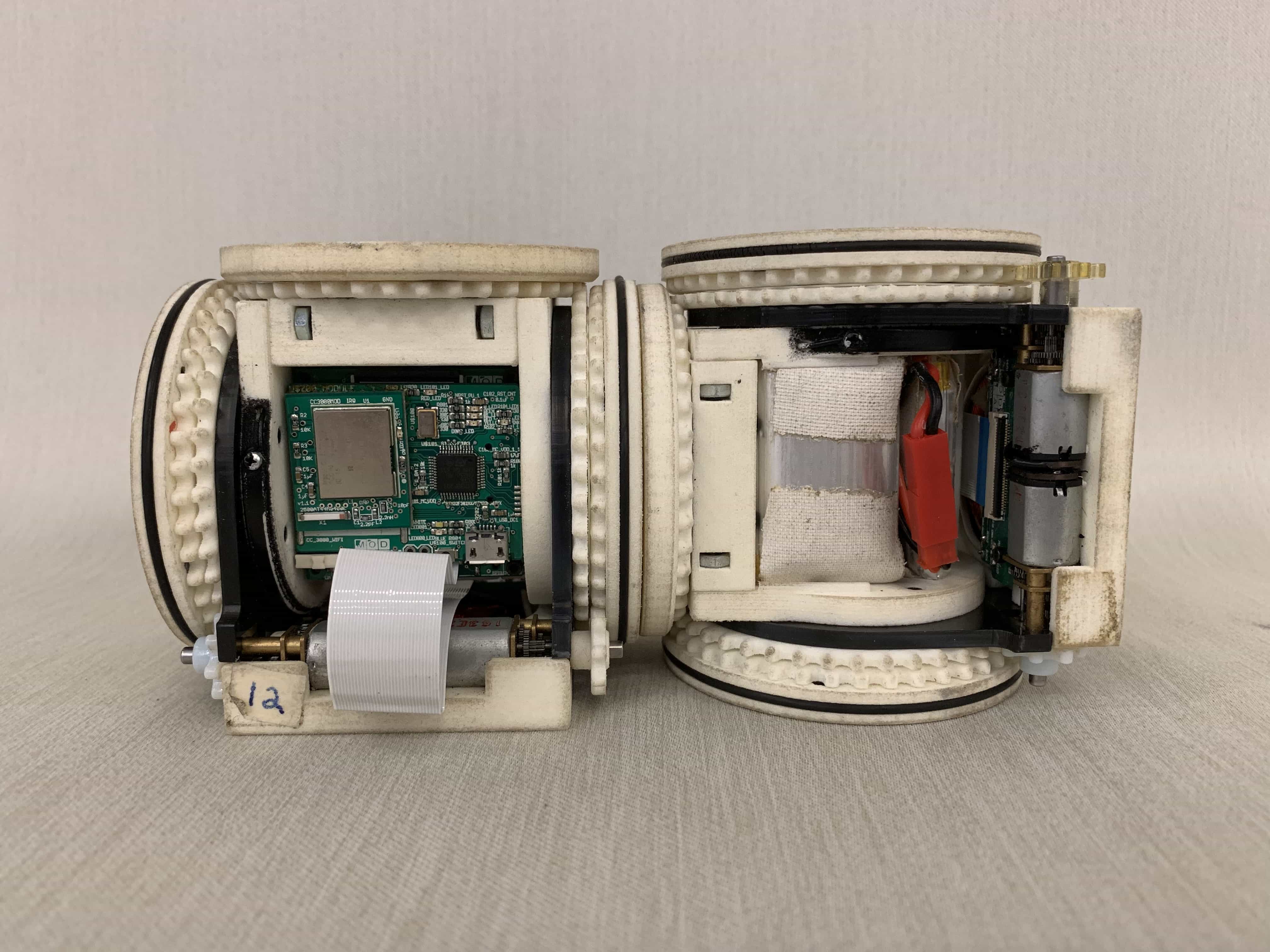}}\\
    &\quad&\subfloat[]{\includegraphics[width=0.2\textwidth]{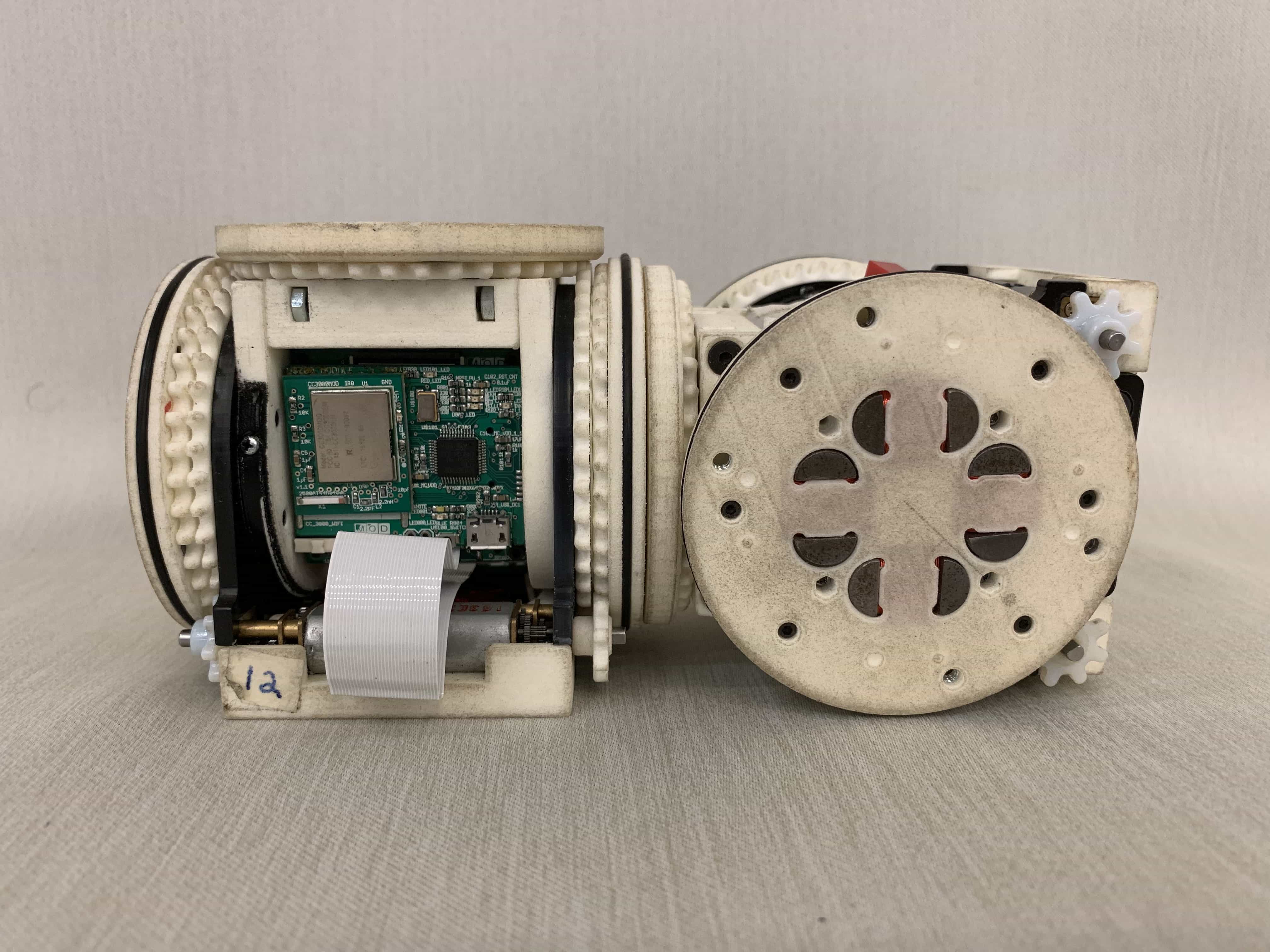}}
    &\subfloat[]{\includegraphics[width=0.2\textwidth]{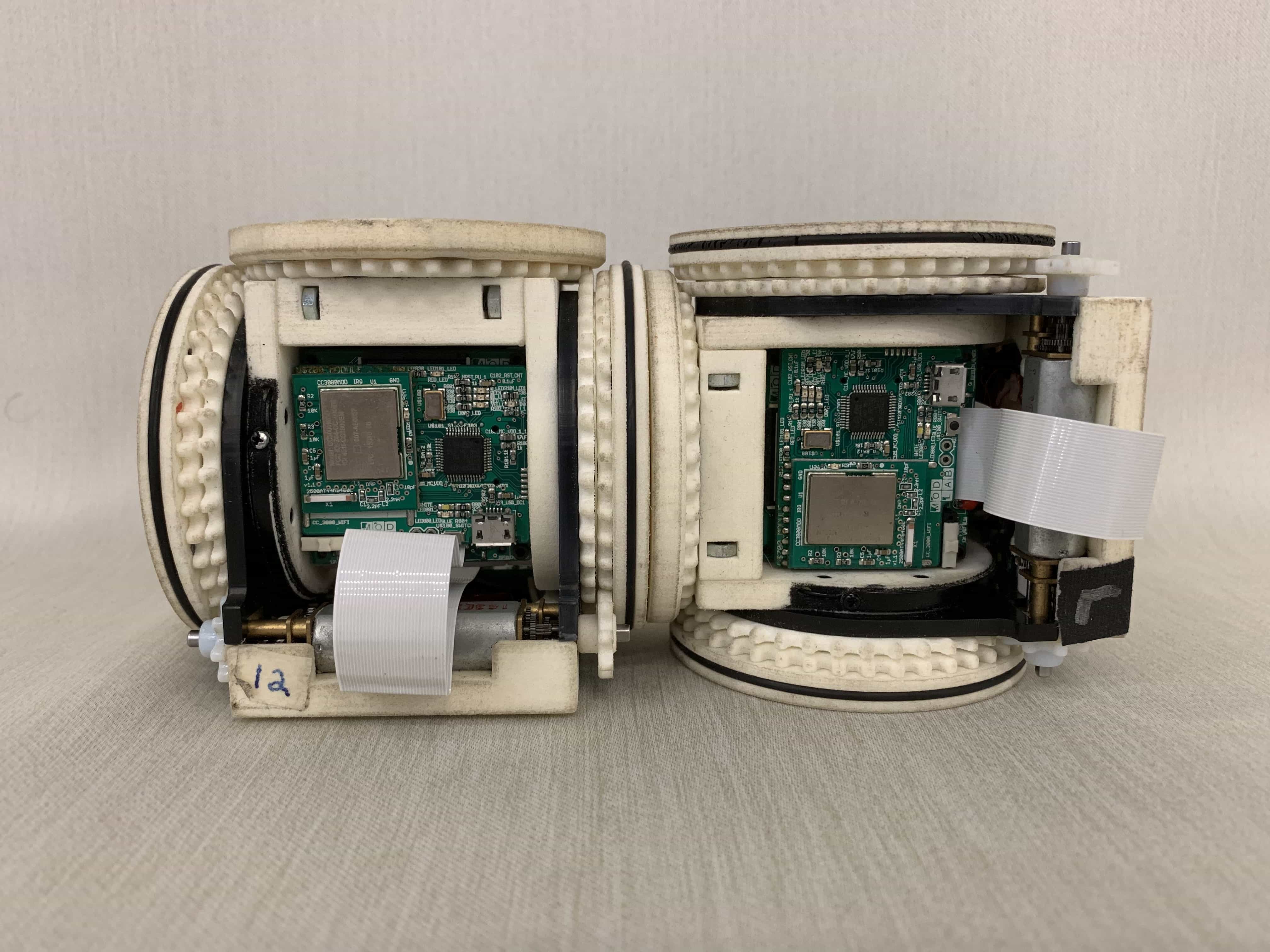}\label{fig:smores-con-4}}\\
  \end{tabular}
  \caption{(a) Kinematics for SMORES modules. (b) --- (e) Four cases to
    connect $\mathcal{R}$ and $\mathcal{T}$.}
\end{figure}

\subsection{Kinematics for Chains}
\label{sec:reach-jac}

A kinematic chain from frame $\mathcal{S}$ to $\mathcal{F}$ can be
obtained as $G_K: \mathcal{S} \rightsquigarrow \mathcal{F}$ where
$\mathcal{S}$ and $\mathcal{F}$ are two vertices of $G_K$. In this
kinematic chain, all homogeneous transformations between connectors
(e.g., $g_{\mathcal{T}_1 \mathcal{B}_2}$ in
Fig.~\ref{fig:kinematics-chain-model}) are fixed and can be easily
computed. The relative orientation between connectors is determined by
examining the connector design. For example, there are four cases for
connecting SMORES-EP modules shown in Fig.~\ref{fig:smores-con-1} ---
Fig.~\ref{fig:smores-con-4} due to the arrangement of the magnets on
the connector. The homogeneous transformation $g_{\mathcal{S F}}$ can
be computed by multiplying the homogeneous transformation of each edge
of path $G_K: \mathcal{S} \rightsquigarrow \mathcal{F}$ in order. In
particular, let $\mathcal{S}$ be world frame $\mathcal{W}$, if module
$m_1, m_2, \cdots, m_N$ are involved in this chain, then the position of
the origin of $\mathcal{F}$ in $\mathcal{W}$ is given by
\begin{equation}
  \label{eq:frame-pos}
  p_{\mathcal{F}}^{\mathcal{W}} =
  g_{\mathcal{W}\mathcal{F}}
  \left[\begin{array}{cccc}
          0&0&0&1
        \end{array}\right]^\intercal
\end{equation}
and the instantaneous spatial velocity of $\mathcal{F}$ is given by
the twist
\begin{equation}
  \label{eq:tool-jacobian}
    \widehat{V}_{\mathcal{W F}}^s = \sum_{i=1}^N \sum_{j=1}^{N_i}\left(\frac{\partial g_{\mathcal{W
          F}}}{\partial\theta_{ij}} g_{\mathcal{W
        F}}^{-1}\right)\dot{\theta}_{ij}
\end{equation}
in which $\theta_{ij}$ is the $j$th joint parameter of module $m_i$
involved in this chain and the number of joints of module $m_i$
involved in this chain is $N_i$. Rewrite Eq.~\eqref{eq:tool-jacobian}
in twist coordinates as
\begin{equation}
  \label{eq:tool-jacobian-twist}
  V_{\mathcal{W F}}^s = J_{\mathcal{W F}}^s \dot{\Theta}^{\mathcal{W F}}
\end{equation}
in which
\begin{gather}
  \label{eq:tool-jacobian-rep}
  \Theta^{\mathcal{W F}} = \left[\theta_{11} \cdots \theta_{1 N_1}\ \theta_{21} \cdots \theta_{2 N_1}\ \cdots\ \theta_{N1} \cdots
    \theta_{N N_n}\right]^\intercal\\
  J_{\mathcal{W F}}^s = \left[
    \begin{array}{cccc}
      J_1&J_2&\cdots&J_N
    \end{array}
  \right]\\
  J_i = \left[
    \begin{array}{cccc}
      \left(\frac{\partial g_{\mathcal{W F}}}{\partial\theta_{i1}} g_{\mathcal{W
      F}}^{-1}\right)^\vee&\left(\frac{\partial g_{\mathcal{W F}}}{\partial\theta_{i2}} g_{\mathcal{W
                   F}}^{-1}\right)^\vee&\cdots&\left(\frac{\partial g_{\mathcal{W F}}}{\partial\theta_{i
                                 N_i}} g_{\mathcal{W F}}^{-1}\right)^\vee
    \end{array}
  \right]
\end{gather}
and $J_{\mathcal{W F}}^s$ is the \textit{spatial chain Jacobian}.

Define the twist of the $j$th joint of module $m_i$ with respect to
$\mathcal{W}$ as $\xi_{ij}^\prime$ that is
\begin{equation*}
  \xi_{ij}^\prime = \left(\frac{\partial g_{\mathcal{W F}}}{\partial\theta_{ij}} g_{\mathcal{W F}}^{-1}\right)^\vee =
  \mathrm{Ad}_{g_{\mathcal{W}\mathcal{M}_i}}\xi_{ij}
\end{equation*}
in which $\mathrm{Ad}_{g_{\mathcal{W} \mathcal{M}_i}}$ is the adjoint
transformation\footnote{Refer to Chapter 2 in~\cite{robotics-math} for
  adjoint transformation definition.}  and $\xi_{ij}$ is defined in
Sec.~\ref{sec:mod-kinematics} for each joint in a module with respect
to its module body frame. Then $J_i$ becomes
\begin{equation}
  \label{eq:tool-jacobian-twist-rep}
  J_i = \left[
    \begin{array}{cccc}
      \xi_{i1}^\prime&\xi_{i2}^\prime&\cdots&\xi_{i N_i}^\prime
    \end{array}
\right]
\end{equation}
With this spatial chain Jacobian, the velocity of the origin of frame
$\mathcal{F}$ is
\begin{equation}
  \label{eq:tool-vel}
  v_{\mathcal{F}}^s = \widehat{V}_{\mathcal{W F}}^s
  p_{\mathcal{F}}^{\mathcal{W}} = \left(J_{\mathcal{W F}}^s
  \dot{\Theta}^{\mathcal{W F}}\right)^\wedge p_{\mathcal{F}}^{\mathcal{W}}
\end{equation}

For a module $m_i$ in the kinematic chain
$G_K: \mathcal{W}\rightsquigarrow \mathcal{F}$ ($\mathcal{M}_i$ is a
vertex in the corresponding path), a sub-kinematic chain
$G_K: \mathcal{W}\rightsquigarrow \mathcal{M}_i$ can be defined with
joint parameter vector
$\Theta^{\mathcal{W}\mathcal{M}_i} = \left[\theta_{11}, \theta_{12}, \cdots,
  \theta_{\bar{i}\bar{j}_i}\right]^\intercal$ where
$\theta_{\bar{i}\bar{j}_i}$ is the parameter of the $\bar{j}_i$th joint of
module $m_{\bar{i}}$. For example, take the sub-kinematic chain from
$\mathcal{W}$ to $\mathcal{M}_2$ in
Fig.~\ref{fig:kinematics-chain-model}, then $i=2$, $\bar{i} = 1$,
$\bar{j}_i = 1$, since there is only one joint between $\mathcal{W}$
and $\mathcal{M}_2$ which is the 1st joint of module $m_1$. Then the
\textit{spatial module Jacobian} $J_{\mathcal{W}\mathcal{M}_i}^s$ or
$J_{\mathcal{M}_i}^s$ for simplicity can be defined as
\begin{equation}
  \label{eq:module-jac}
  J_{\mathcal{M}_i}^s = \left[
    \begin{array}{cccc}
      \xi_{11}^\prime&\xi_{12}^\prime&\cdots&\xi_{\bar{i}\bar{j}_i}^\prime
    \end{array}
  \right]
\end{equation}
and the velocity of the origin of $\mathcal{M}_i$ is
\begin{equation}
  \label{eq:module-vel}
    v_{\mathcal{M}_i}^s =
    \left(J_{\mathcal{M}_i}^s\dot{\Theta}^{\mathcal{W}\mathcal{M}_i}\right)^\wedge
    p_{\mathcal{M}_i}^{\mathcal{W}}
\end{equation}
By replacing all twists associated with joints after the $\bar{j}_i$th
joint of module $m_{\bar{i}}$ in the spatial chain Jacobian of chain
$G_K: \mathcal{W}\rightsquigarrow \mathcal{F}$ with $6\times1$ zero
vectors, the spatial module Jacobian can also be written as
\begin{equation}
  \label{eq:module-jac-new}
  J_{\mathcal{M}_i}^s = \left[
    \begin{array}{ccccccc}
      \xi_{11}^\prime&\xi_{12}^\prime&\cdots&\xi_{\bar{i}\bar{j}_i}^\prime&0_{6\times1}&\cdots&0_{6\times1}
    \end{array}
  \right]
\end{equation}
then the velocity of the origin of $\mathcal{M}_i$ is represented as
\begin{equation}
  \label{eq:module-vel-new}
    v_{\mathcal{M}_i}^s =
    \left(J_{\mathcal{M}_i}^s\dot{\Theta}^{\mathcal{W F}}\right)^\wedge
    p_{\mathcal{M}_i}^{\mathcal{W}}
  \end{equation}

  \section{Control and Motion Planning}
\label{sec:control-planning}

\subsection{Control}
\label{sec:control}

Given the kinematic chain
$G_K: \mathcal{W}\rightsquigarrow \mathcal{F}$, the goal of the
control task is to move $p_{\mathcal{F}}^{\mathcal{W}}$ (or
$p_{\mathcal{F}}$ for simplicity) --- the position of $\mathcal{F}$ --- to
follow a desired trajectory.

Let $\tilde{p}_{\mathcal{F}} = \tilde{p}_{\mathcal{F}}(t)$ be the
desired trajectory for the robot to track and
$\tilde{v}_{\mathcal{F}}^s$ (or $\tilde{v}_{\mathcal{F}}$ for
simplicity) is the derivative of $\tilde{p}_{\mathcal{F}}$, and the
error and its derivative are defined as
\begin{gather*}
  e = \tilde{p}_{\mathcal{F}} - p_{\mathcal{F}}\\
  \dot{e} = \dot{\tilde{p}}_{\mathcal{F}} - \dot{p}_{\mathcal{F}} =
  \tilde{v}_{\mathcal{F}}  - v_{\mathcal{F}}
\end{gather*}
The error $e$ can converge exponentially to zero as long as it
satisfies
\begin{equation}
  \label{eq:control-eq}
  \dot{e} + Ke = 0
\end{equation}
in which $K$ is positive definite. Substitute $e$ and $\dot{e}$
\begin{equation}
  \label{eq:control-eq-state}
  \tilde{v}_{\mathcal{F}}^s - v_{\mathcal{F}}^s +
  K(\tilde{p}_{\mathcal{F}} - p_{\mathcal{F}}) = 0
\end{equation}
With Eq.~\eqref{eq:tool-vel}, Eq.~\eqref{eq:control-eq-state} can be
rewritten as
\begin{equation}
  \label{eq:control-law}
  {(J_{\mathcal{W F}}^s\dot{\Theta}^{\mathcal{W F}})}^\wedge p_{\mathcal{F}} =
  \tilde{v}_{\mathcal{F}}^s + K(\tilde{p}_{\mathcal{F}} - p_{\mathcal{F}})
\end{equation}
Eq.~\eqref{eq:control-law} is the control law to control the position
of frame $\mathcal{F}$, namely $\dot{\Theta}^{\mathcal{W F}}$ (or
$\dot{\Theta}^{\mathcal{F}}$ for simplicity) --- the velocities of all
involved joints that satisfy this equation --- can move
$p_{\mathcal{F}}$ to $\tilde{p}_{\mathcal{F}}$ in exponential time.

Suppose there are $\alpha$ motion goals $\tilde{p}_{\mathcal{F}_1}$,
$\tilde{p}_{\mathcal{F}_2}$, $\cdots$,
$\tilde{p}_{\mathcal{F}_\alpha}$, then the control law for all motion goals
can be written as
\begin{equation}
  \label{eq:control-law-full}
  \mathbf{J}\mathbf{P} = \widetilde{\mathbf{V}} +
  \mathbf{K}(\widetilde{\mathbf{P}} - \mathbf{P})
\end{equation}
which is the stack of Eq.~\eqref{eq:control-law} for each motion
goal. This makes the control problem for multiple motion goals easier
without considering the fact that some motion goals may be
coupled. That is, some kinematic chains may share DoFs.  We need only
build an Eq.~\eqref{eq:control-law} for each individual motion goal
and then stack them as linear constraints. Building a specific model
for different combinations of motion goals is not necessary.

Recall that a modular robotic system is usually redundant so that
there can be an infinite number of solutions to
Eq.~\eqref{eq:control-law-full}. This problem is formulated as a
quadratic program
\begin{equation}
  \label{eq:qp-program}
  \begin{aligned}
    \mathrm{minimize}\quad &\dfrac{1}{2}\dot{\Theta}^\intercal \dot{\Theta}\\
    \mathrm{subject\ to}\quad &\mathbf{J}\mathbf{P} = \widetilde{\mathbf{V}} +
  \mathbf{K}(\widetilde{\mathbf{P}} - \mathbf{P})
  \end{aligned}
\end{equation}
where $\Theta$ is the set of joint parameters in kinematic chains
$G_K: \mathcal{W}\rightsquigarrow \mathcal{F}_1$,
$G_K: \mathcal{W}\rightsquigarrow \mathcal{F}_2$, $\cdots$,
$G_K: \mathcal{W}\rightsquigarrow \mathcal{F}_\alpha$. Then
solving~\eqref{eq:qp-program} yields the minimum norm solution of
joint velocities at every moment.

The joint position and velocity limits can be added to the quadratic
program as inequality constraints
\begin{gather}
  \frac{\Theta_{\min} - \Theta}{\Delta t} \le \dot{\Theta} \le \frac{\Theta_{\max} - \Theta}{\Delta
    t}\label{eq:joint-pos-limit}\\
  \dot{\Theta}_{\min} \le \dot{\Theta} \le \dot{\Theta}_{\max}\label{eq:joint-vel-limit}
\end{gather}
in which $\Delta t$ is the time duration for the current step. Due to these
two constraints, $K$ cannot be too aggressive or solutions may not be
obtained.

This optimization approach is helpful for many types of motion
tasks. The controller can be used to move $p_{\mathcal{F}}$ to a
desired position $\tilde{p}_{\mathcal{F}}$ by setting
$\tilde{v}_{\mathcal{F}}^s = 0$, and it can also control
$p_{\mathcal{F}}$ to move at a desired velocity by increasing
$\tilde{p}_{\mathcal{F}}$ by $\tilde{v}_{\mathcal{F}}\Delta t$ for every
time step.

\subsection{Motion Planning}
\label{sec:planning}

The goal of the motion planning task is to enable a cluster of modules
to navigate collision-freely in an environment with obstacles.

\subsubsection{Frame Boundaries}
\label{sec:frame-boundary}

\begin{figure}[t!]
  \centering
  \subfloat[]{\includegraphics[width=0.25\textwidth]{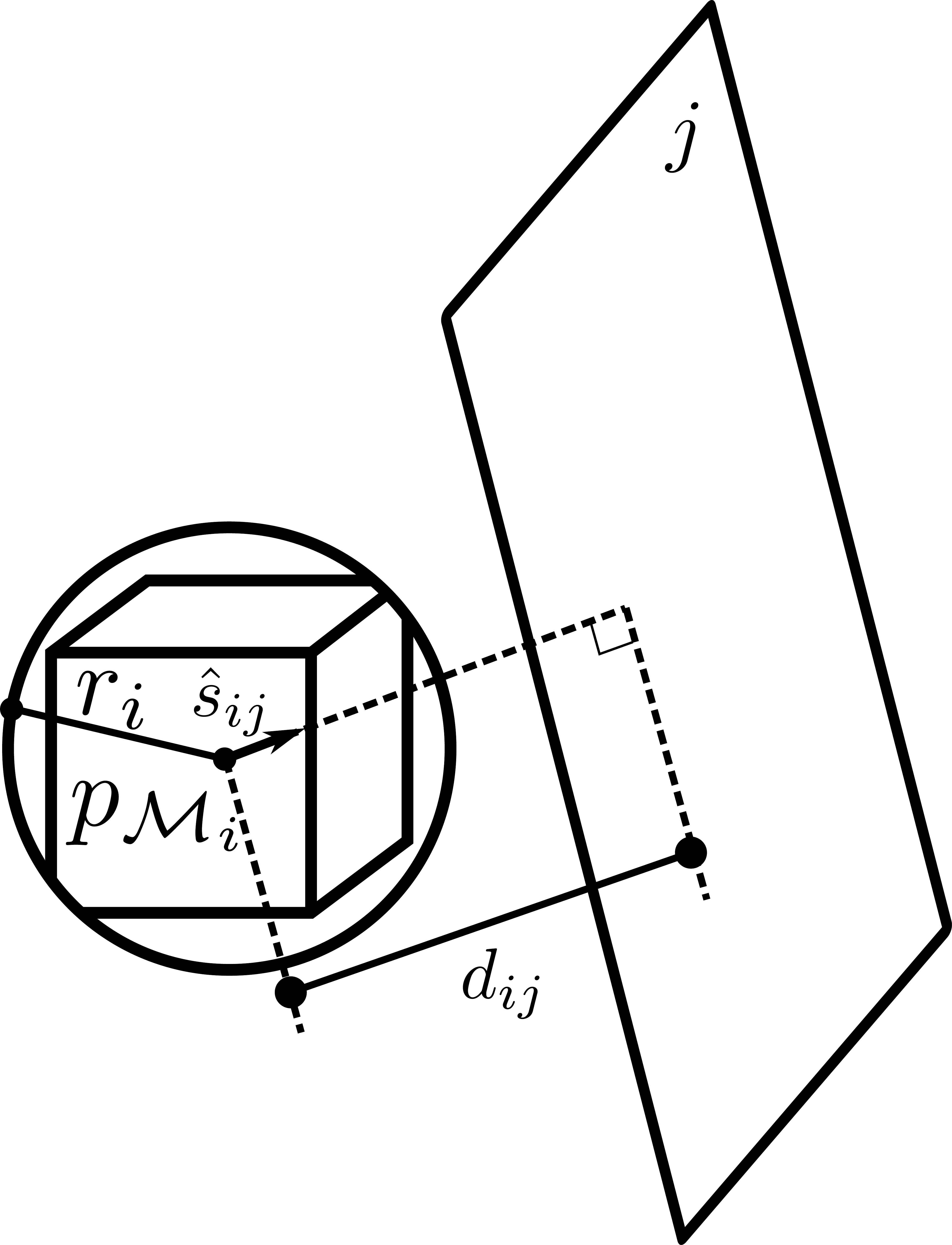}\label{fig:frame-boundary}}
  \hfil
  \subfloat[]{\includegraphics[width=0.4\textwidth]{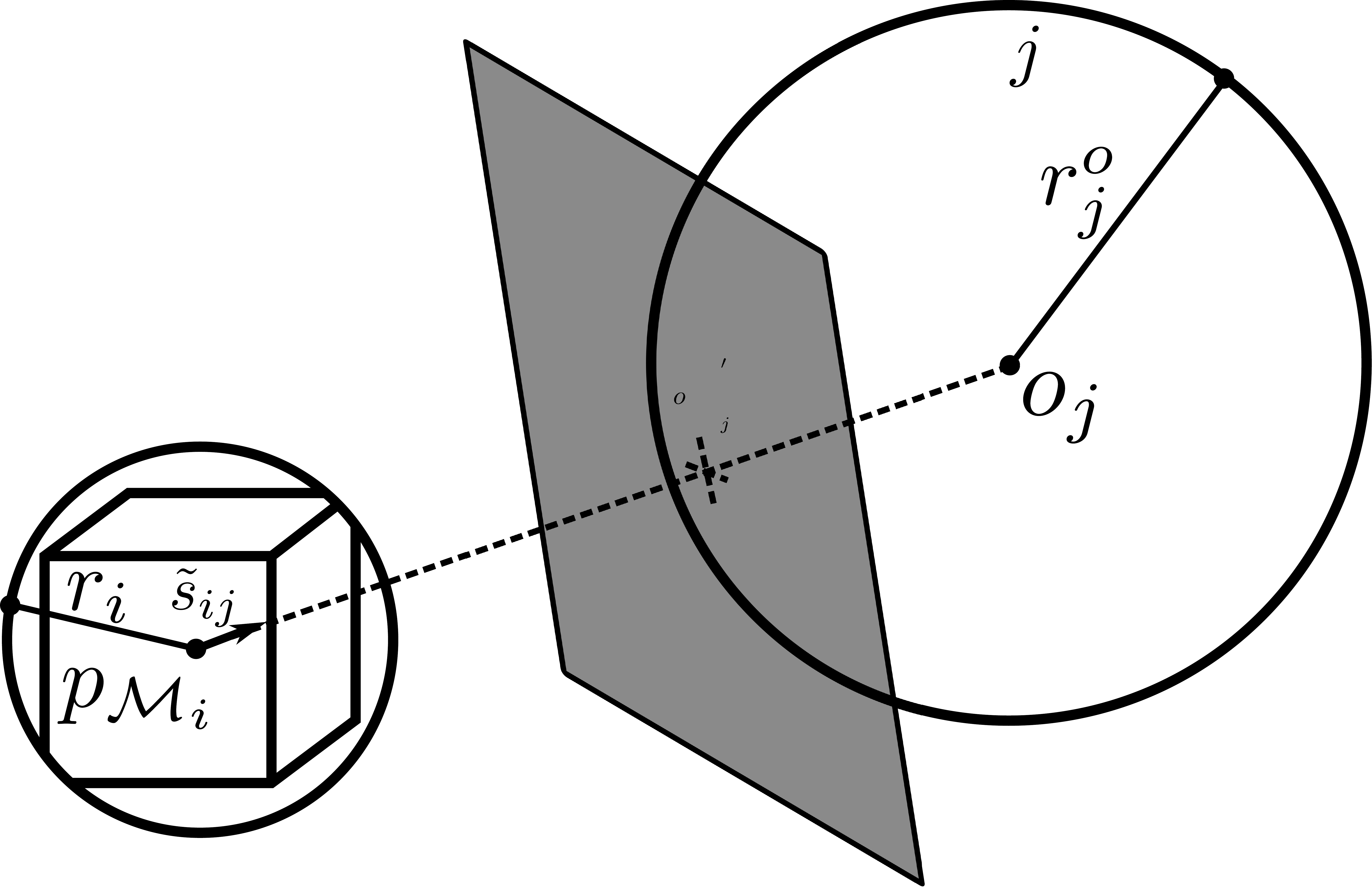}\label{fig:obstacle}}
  \caption{(a) Environment boundary. (b) Sphere obstacle avoidance.}
\end{figure}

The cluster of modules can be kept in any polyhedral region in space
which is defined by the boundaries of the environment. For a module
$m_i$ in the kinematic chain
$G_K: \mathcal{W}\rightsquigarrow \mathcal{F}$, let $\hat{s}_{ij}$ be
the unit direction vector from $p_{\mathcal{M}_i}^{\mathcal{W}}$ (or
$p_{\mathcal{M}_i}$ for simplicity) --- the origin of $\mathcal{M}_i$ in
world frame $\mathcal{W}$ --- to the $j$th face of the environment
polyhedron perpendicular with distance $d_{ij}$, then if we enforce
the constraint
\begin{equation}
  \label{eq:frame-boundary-cons}
  v_{\mathcal{M}_i}^s \bullet \hat{s}_{ij} = (J_{\mathcal{M}_i}^s\dot{\Theta})^\wedge
  p_{\mathcal{M}_i} \bullet \hat{s}_{ij}\le d_{ij}
\end{equation}
for every side of the environment polyhedron, $p_{\mathcal{M}_i}$ will
never cross the boundary of the environment as long as this kinematic
chain follows the velocity for much less than 1 second. Using a sphere
with radius $r_i$ to approximate the geometry size of module $m_i$,
then the constraint
\begin{equation}
  \label{eq:frame-boundary-cons-mod}
  v_{\mathcal{M}_i}^s \bullet \hat{s}_{ij} = (J_{\mathcal{M}_i}^s\dot{\Theta})^\wedge
  p_{\mathcal{M}_i} \bullet \hat{s}_{ij}\le d_{ij} - r_i
\end{equation}
will ensure that the module body will always be inside the environment
boundaries (Fig.~\ref{fig:frame-boundary}). Thus, applying
constraint~\eqref{eq:frame-boundary-cons-mod} to all modules in the
kinematic chain will ensure the chain will stay inside the
environment.

\subsubsection{Obstacle Avoidance}
\label{sec:obstacle}

\begin{figure}[t]
  \centering
  \subfloat[]{\includegraphics[width=0.25\textwidth]{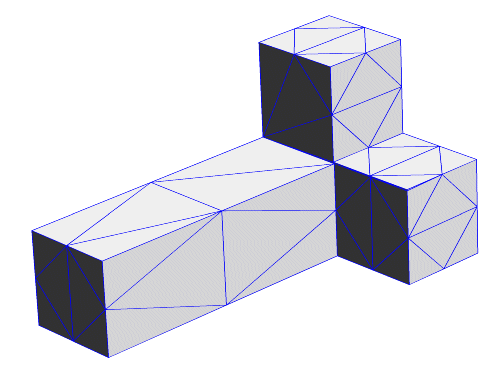}\label{fig:block-raw}}
  \subfloat[]{\includegraphics[width=0.25\textwidth]{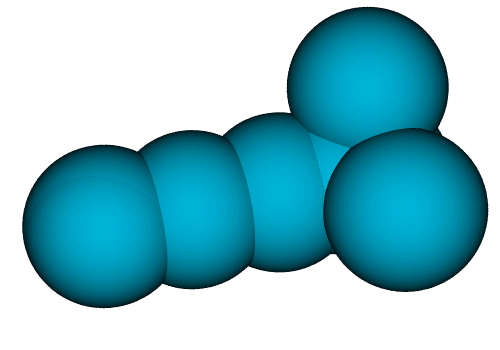}\label{fig:block-level1}}
  \subfloat[]{\includegraphics[width=0.25\textwidth]{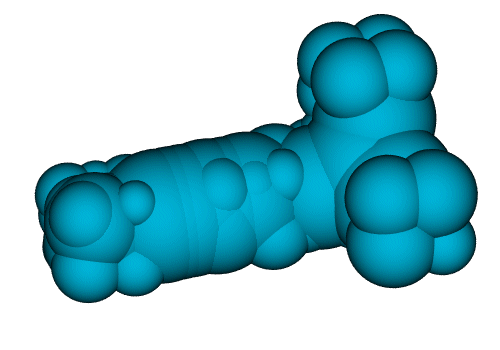}\label{fig:block-level2}}
  \subfloat[]{\includegraphics[width=0.25\textwidth]{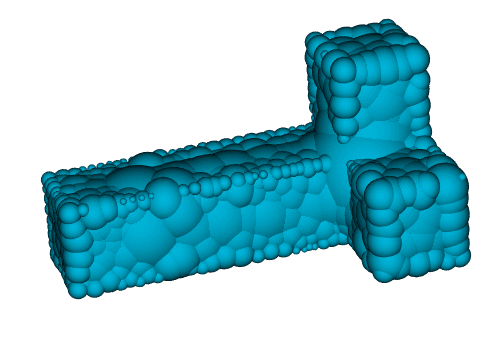}\label{fig:block-level3}}
  \caption{A block obstacle (a) is approximated with 3 levels of
    spheres. (b) 8 spheres in level 1. (c) 64 spheres in level 2. (d)
    470 spheres in level 3.}
\end{figure}

It is hard to represent the collision-free space analytically in joint
space due to the high DoFs of modular robotic systems. Here we propose
an alternative. The obstacles can be approximated by a set of spheres
using a sphere-tree construction
algorithm~\cite{sphere-tree-construct-2004}. Similar ideas have been
explored
in~\cite{Fromherz-modular-robot-control-2001,Tomizuka-arm-trajectory-2018}. There
are two issues using this idea. This collision-avoidance constraint is
modeled as the condition that the distance between every sphere
approximating the robot and every sphere approximating the obstacles
is greater than the sum of their radius. This leads to quadratic
constraints which are not suitable for real-time applications of large
systems due to numerical issues. In addition, in order to approximate
obstacles with decent accuracy, many spheres have to be generated. For
example, a block object shown in Fig.~\ref{fig:block-raw} is
constructed by multiple spheres. A more accurate approximation of this
object requires more spheres (Fig.~\ref{fig:block-level1} ---
Fig.~\ref{fig:block-level3}). An advantage of this approach is that
obstacles automatically have some level of buffer that can further
guarantee motion safety. However, using a small number of spheres to
approximate an obstacle can lead to too conservative planning space
not finding collision-free paths when they exist. On the other hand,
if there are a large number of spheres, there will also be a large
number of constraints which can prohibit the optimization problem
being solved efficiently without numerical issues.

In this paper, the obstacle avoidance requirement is modeled as linear
constraints which are efficient to solve stably. For a module $m_i$ in
the kinematic chain $G_K: \mathcal{W}\rightsquigarrow \mathcal{F}$,
let $\tilde{s}_{ij}$ be the unit direction vector from
$p_{\mathcal{M}_i}$ to the center of the $j$th obstacle sphere $o_j$
in world frame $\mathcal{W}$ with radius $r_j^o$. Imaging a plane
$P_{ij}$ with $\tilde{s}_{ij}$ as its normal vector and $o_j^\prime$ being
the point of tangency to this sphere, then if we enforce the
constraint
\begin{equation}
  \label{eq:obstacle-cons}
  v_{\mathcal{M}_i}^s \bullet \tilde{s}_{ij} = (J_{\mathcal{M}_i}^s\dot{\Theta})^\wedge
  p_{\mathcal{M}_i}\bullet \tilde{s}_{ij} \le \|o_j^\prime - p_{\mathcal{M}_i}\| - r_i
\end{equation}
in which $o_j^{\prime} = o_j - r_j^os_{{ij}}$ for every obstacle sphere,
$p_{\mathcal{M}_i}$ will never touch an obstacle
(Fig.~\ref{fig:obstacle}). In order to enable the system to safely
navigate the environment, we need to apply this constraint for every
module.

\begin{figure}[b!]
  \centering
  \subfloat[]{\includegraphics[width=0.45\textwidth]{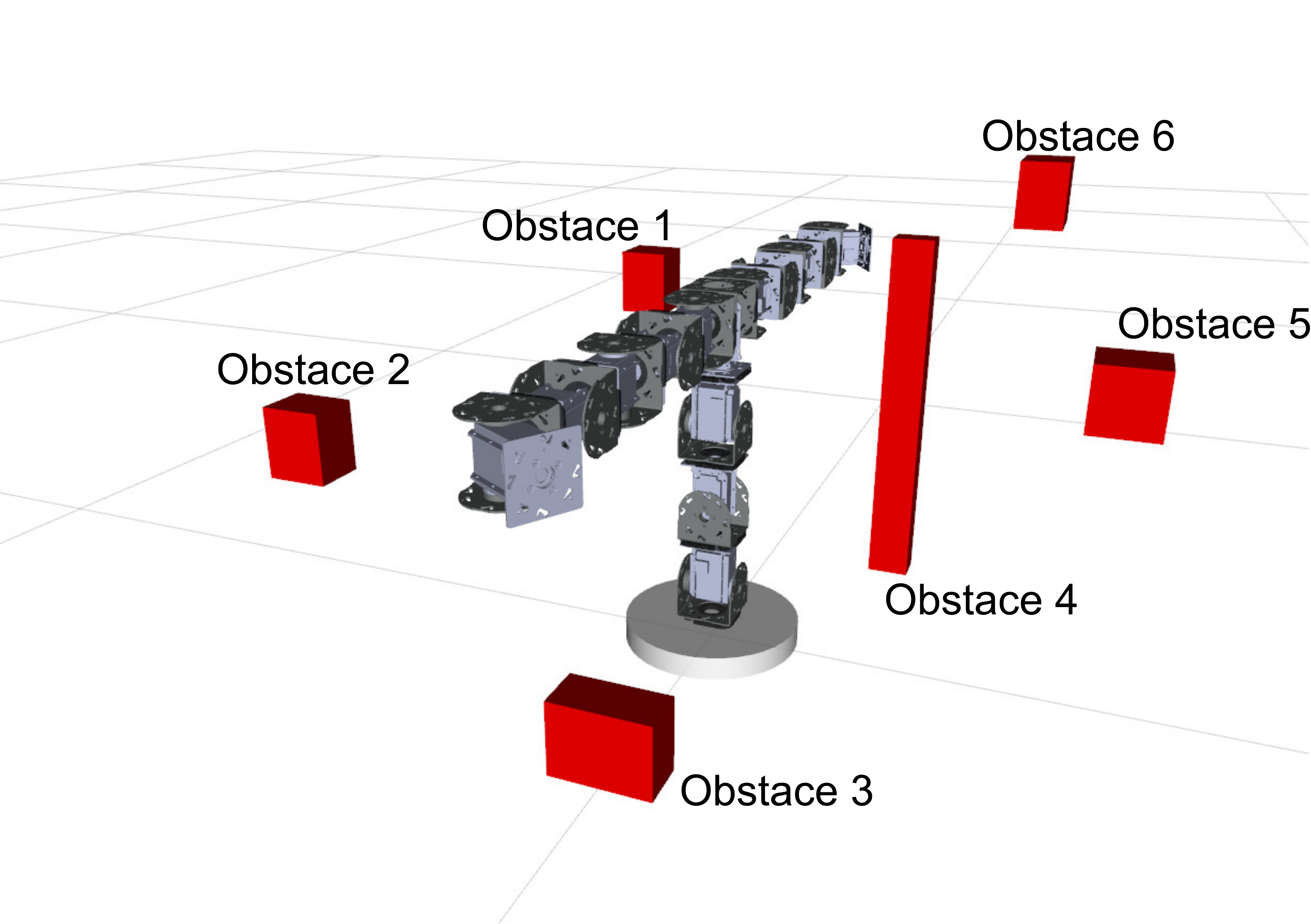}\label{fig:raw-obstacle}}
  \hfil
  \subfloat[]{\includegraphics[width=0.45\textwidth]{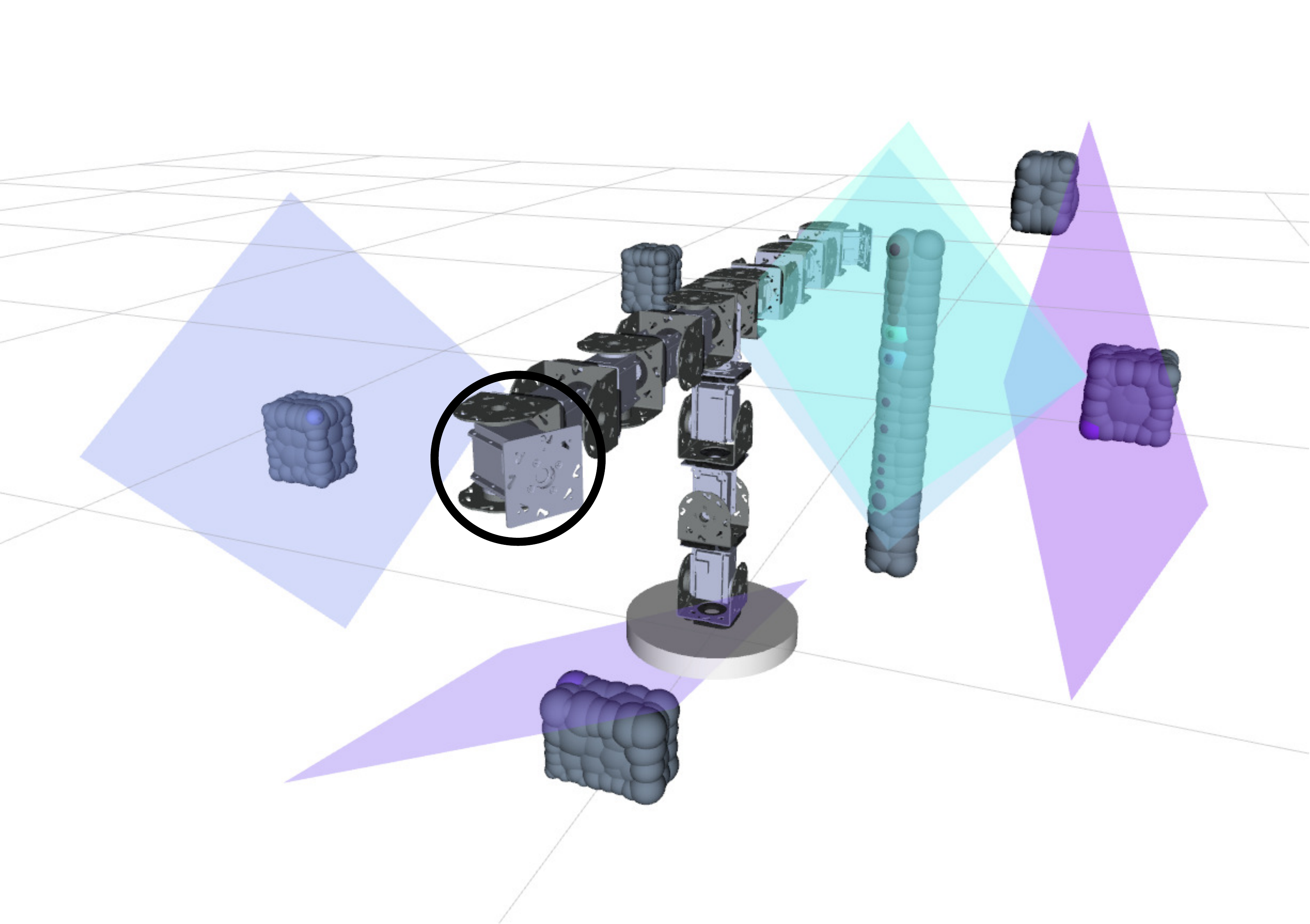}\label{fig:sphere-obstacles}}
  \caption{(a) A configuration built by 14 CKBot UBar modules is
    placed in a cluttered environment with 6 obstacles. (b) Apply the
    sphere-tree construction algorithm on all obstacles and the total
    number of obstacle spheres is 373. For the module inside the
    circle at its current state, only 5 obstacle spheres are necessary
    for collision avoidance and their obstacle planes are shown.}
  \label{fig:obstacle-simplify}
\end{figure}

In order to resolve the difficulty that there can be a large number of
obstacle spheres leading to a large number of constraints, we present
a novel way to significantly simplify these constraints as a
pre-processing step for optimization. As mentioned, for module $m_i$
and the $j$th obstacle sphere, we can compute an obstacle plane
$P_{ij}$ to build a linear constraint. If another obstacle sphere and
module $m_i$ are perfectly separated by $P_{ij}$, then this obstacle
sphere can be ignored for $m_i$ at the current planning step, because
as long as module $m_i$ never intersects with obstacle plane $P_{ij}$,
$m_i$ cannot touch this obstacle sphere. By this simple rule, we can
refine the set of obstacle spheres by iteratively applying an
erase-remove idiom technique efficiently for real-time
performance. For example, a robot configuration built by fourteen
CKBot UBar modules is placed in a cluttered environment where there
are six obstacles (Fig.~\ref{fig:raw-obstacle}). After applying the
sphere-tree construction algorithm, we derive 373 obstacle spheres in
total shown in Fig.~\ref{fig:sphere-obstacles}. For the module circled
in the figure, after refining the set of all obstacle spheres, we only
need to consider five obstacle spheres. Their obstacle planes are
shown in Fig.~\ref{fig:sphere-obstacles}. In the current step,
Obstacle 1 and Obstacle 6 are behind other obstacles so they can be
ignored. And for the rest of the obstacles, we only need five spheres
to approximate these obstacle-avoidance constraints.

\subsection{Integrated Control and Motion Planning}
\label{sec:control-planning-combine}

With the control law in Sec.~\ref{sec:control} and the motion
constraints in Sec.~\ref{sec:planning}, we can formalize the control
and motion planning problem for multiple kinematic chains
$G_K: \mathcal{W}\rightsquigarrow \mathcal{F}_i, i=1,2,\cdots,\alpha$ as the
following quadratic program with linear constraints
\begin{equation}
  \label{eq:qp-program-full}
  \begin{aligned}
    \mathrm{minimize}\quad &\frac{1}{2}\dot{\Theta}^\intercal \dot{\Theta}\\
    \mathrm{subject\ to}\quad &\mathbf{J}\mathbf{P} = \widetilde{\mathbf{V}} +
  \mathbf{K}(\widetilde{\mathbf{P}} - \mathbf{P})\\
    &\frac{\Theta_{\min} - \Theta}{\Delta t} \le \dot{\Theta} \le \frac{\Theta_{\max} - \Theta}{\Delta
    t}\\
    &\dot{\Theta}_{\min} \le \dot{\Theta} \le \dot{\Theta}_{\max}\\
    &(J_{\mathcal{M}_i}^s\dot{\Theta})^\wedge
    p_{\mathcal{M}_i} \bullet \hat{s}_{ij}\le d_{ij} - r_i\\ &\qquad\qquad\forall (\mathcal{M}_i, f_j)\in
    V_K\times F\\
    &(J_{\mathcal{M}_i}^s\dot{\Theta})^\wedge
    p_{\mathcal{M}_i}\bullet \tilde{s}_{ik} \le \|o_k^\prime - p_{\mathcal{M}_i}\| - r_i\\ &\qquad\qquad\forall (\mathcal{M}_i, S_k) \in
    V_K\times \mathbf{S}_i
  \end{aligned}
\end{equation}
in which $F$ is the set of all faces of the environment polyhedron and
$f_j$ is the $j$th face, $\mathbf{S}$ is the set of all spheres
approximating the environment obstacles,
$\mathbf{S}_i\subseteq \mathbf{S}$ is the current set of obstacle spheres
under consideration for module $m_i$, and $S_k$ is the $k$th sphere in
$\mathbf{S}_i$. By solving this quadratic program, the minimum norm
solution that satisfies the hardware limits, control requirement, and
motion constraints can be obtained for the current time step given the
current state of every kinematic chain
$G_K: \mathcal{W}\rightsquigarrow \mathcal{F}_i$ where
$i=1,2,\cdots,\alpha$, the desired velocity, and the position of the origin of
each frame $\mathcal{F}_i$.

This formulation can be used for motion tasks with simple constraints
(e.g., when obstacles are far from robots and motion goals). The
equality constraint enables the motion goal to be achieved very fast
with suitable gains. However, this can also cause difficulties for
optimization. For complicated scenarios, we propose a sequential
convex optimization formulation in which the objective function and
constraints are updated when encountering obstacles. Initially the
objective function is in the following form
\begin{equation}
  \label{eq:objective}
  f(\dot{\Theta}) = \|\dot{\Theta}\|^2 + \lambda\|\mathbf{JP}-(\widetilde{\mathbf{V}}+\mathbf{K}(\widetilde{\mathbf{P}}-\mathbf{P})\|^2
\end{equation}
in which $\lambda$ is a weight to address the significance of the feedback
controller. In order to avoid entering space that is hard to maintain
safety, we penalize aggressive motions towards obstacles by adding
$\|v_{\mathcal{M}_i}^s\bullet s_{ij}\|^2$ to the objective function when the
distance between module body frame $\mathcal{M}_i$ of the robot and
the $j$th obstacle is less than $d_{\min}$, and the objective function
becomes
\begin{equation}
  \label{eq:objective-penalty}
  f(\dot{\Theta}) = \|\dot{\Theta}\|^2 +
  \lambda\|\mathbf{JP}-(\widetilde{\mathbf{V}}+\mathbf{K}(\widetilde{\mathbf{P}}-\mathbf{P})\|^2
  + \mu_{ij}\|v_{\mathcal{M}_i}^s\bullet s_{ij}\|^2
\end{equation}
in which $\mu_{ij}$ is also a weight. The new penalizing term means that
we want this module $m_i$ to minimize its motion towards the $j$th
obstacle sphere. We have to check every module to update the objective
function and this can be computed easily by sphere-to-sphere distance.

If module $\mathcal{M}_i$ makes contact with the $j$th obstacle
sphere, this module will be forced to move away by defining a
repulsive velocity as the normal to the obstacle plane. This can be
done by adding a hard inequality constraint
\begin{equation}
  \label{eq:repulsive}
  v_{\mathcal{M}_i}^s\bullet s_{ij} = \left(J_{\mathcal{M}_i}^s\dot{\Theta}\right)^\wedge
  p_{\mathcal{M}_i}\bullet s_{ij} \le \gamma_{ij}
\end{equation}
in which $\gamma_{ij}$ is bounded between $-\|v_{{\mathcal{M}_i}}^s\|$ and
$0$. This constraint can force module body frame $\mathcal{M}_i$ to
move away from the $j$th obstacle sphere. Note that the previously
added penalizing term for this module
$\mu_{ij}\|v_{\mathcal{M}_i}^s\bullet s_{ij}\|^2$ is removed from the
objective function. The larger $\gamma_{ij}$ is, the faster the module
$\mathcal{M}_i$ moves away from the $j$th obstacle.

\subsection{Iterative Algorithm for Manipulation Planning}
\label{sec:algorithm}

The set of module graph $\mathbf{G}_m$ described in
Sec.~\ref{sec:configuration} and the twist set $\bm{\xi}$ described in
Sec.~\ref{sec:mod-kinematics} associated with all the joints in
different type of modules are computed and stored. For a modular robot
configuration $G$, assuming the base module $\bar{m}$ and how it is
attached to the world frame $\mathcal{W}$ as well as the motion goals
$\tilde{p}_{\mathcal{F}_1}$, $\tilde{p}_{\mathcal{F}_2}$, $\cdots$,
$\tilde{p}_{\mathcal{F}_\alpha}$ for frame $\mathcal{F}_1$,
$\mathcal{F}_2$, $\cdots$, $\mathcal{F}_\alpha$ respectively are known, the set
of all faces of the environment polyhedron is $F$ and the set of all
spheres approximating environmental obstacles is $\mathbf{S}$, the
full algorithm framework is shown in Algorithm~\ref{alg:framework}
with following functions:
\begin{itemize}
\item $\mathtt{BFS}(G, \mathbf{G}_m, \bar{m})$: Traverse a modular
  robotic configuration $G$ in breadth-first-search order starting
  from $\bar{m}$ to construct the kinematics graph $G_K = (V_K, E_K)$;
\item $\mathtt{GetChain}(G_K, \mathcal{F})$: Return the kinematic
  chain from $\mathcal{W}$ to $\mathcal{F}$ in $G_K$;
\item
  $\mathtt{SolveQP}(G_K, \mathcal{F}_1, \mathcal{F}_2, \cdots,
  \mathcal{F}_\alpha, \widetilde{\mathbf{P}}(t), \widetilde{\mathbf{V}}(t),
  \mathbf{P}, \mathbf{K}, \Delta t)$: Construct and try to solve the
  quadratic program described in
  Sec.~\ref{sec:control-planning-combine}. If failed to solve this
  program, then return \texttt{Null} as an invalid solution.
\end{itemize}

\begin{algorithm}[t]
  \caption{Control and Motion Planning}\label{alg:framework}
  \SetKwData{Result}{result}
  \KwIn{$\bm{\xi}$, $\mathbf{G}_m$, $\bar{m}$, $\mathcal{F}_1$,
    $\mathcal{F}_2$, $\cdots$, $\mathcal{F}_\alpha$, $\{\tilde{p}_{\mathcal{F}_i}(t)|
    0\le t\le T, i = 1,2,\cdots,\alpha\}$, $F$, $\mathbf{S}$}
  \KwOut{\Result}
  \SetKwFunction{SolveQP}{SolveQP}
  \SetKwFunction{BFS}{BFS}
  \SetKwFunction{GetChain}{GetChain}
  $G_K\leftarrow$ \BFS{$G, \mathbf{G}_m, \bar{m}$}\;
  $G_K: \mathcal{W}\rightsquigarrow \mathcal{F}_i\leftarrow$
  \GetChain{$G_K, \mathcal{F}_i$}, $i\in \left[1, \alpha\right]$\;
  Initialize $\Theta$\;
  Initialize $\mathbf{K}$ and $\Delta t$\;
  $t\leftarrow 0$\;
  \While{$\sum\limits_{i=1}^\alpha\| p_{\mathcal{F}_i} - \tilde{p}_{\mathcal{F}_i}(T)\|\ge \epsilon $}{
    Compute $\hat{s}_{ij}$ $\forall (\mathcal{M}_i, f_j)\in V_K\times F$\;
    Compute $\mathbf{S}_i\subseteq \mathbf{S}$ $\forall \mathcal{M}_i\in V_K$ and
    $\tilde{s}_{ik}$ $\forall S_k\in \mathbf{S}_i$\;
    $\dot{\Theta} \leftarrow$ \SolveQP{$G_K,
    \mathcal{F}_1, \mathcal{F}_2, \cdots, \mathcal{F}_\alpha,
    \tilde{\mathbf{P}}(t),
    \tilde{\mathbf{V}}(t), \mathbf{P}, \mathbf{K}, \Delta t$}\;
    \If{$\dot{\Theta}=\mathtt{Null}$}{
       \Return \Result $\leftarrow$ False;
    }
    Publish $\dot{\Theta}$ to the system\;
    $t\leftarrow t+\Delta t$;
  }
  \Return \Result $\leftarrow$ True;
\end{algorithm}

After initializing all the parameters, compute the unit direction
vector $\hat{s}_{ij}$ between every $\mathcal{M}_i\in V_K$ and every
face of the environment $f_j\in F$, compute
$\mathbf{S}_i\subseteq \mathbf{S}$ for every $\mathcal{M}_i\in V_K$ and the
corresponding unit direction vector $\tilde{s}_{ik}$ between
$\mathcal{M}_i$ and every obstacle sphere $S_k\in \mathbf{S}_i$. If
there is no valid solution, the program should stop, or the program
will continue until every $p_{\mathcal{F}_i}$ is close enough to the
destination $\tilde{p}_{\mathcal{F}_i}(T)$. If the trajectory
$\tilde{p}_{\mathcal{F}_i}(t)$ is not specified and only
$\tilde{p}_{\mathcal{F}_i}(T)$ where $T\to \infty$ is given, then this
algorithm can automatically find a trajectory for modules to navigate
the environment. The output from the planner can be applied directly
online (e.g., running on robot modules) to achieve real-time
performance, or can be integrated over time (one-step Euler
integration) to generate the trajectory for each module or joint.

\section{Experiments}
\label{sec:experiment}

We performed several experiments on two hardware platforms to verify
the approach. Here, we show that the framework is able to execute a
motion task with guaranteed control performance on real hardware while
satisfying all hardware constraints, frame boundary constraint, and
obstacle avoidance. Finally the framework is tested in a complex
scenario showing its ability for online trajectory optimization for
navigation tasks.

\subsection{Real-Time Control}
\label{sec:experiment-control}

\subsubsection{CKBot Chain}
\label{sec:ckbot-experiment}

\begin{figure}[t!]
  \centering
  \subfloat[]{\includegraphics[width=0.20\textwidth]{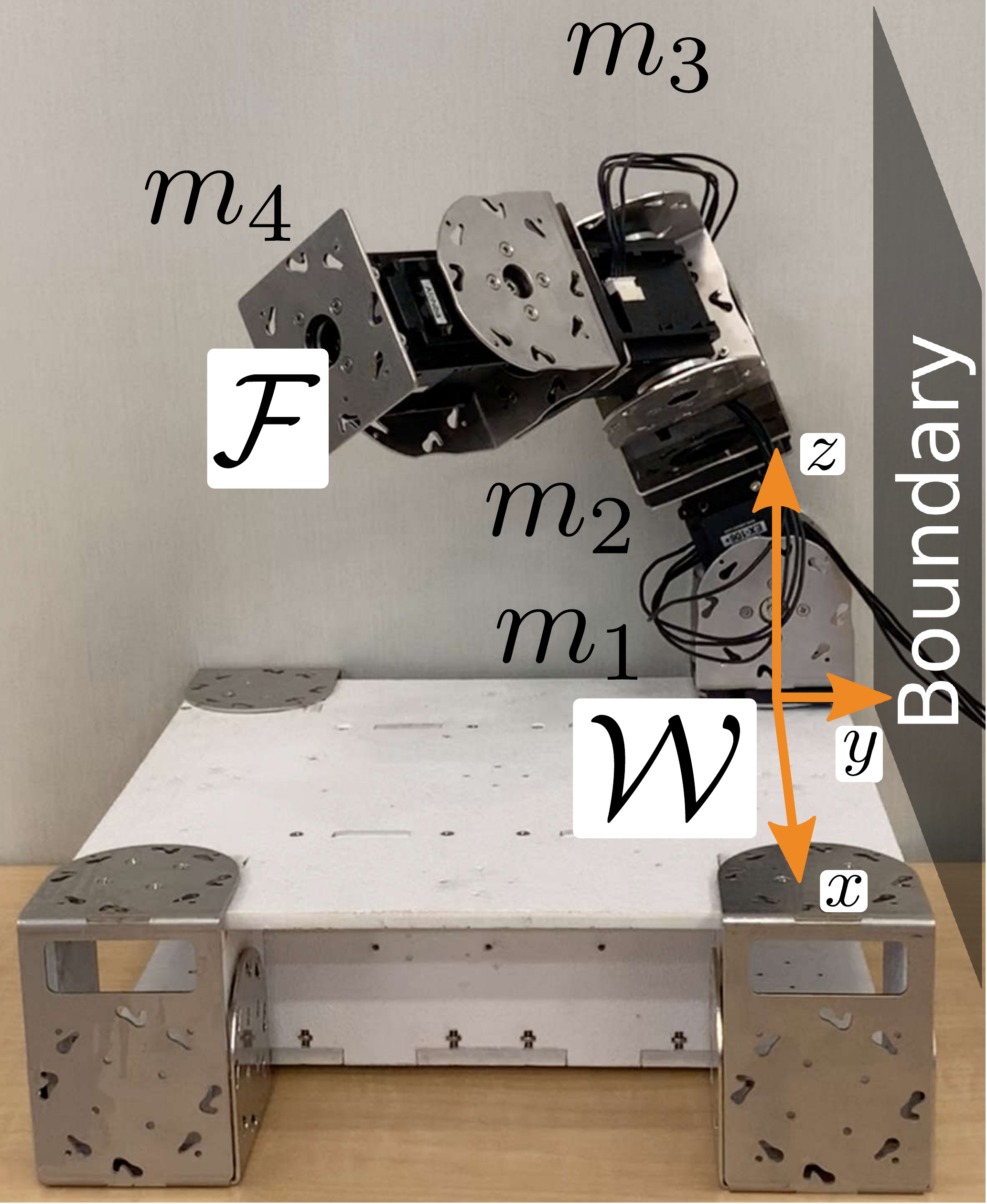}\label{fig:work-1}}
  \hfil
  \subfloat[]{\includegraphics[width=0.20\textwidth]{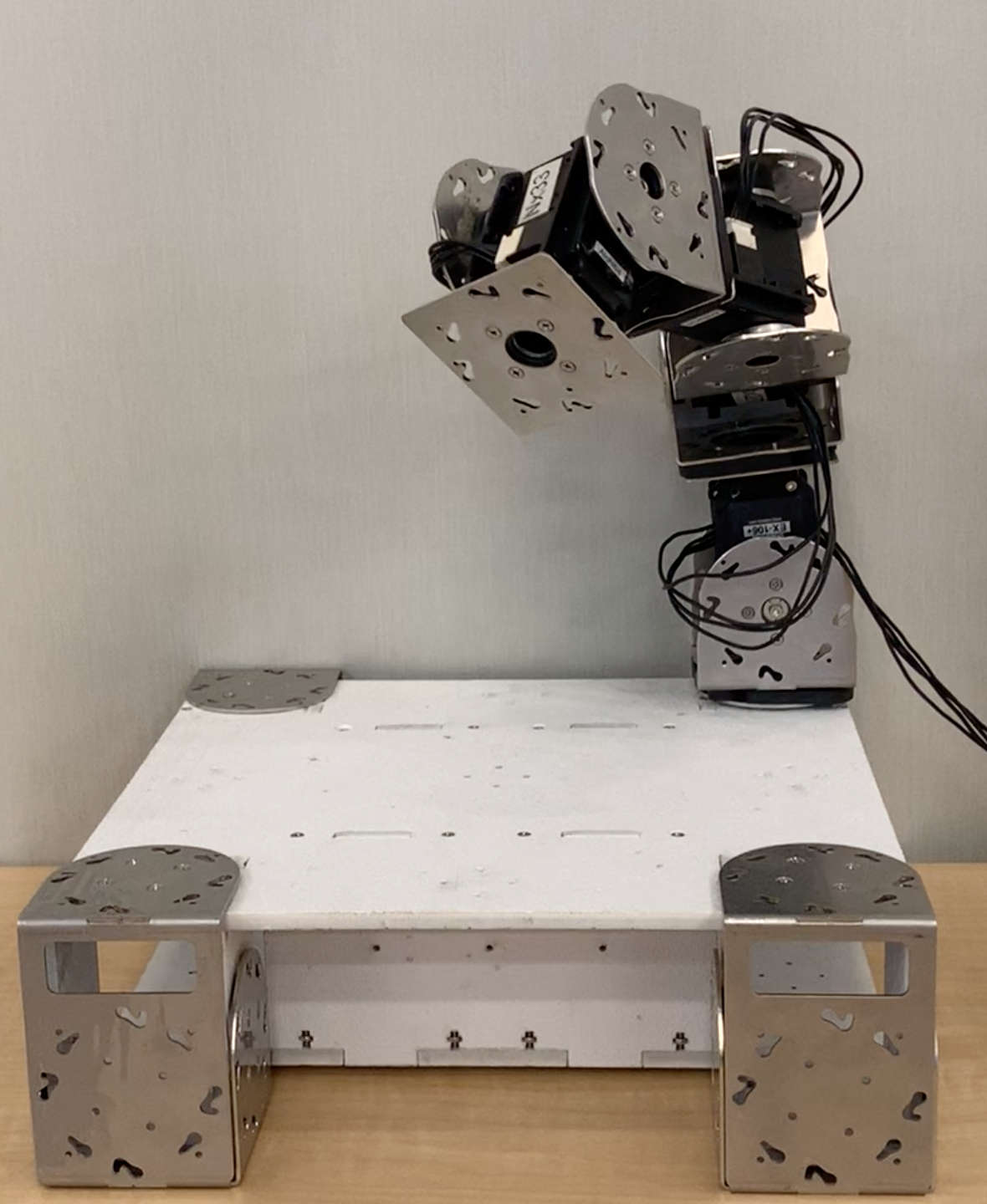}\label{fig:work-2}}
  \hfil
  \subfloat[]{\includegraphics[width=0.20\textwidth]{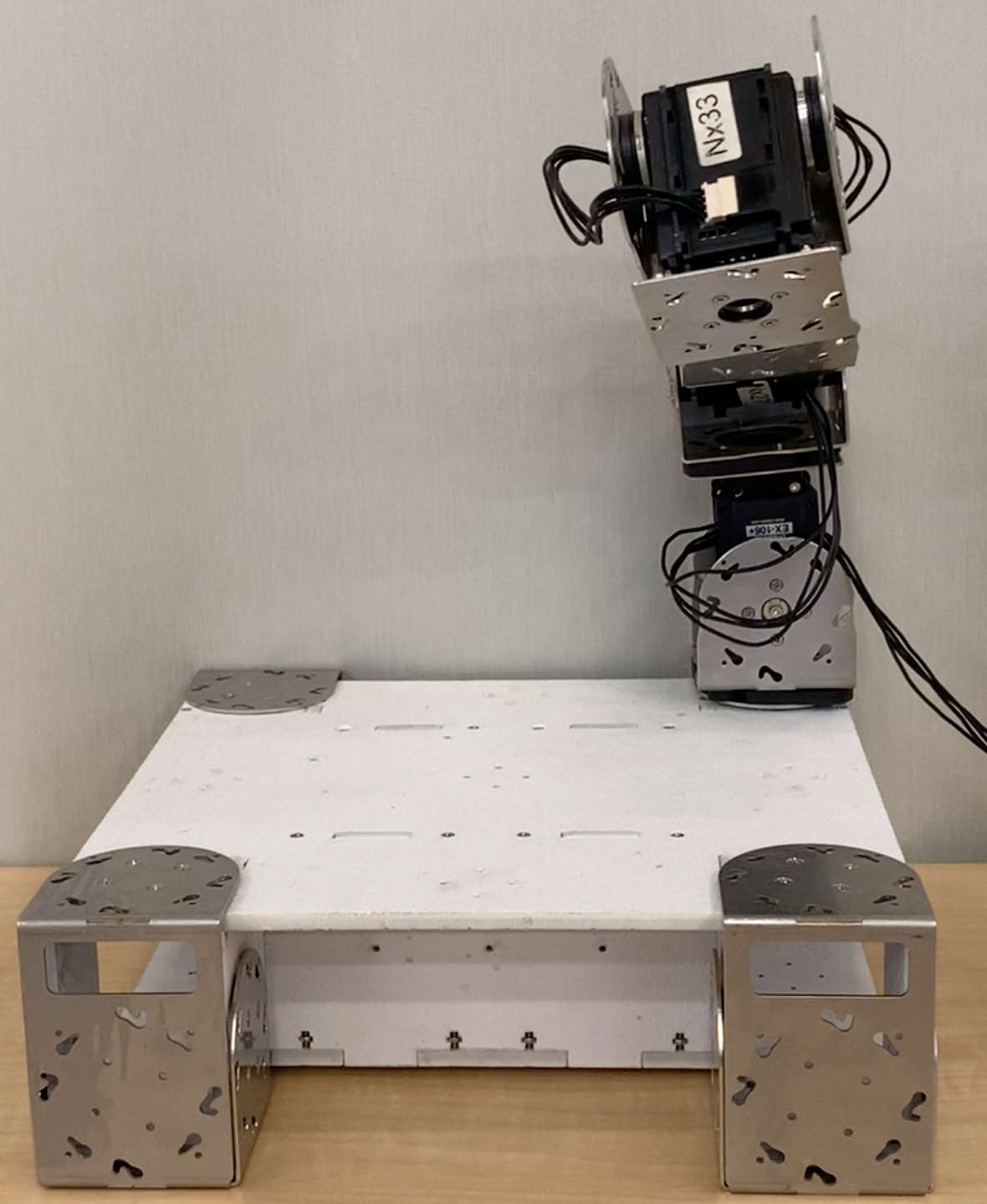}\label{fig:work-3}}
  \hfil
  \subfloat[]{\includegraphics[width=0.20\textwidth]{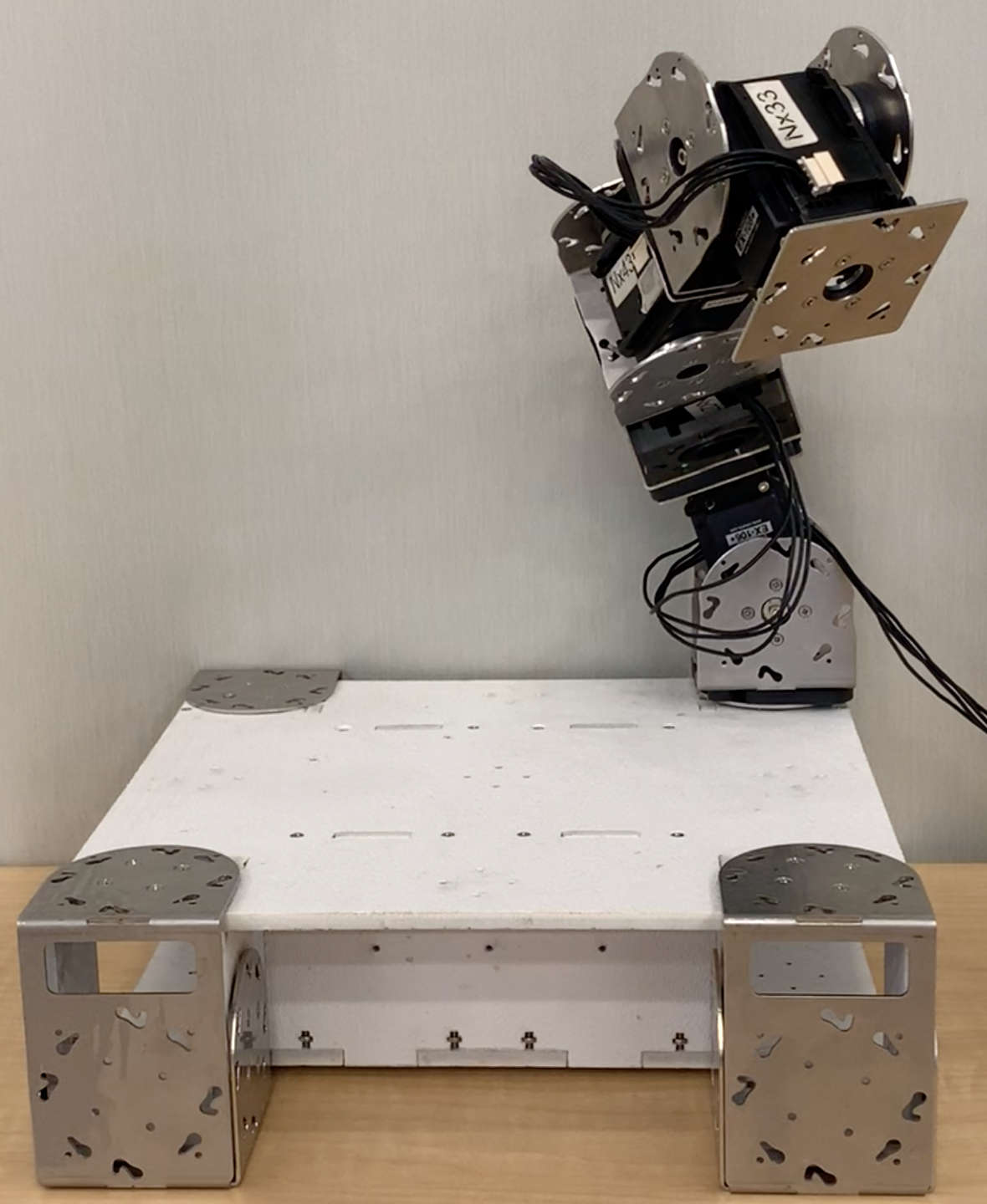}\label{fig:work-4}}
  \caption{Control $p_{\mathcal{F}}$ to follow a given trajectory
    along $+y$-axis of $\mathcal{W}$ by \SI{15}{cm} from the initial
    pose (a) to the final pose (d). All the modules have to be on the
    left side of the boundary. $m_1$, $m_2$, and $m_3$ have to
    approach the boundary first (b) and then move away from the
    boundary (c) to finish the task.}
\end{figure}

\begin{figure}[t!]
  \centering
  \subfloat[]{\includegraphics[width=0.20\textwidth]{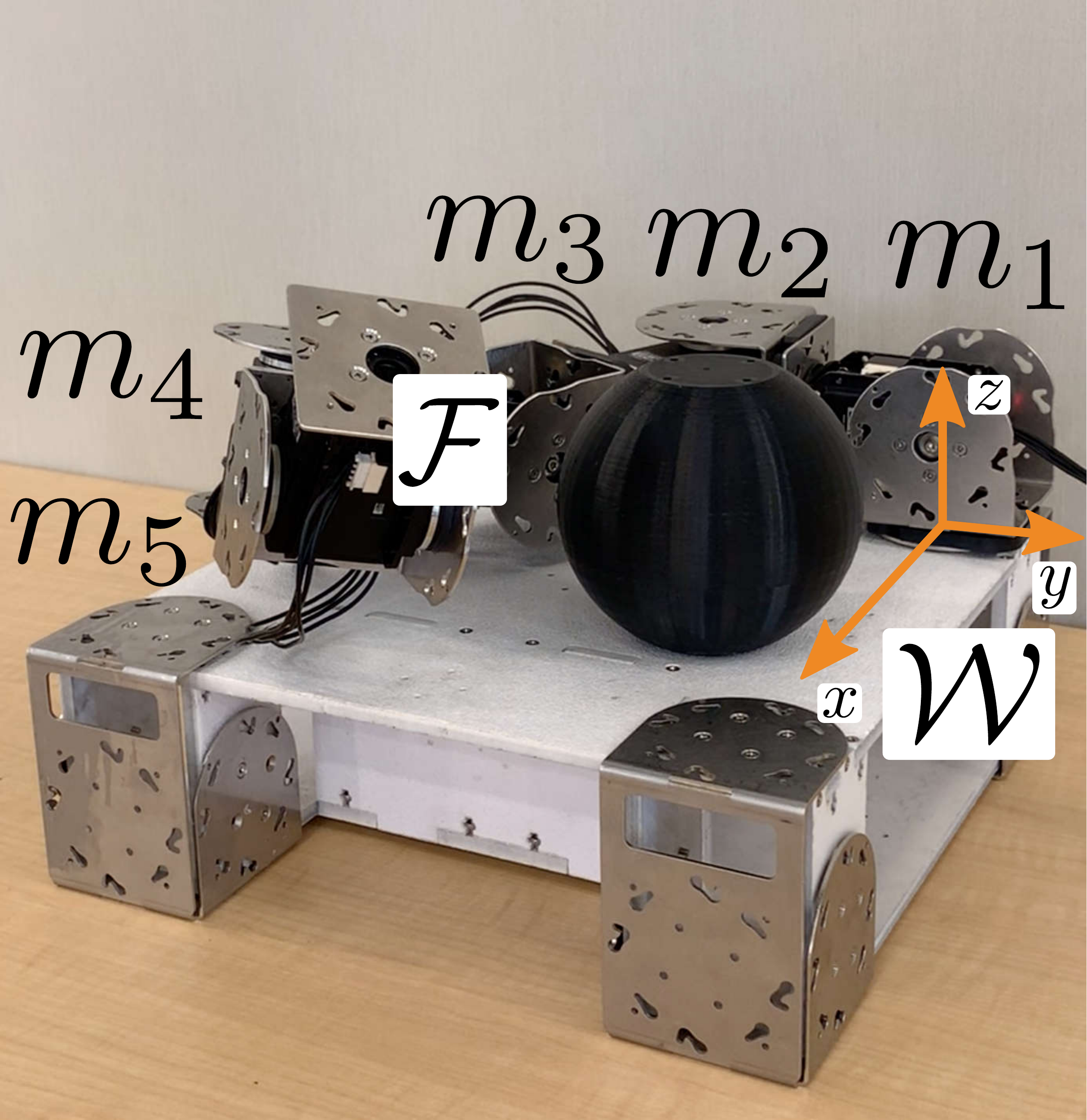}\label{fig:traj-1}}
  \hfil
  \subfloat[]{\includegraphics[width=0.20\textwidth]{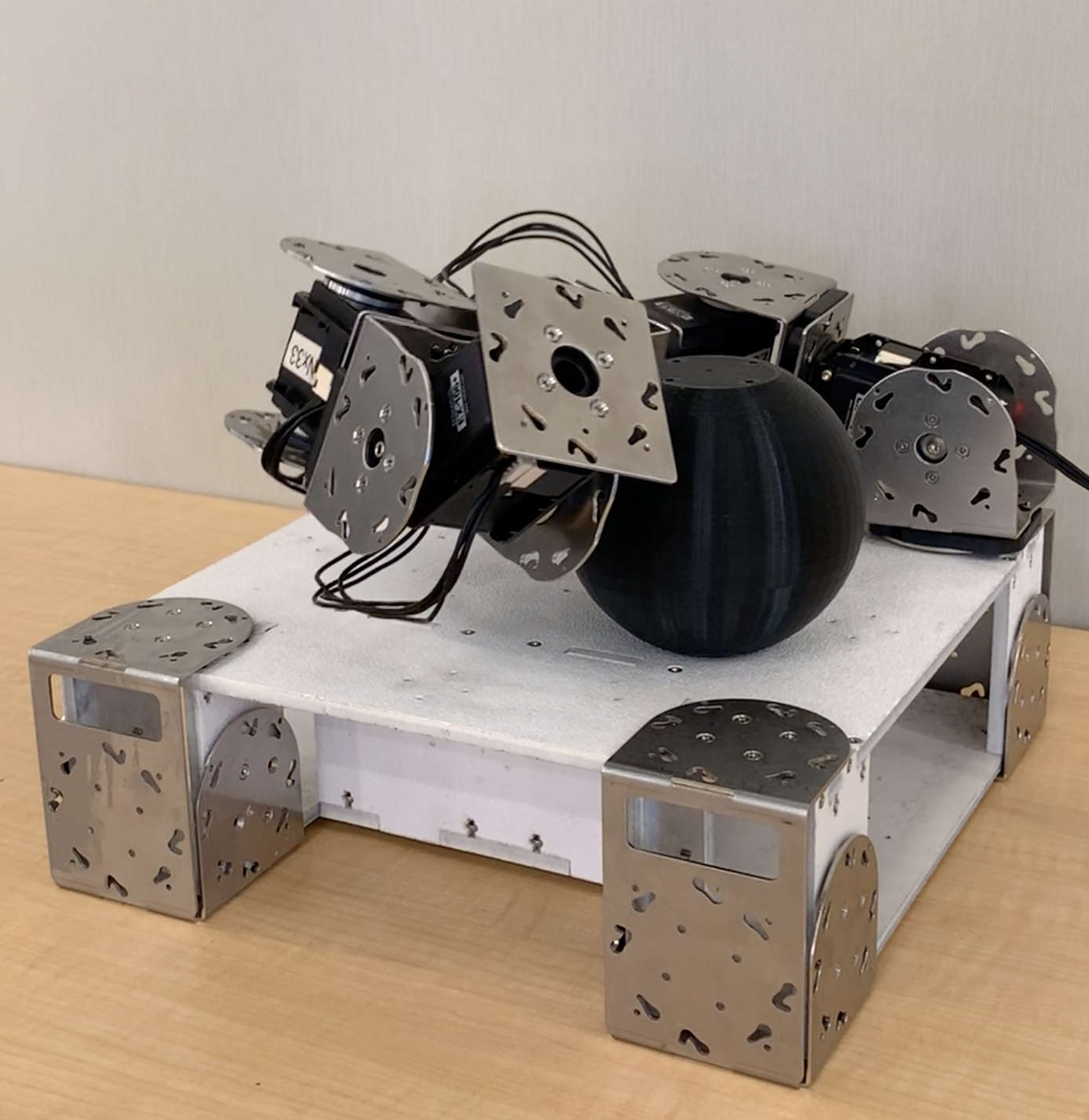}\label{fig:traj-2}}
  \hfil
  \subfloat[]{\includegraphics[width=0.23\textwidth]{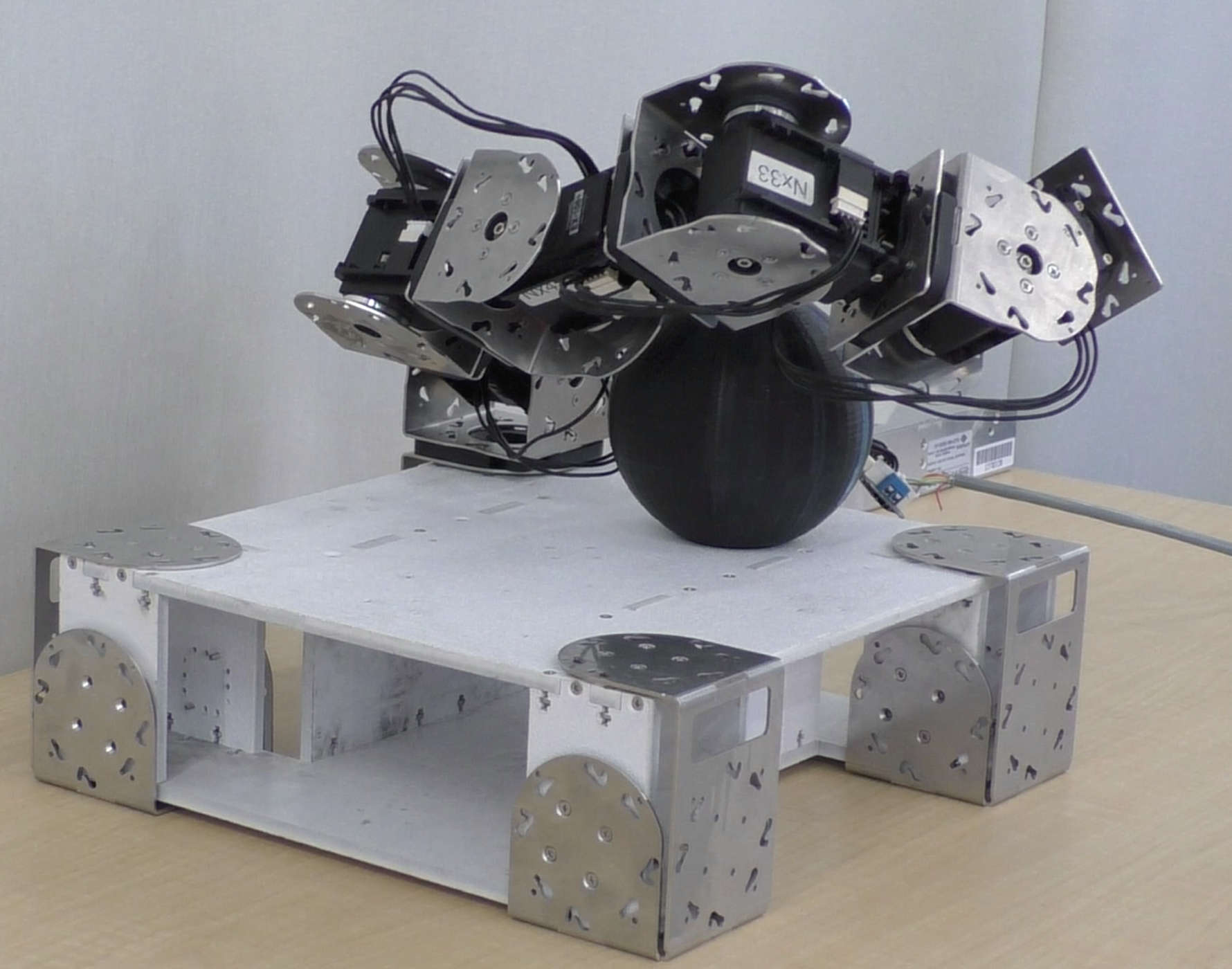}\label{fig:traj-3}}
  \hfil
  \subfloat[]{\includegraphics[width=0.23\textwidth]{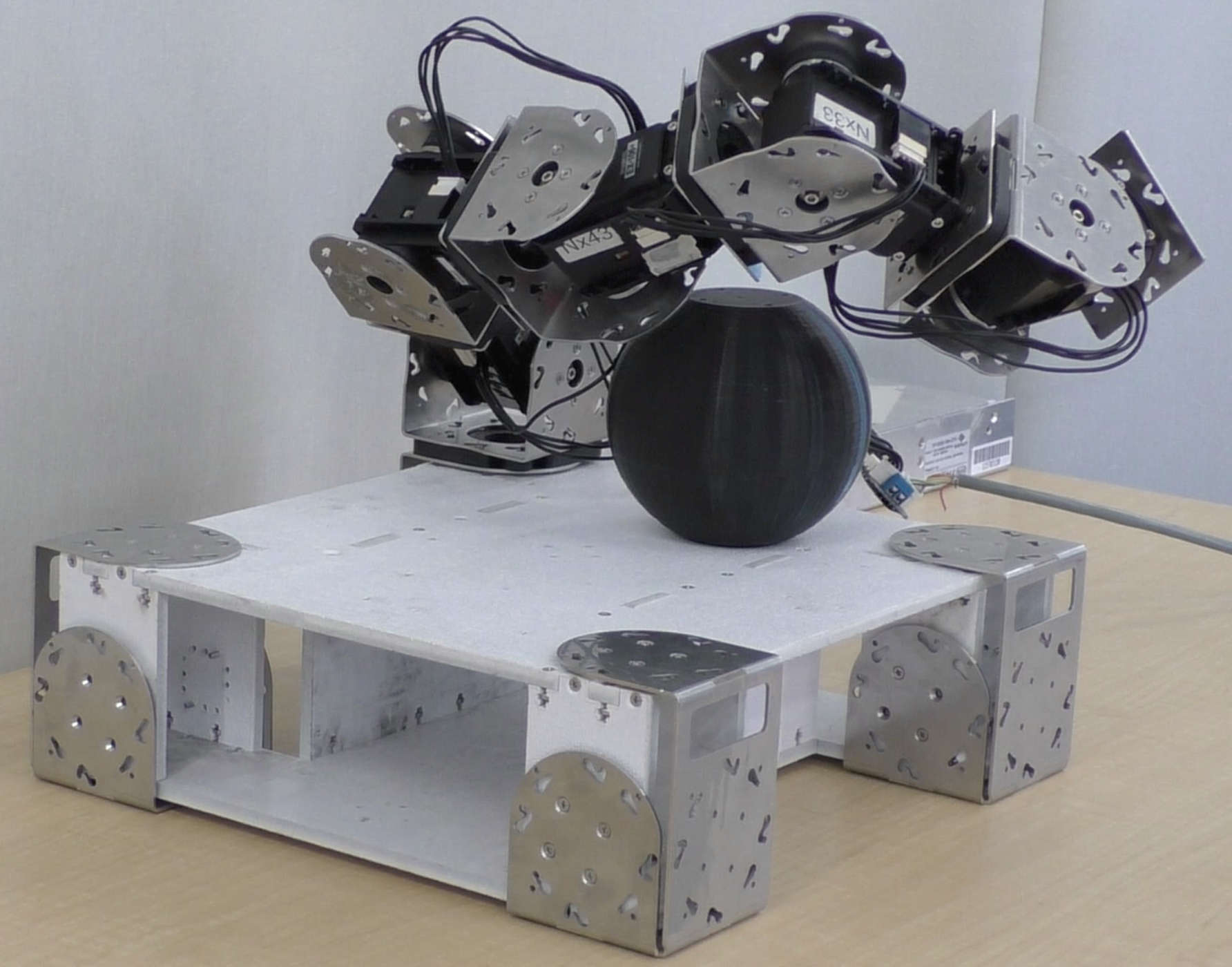}\label{fig:traj-4}}
  \caption{Control $p_{\mathcal{F}}$ from its initial pose (a) to its
    final pose (d) by both following a given trajectory along
    $+y$-axis of $\mathcal{W}$ by \SI{15}{cm} and navigating to the
    destination directly. The modules have to move around the sphere
    obstacle while executing these two tasks.}
\end{figure}

\begin{figure}[t!]
  \centering
  \subfloat[]{\includegraphics[width=0.35\textwidth]{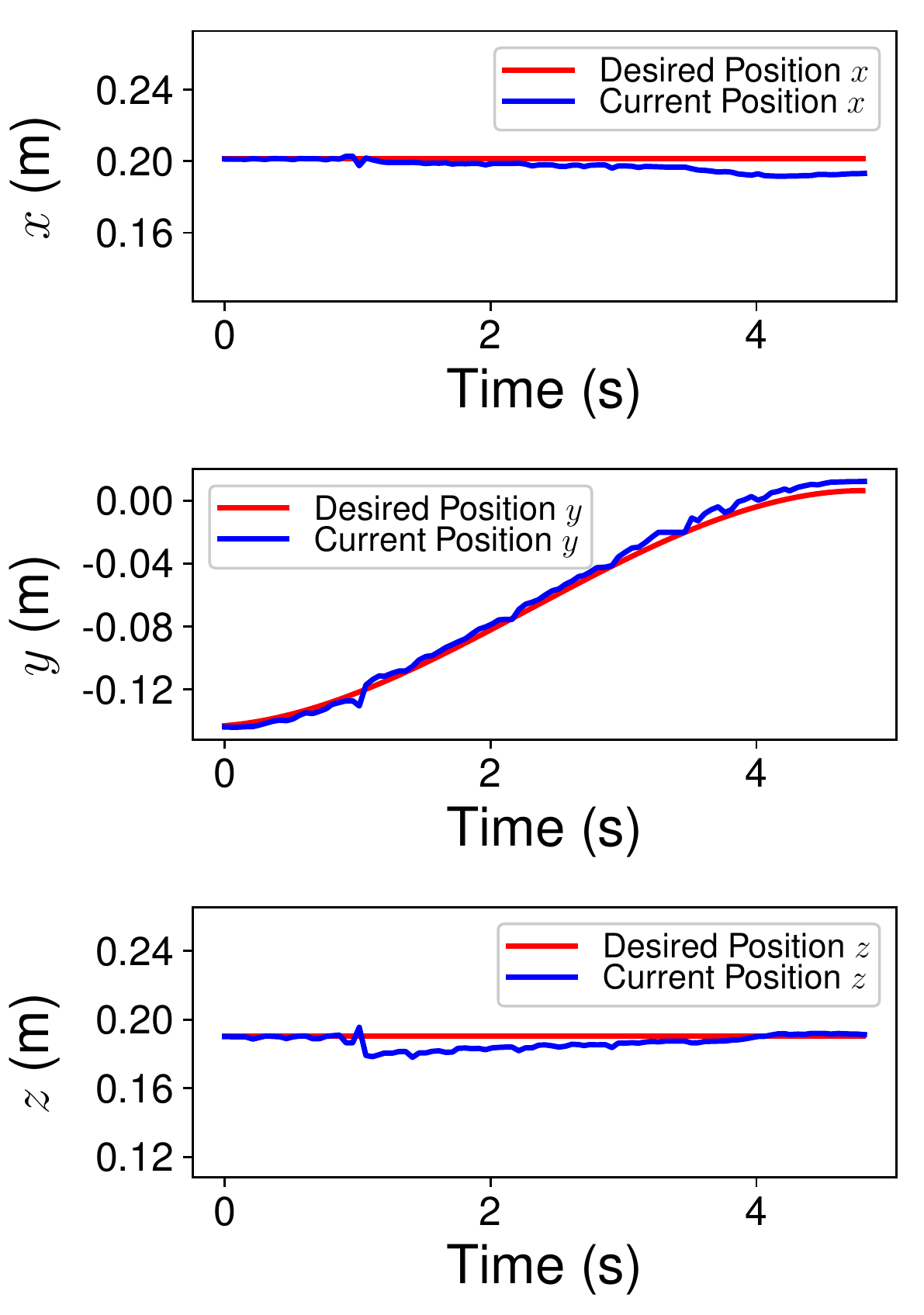}\label{fig:pos-boundary}}
  \hfil
  \subfloat[]{\includegraphics[width=0.35\textwidth]{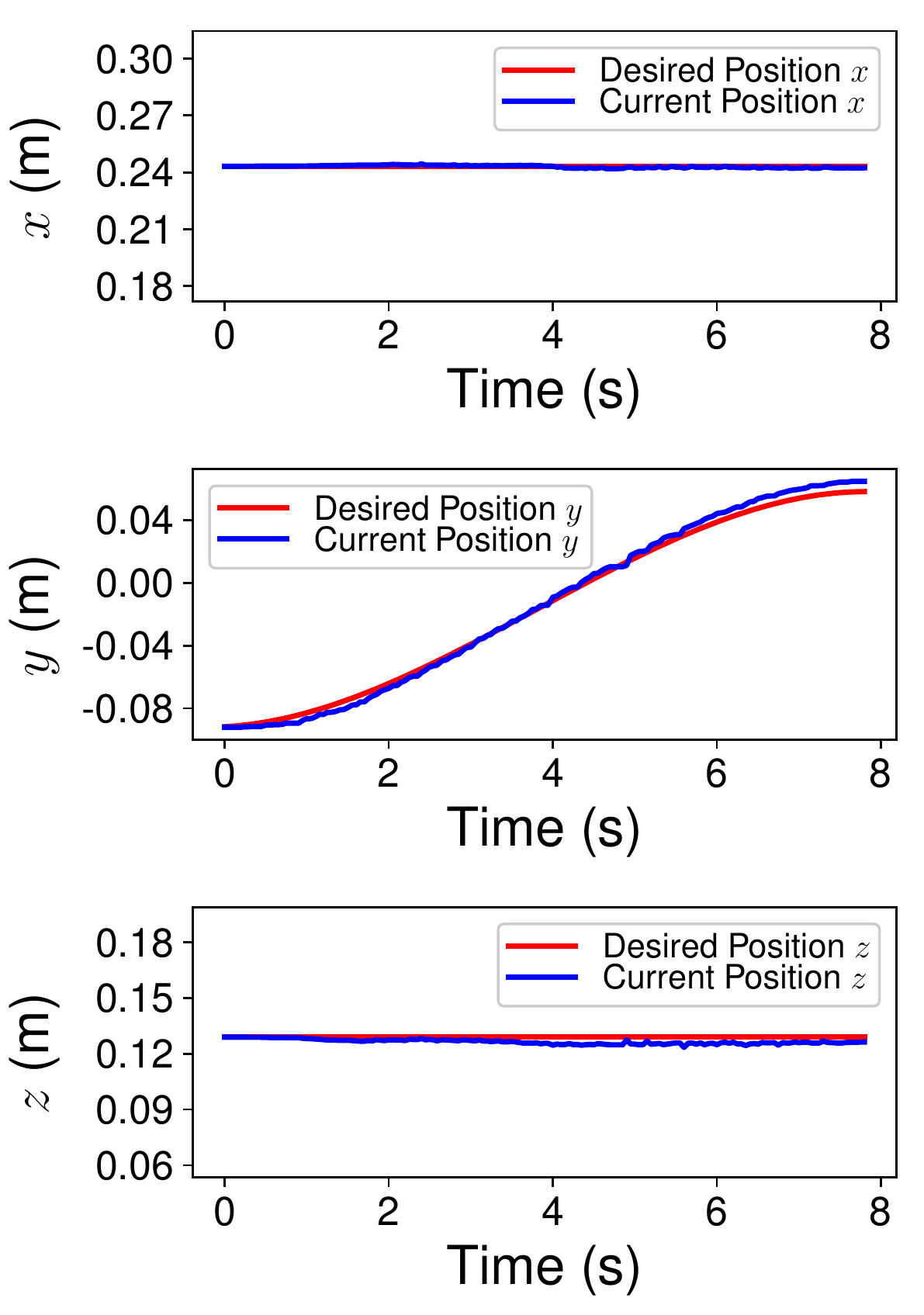}\label{fig:pos-traj}}
  \caption{The motion of $p_{\mathcal{F}}$: (a) the four-module task;
    (b) the five-module trajectory following task.}
  \label{fig:traj-data}
\end{figure}

\begin{figure}[t!]
  \centering
  \subfloat[]{\includegraphics[width=0.35\textwidth]{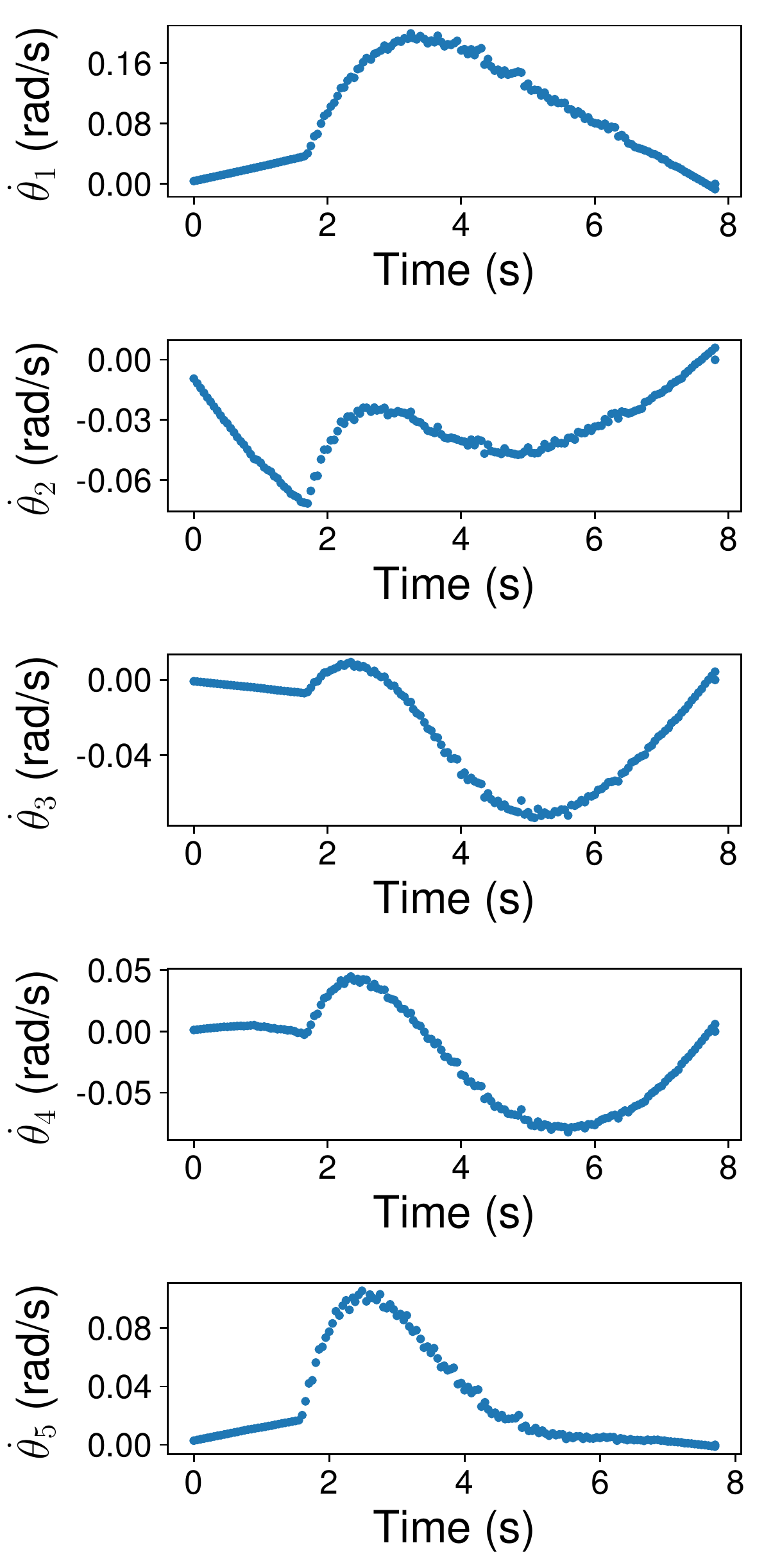}}
  \hfil
  \subfloat[]{\includegraphics[width=0.35\textwidth]{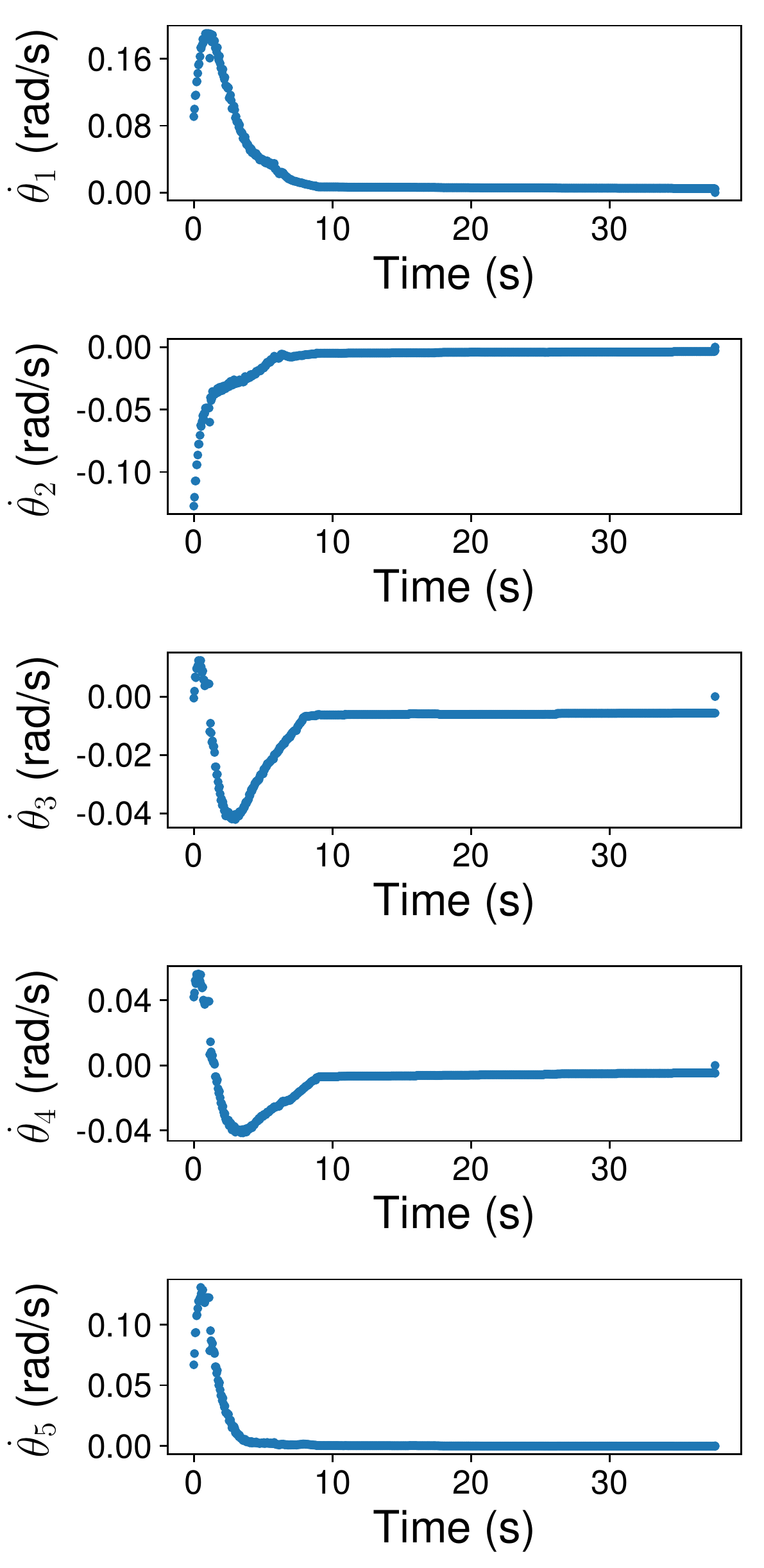}\label{fig:vel-navigate}}
  \caption{The control input $\dot{\Theta}$ for the five-module chain
    experiment: (a) the trajectory following task; (b) the destination
    navigation task.}
  \label{fig:vel-data}
\end{figure}

\begin{figure}[t!]
  \centering
  \subfloat[]{\includegraphics[width=0.35\textwidth]{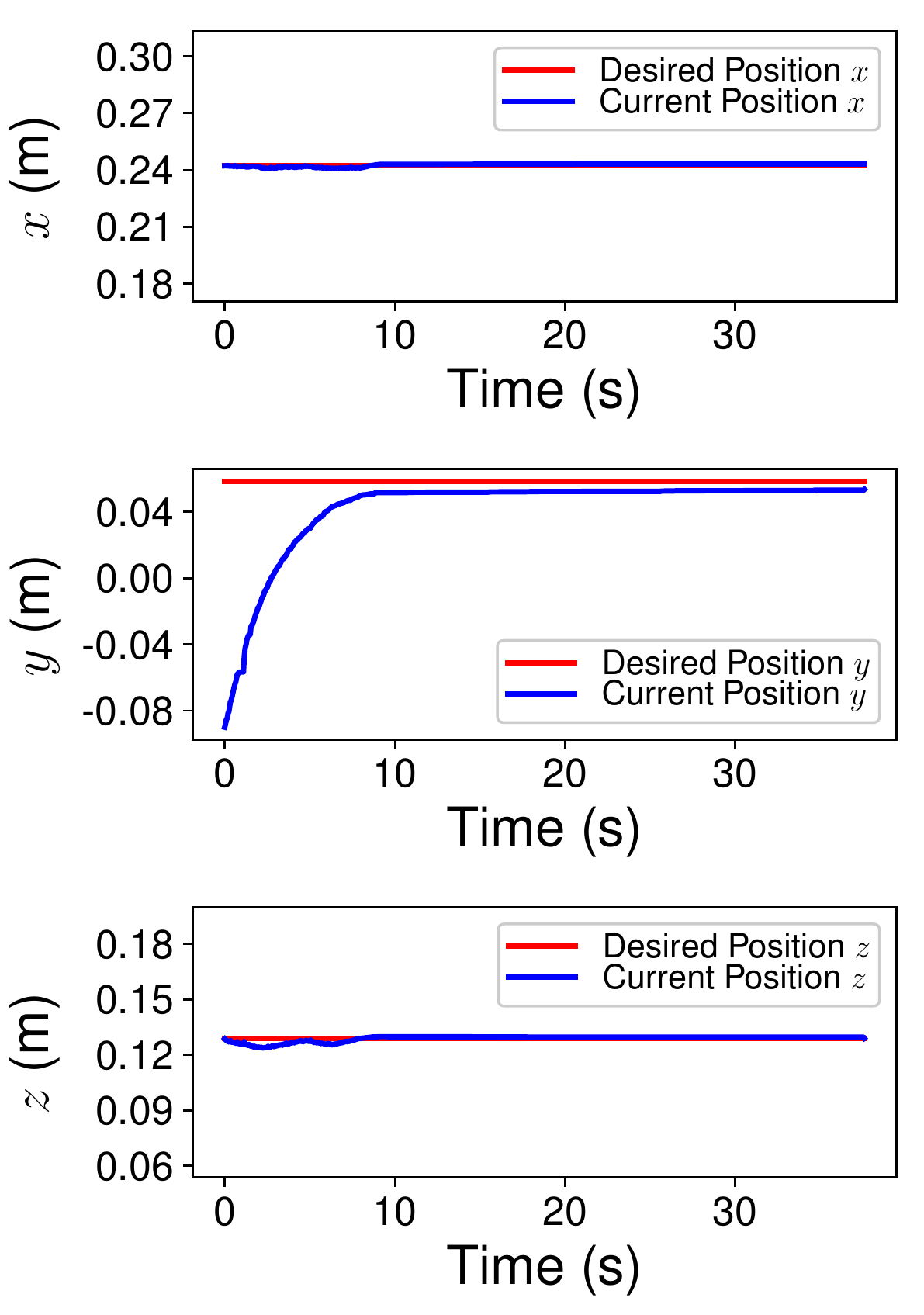}\label{fig:pos-navigate}}
  \hfil
  \subfloat[]{\includegraphics[width=0.35\textwidth]{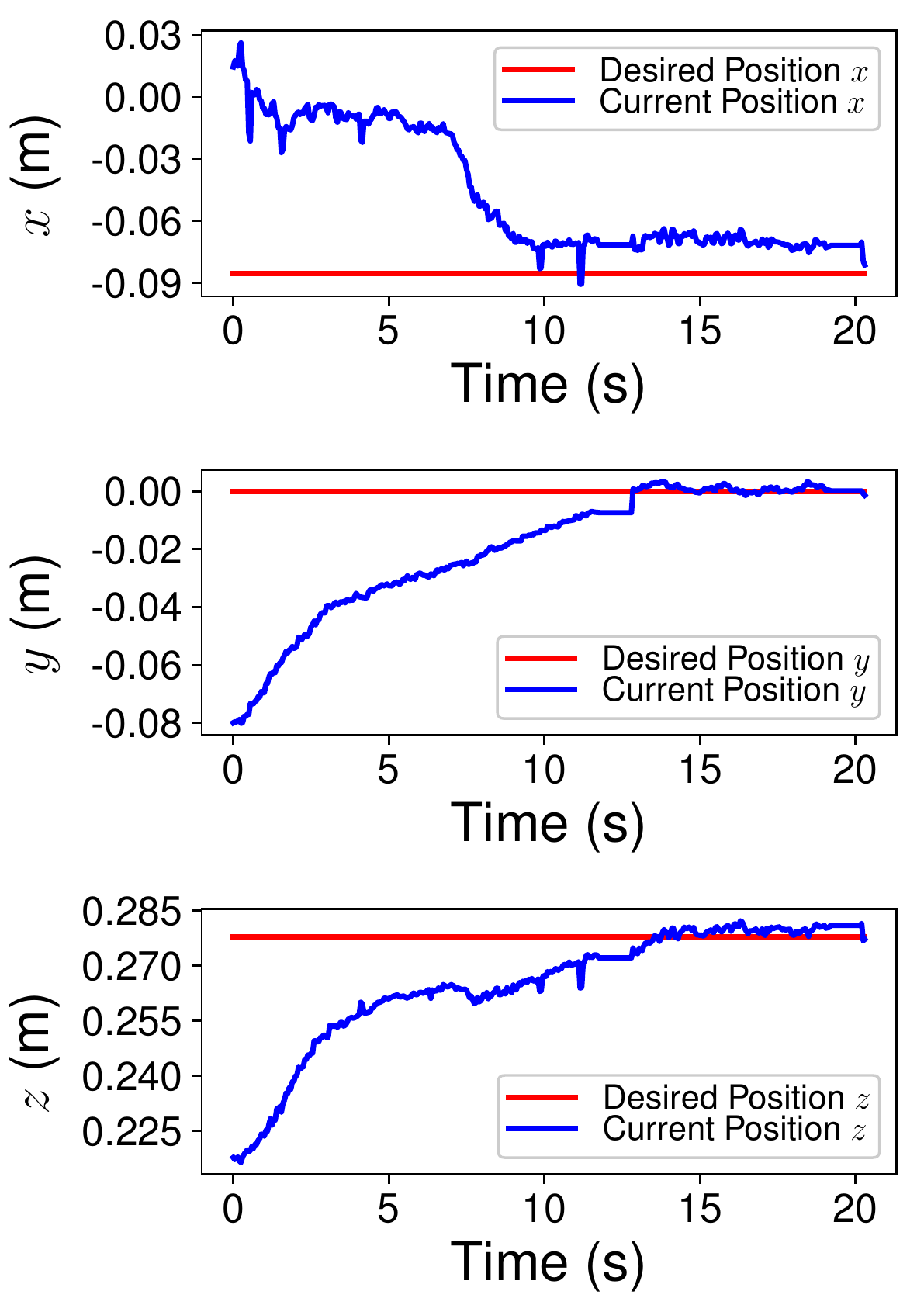}\label{fig:smores-pos}}
  \caption{The motion of $p_{\mathcal{F}}$: (a) the CKBot five-module
    destination navigation task; (b) the SMORES-EP four-module chain
    destination navigation task.}
\end{figure}

A configuration with four CKBot UBar modules is shown in
Fig.~\ref{fig:work-1}. The base module $\bar{m} = m_1$ is attached to
the world frame $\mathcal{W}$ and frame $\mathcal{F}$ is attached to
connector $\mathcal{T}$ of module $m_4$. A virtual frame boundary is
next to the right side of the base. The task is to control
$p_\mathcal{F}$ to follow a given trajectory to the position shown in
Fig.~\ref{fig:work-4}. Another experiment setup with five CKBot UBar
modules is shown in Fig.~\ref{fig:traj-1}. The black sphere is an
obstacle, the base module $\bar{m} = m_1$ and frame $\mathcal{F}$ is
attached to connector $\mathcal{T}$ of module $m_5$. Two tasks are
executed: control $p_{\mathcal{F}}$ to follow a given trajectory and
control $p_{\mathcal{F}}$ to approach a specified destination with the
final position of $p_{\mathcal{F}}$ as shown in
Fig.~\ref{fig:traj-4}. The control loop runs at \SI{20}{Hz} with gain
$K = \mathrm{diag}(1,1,1)$. Fig.~\ref{fig:traj-data} and
Fig.~\ref{fig:pos-navigate} shows $p_{\mathcal{F}}(t)$ and
$\tilde{p}_{\mathcal{F}}$ of these three tests demonstrating tracking
and navigation performance. The velocity commands for all modules in
these two five-module demonstrations are shown in
Fig.~\ref{fig:vel-data} and all commands are within the constraints of
each module. Modules move more aggressively at the beginning when
executing the destination navigation task in order to quickly approach
the destination.

\subsubsection{SMORES-EP Chain}
\label{sec:smores-experiment}

\begin{figure}[t!]
  \centering
  \subfloat[]{\includegraphics[width=0.19\textwidth]{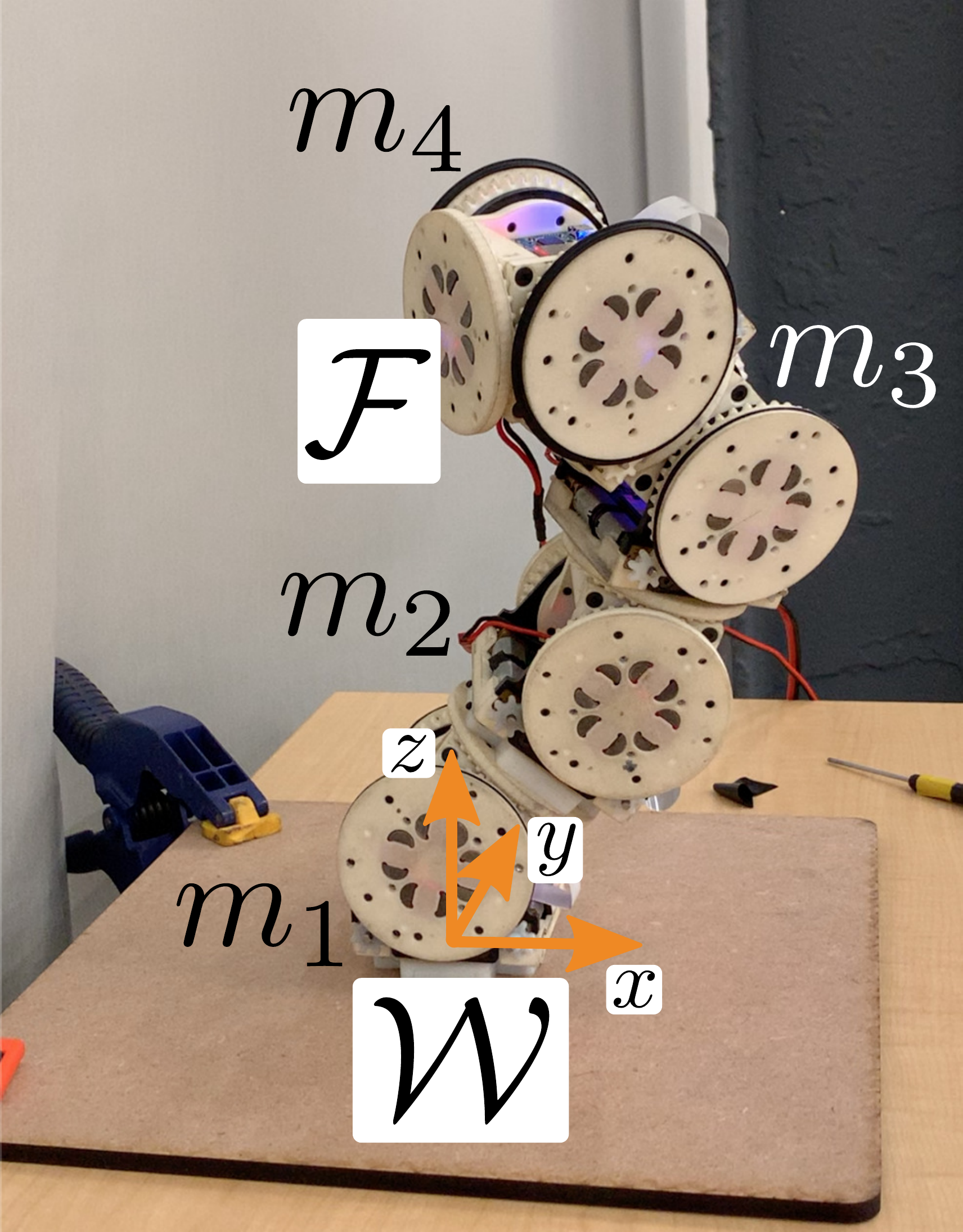}\label{fig:smores-init}}
  \hfil
  \subfloat[]{\includegraphics[width=0.19\textwidth]{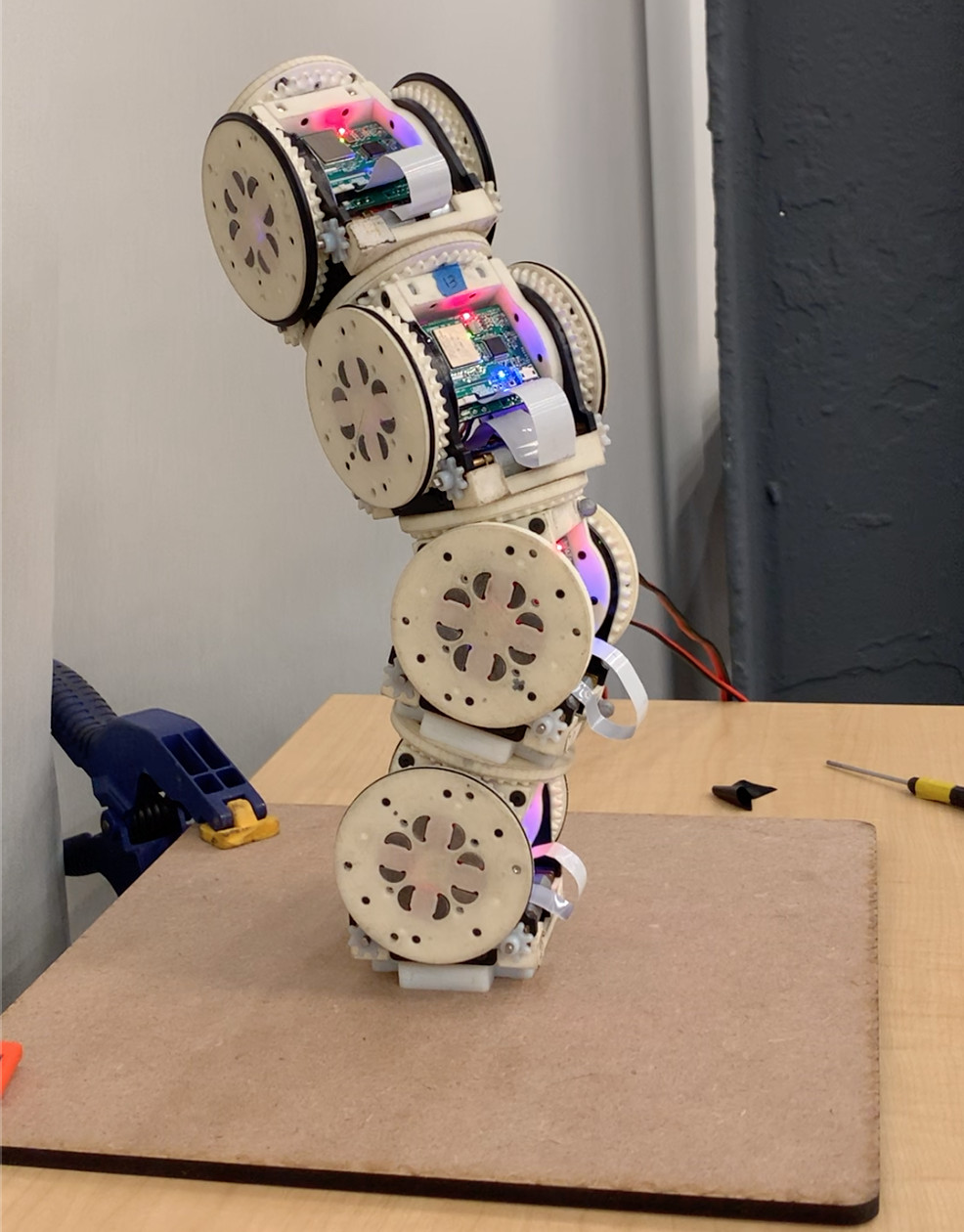}\label{fig:smores-goal}}
  \caption{Control a chain of SMORES-EP modules to navigate from its initial
    position (a) to a goal position (b). This chain is constructed by four
    modules with 16 DoFs.}
  \label{fig:smores-experiment}
\end{figure}

The experiment setup with four SMORES-EP modules is shown in
Fig.~\ref{fig:smores-init}. The base module $\bar{m} = m_1$ is fixed
to the world frame $\mathcal{W}$ and frame $\mathcal{F}$ is attached
to connector $\mathcal{T}$ of module $m_4$. This system has 16 DoFs
and the task is to control $p_{\mathcal{F}}$ to navigate to a
specified destination shown in Fig.~\ref{fig:smores-goal}. The control
loop runs at \SI{20}{Hz} and the gain
$K = \mathrm{diag}(0.5, 0.5, 0.5)$. The experiment result
$p_{\mathcal{F}}(t)$ is shown in Fig.~\ref{fig:smores-pos}. The
position sensors installed in SMORES-EP modules are customizable
potentiometers using
paints~\cite{Tosun-paintpot-icra-2017,Liu-smores-sensor-jmr-2021}. These
low-cost sensors with a modified Kalman filter for nonlinear systems
are used to provide position information of each DoF. Due to the
limitations of the hardware, some noise is evident.

\begin{figure}[t!]
  \centering
  \subfloat[]{\includegraphics[width=0.22\textwidth]{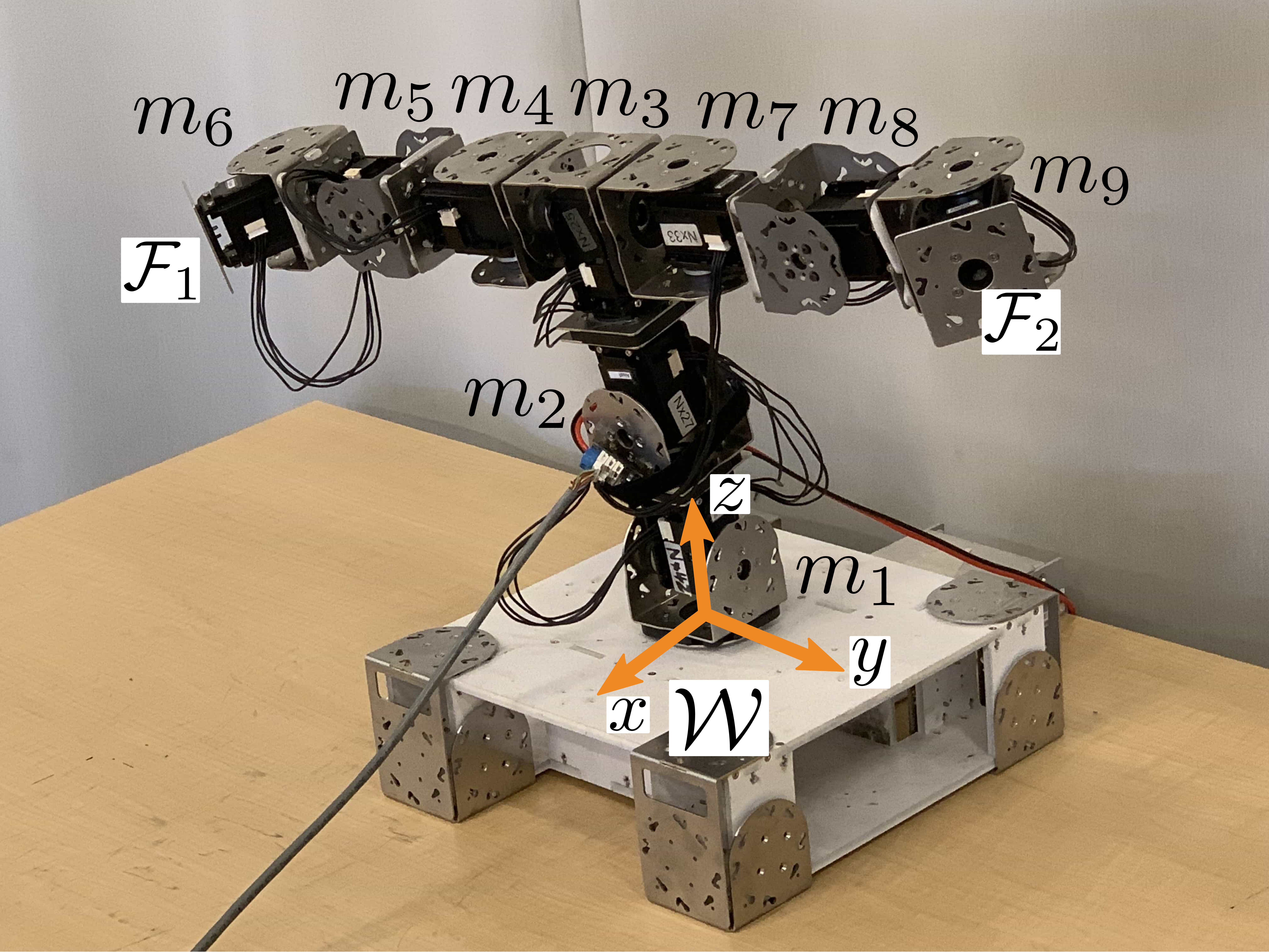}\label{fig:branch-init}}
  \hfil
  \subfloat[]{\includegraphics[width=0.22\textwidth]{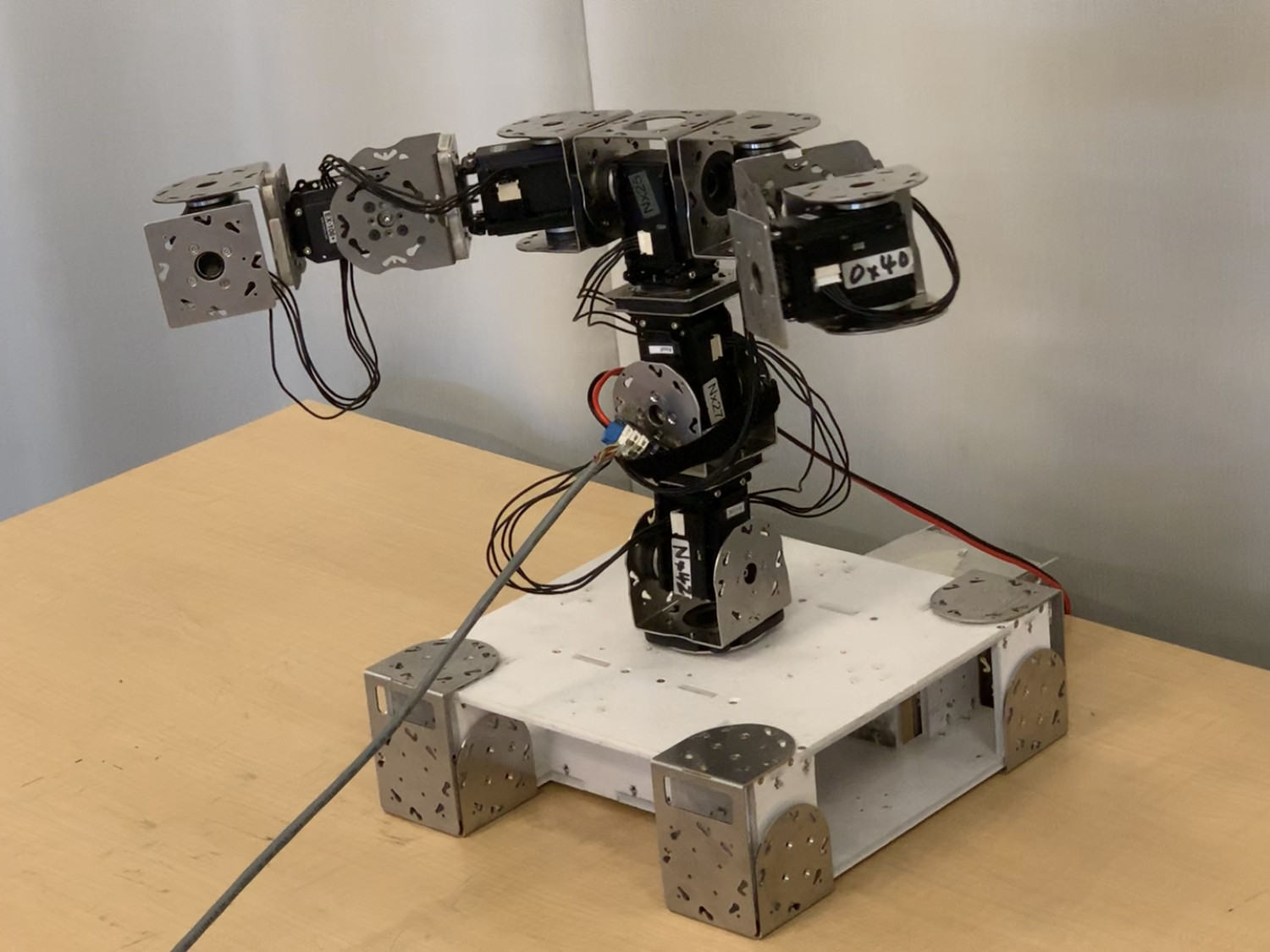}\label{fig:branch-2}}
  \hfil
  \subfloat[]{\includegraphics[width=0.22\textwidth]{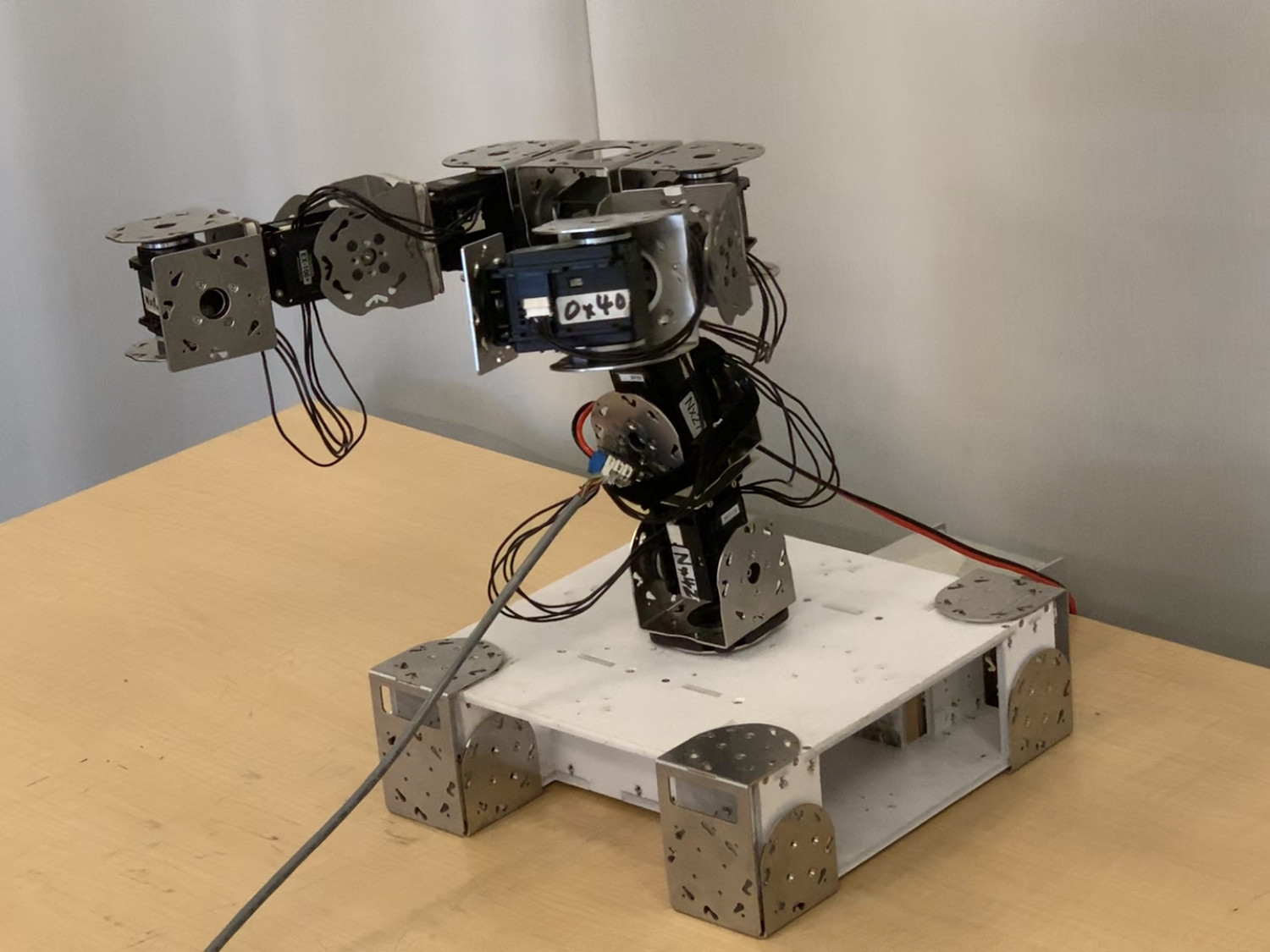}\label{fig:branch-3}}
  \hfil
  \subfloat[]{\includegraphics[width=0.22\textwidth]{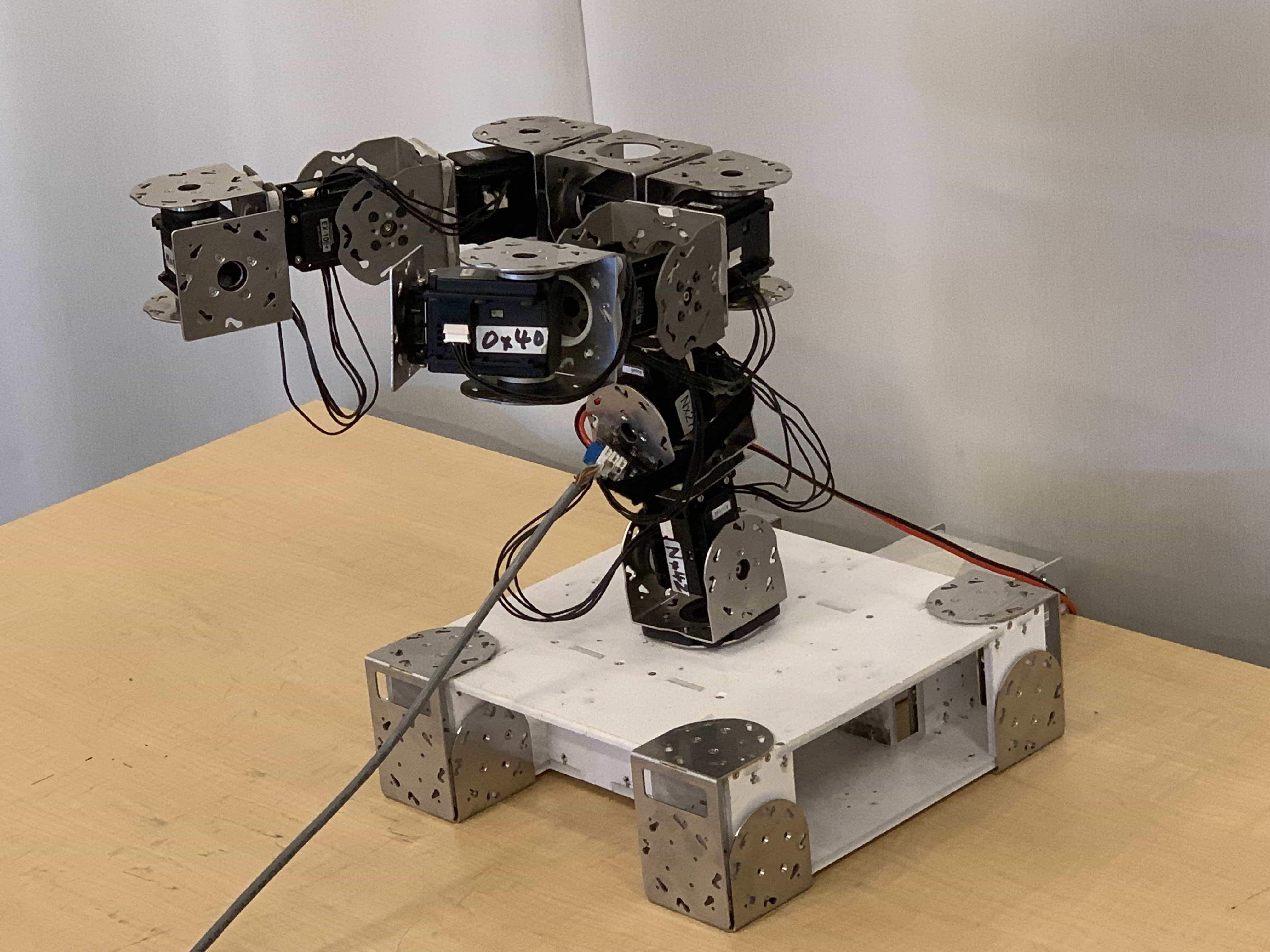}\label{fig:branch-goal}}
  \caption{Control $p_{\mathcal{F}_1}$ and $p_{\mathcal{F}_2}$ to
    follow two given trajectories respectively from their initial
    poses (a) to their final poses (d). Module $m_1$, $m_2$, and $m_3$
    initially have to move backward (b) and then move forward (c) in
    order to control $p_{\mathcal{F}_1}$ and $p_{\mathcal{F}_2}$ to
    follow their trajectories.}
\end{figure}

\begin{figure}[t!]
  \centering
  \subfloat[]{\includegraphics[width=0.35\textwidth]{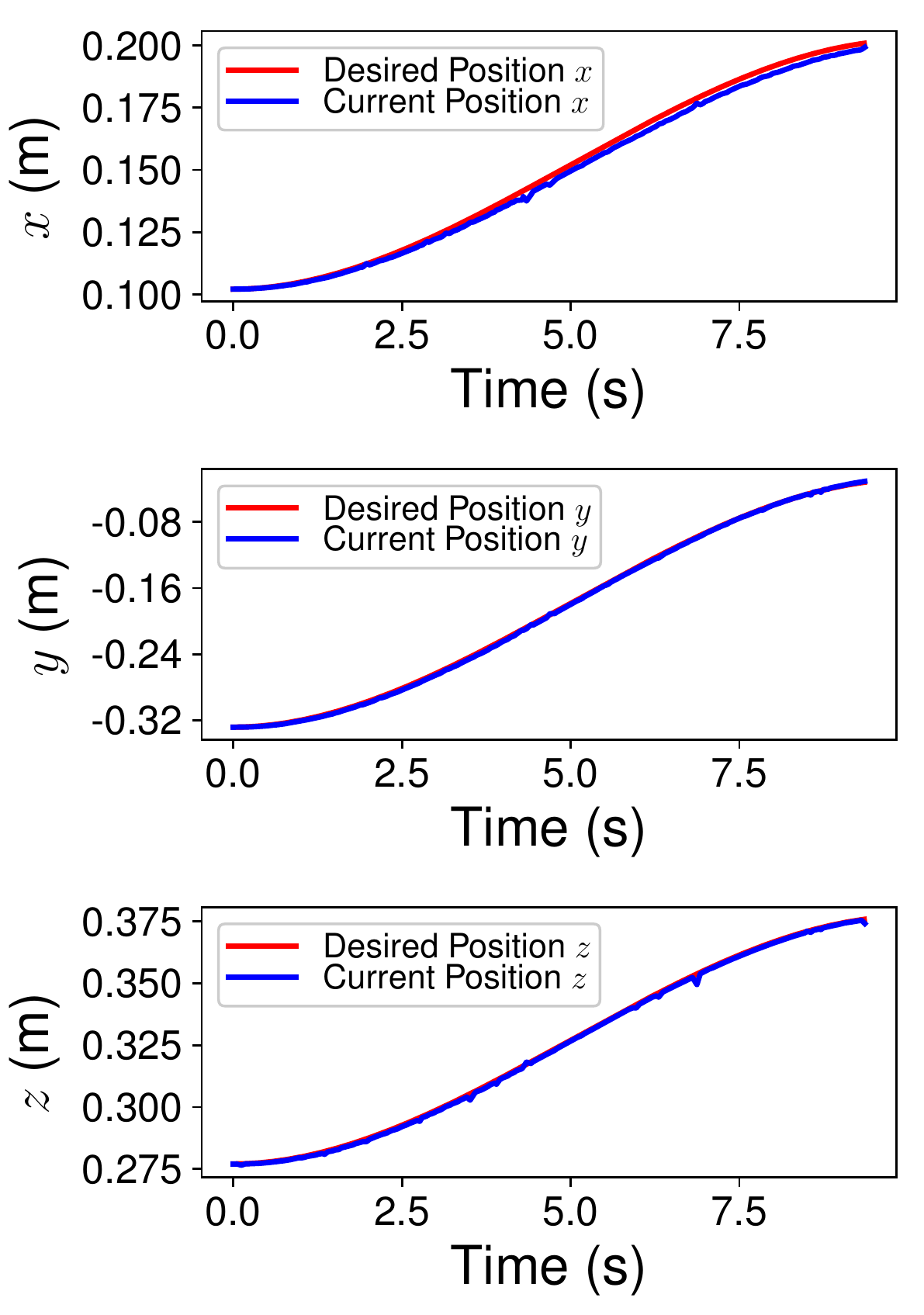}\label{fig:branch-traj-1}}
  \hfil
  \subfloat[]{\includegraphics[width=0.35\textwidth]{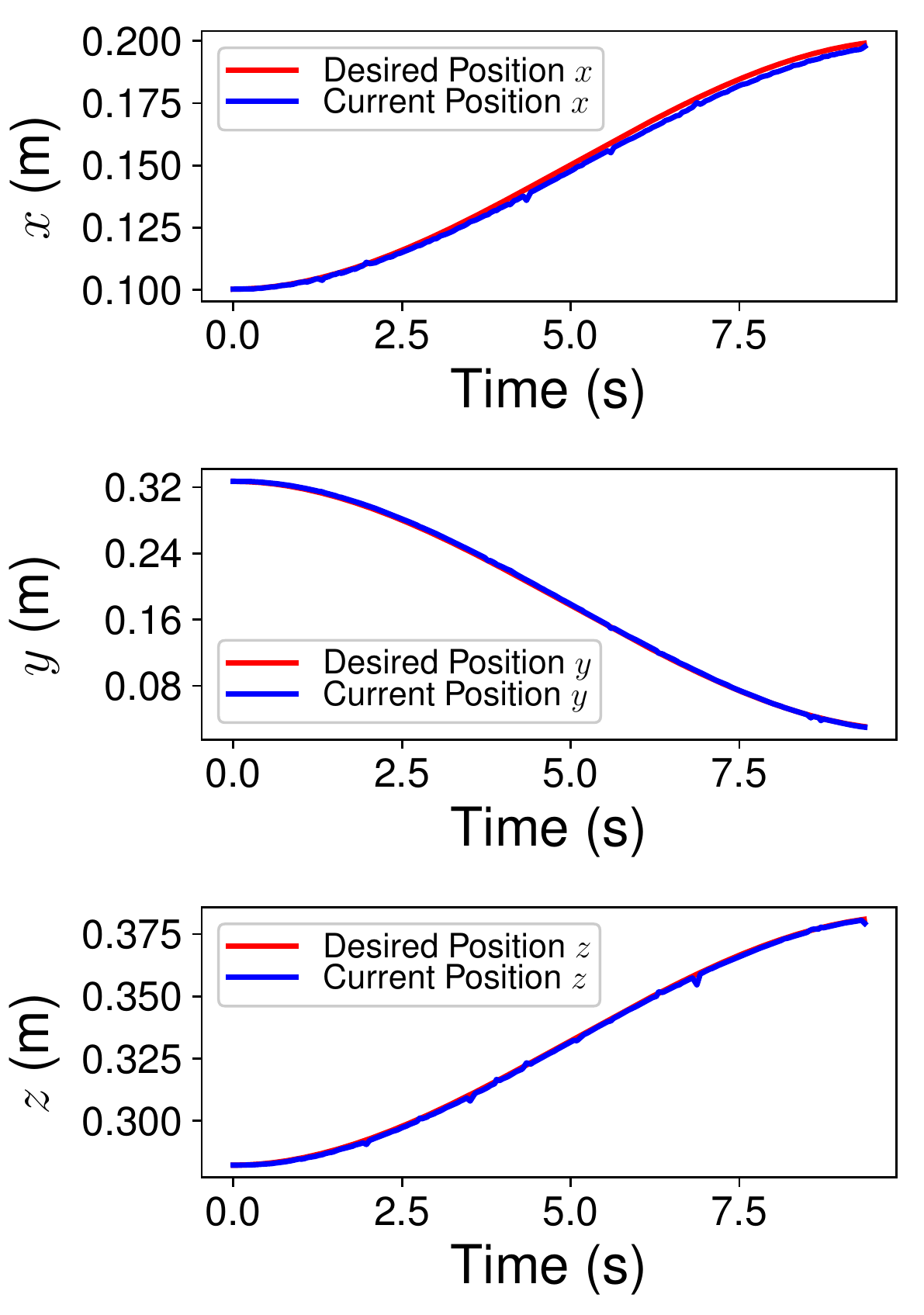}\label{fig:branch-traj-2}}
  \caption{(a) The tracking result for $p_{\mathcal{F}_1}$. (b) The
    tracking result for $p_{\mathcal{F}_2}$.}
\end{figure}

\subsubsection{CKBot Branch}
\label{sec:ckbot-chains}

A configuration with nine CKBot UBar modules is shown in
Fig.~\ref{fig:branch-init}. The base module $\bar{m} = m_1$ is fixed
to the world frame $\mathcal{W}$. Frame $\mathcal{F}_1$ is attached to
connector $\mathcal{T}$ of module $m_6$ and frame $\mathcal{F}_2$ is
attached to connector $\mathcal{T}$ of module $m_9$. Chain
$G_K:\mathcal{W}\rightsquigarrow \mathcal{F}_1$ and
$G_K:\mathcal{W}\rightsquigarrow \mathcal{F}_2$ have common parts
composed by module $m_1$, $m_2$, and $m_3$. The task is to control
$p_{\mathcal{F}_1}$ and $p_{\mathcal{F}_2}$ to follow trajectories
respectively to the pose shown in Fig.~\ref{fig:branch-goal}. The
control loop runs at \SI{20}{Hz} and the gain is
$\mathrm{diag}(0.1, 0.1, 0.1)$ for both motion goals. The tracking
performance is shown in Fig.~\ref{fig:branch-traj-1} and
Fig.~\ref{fig:branch-traj-2}.

\begin{figure}[t!]
  \centering
  \subfloat[]{\includegraphics[width=0.3\textwidth]{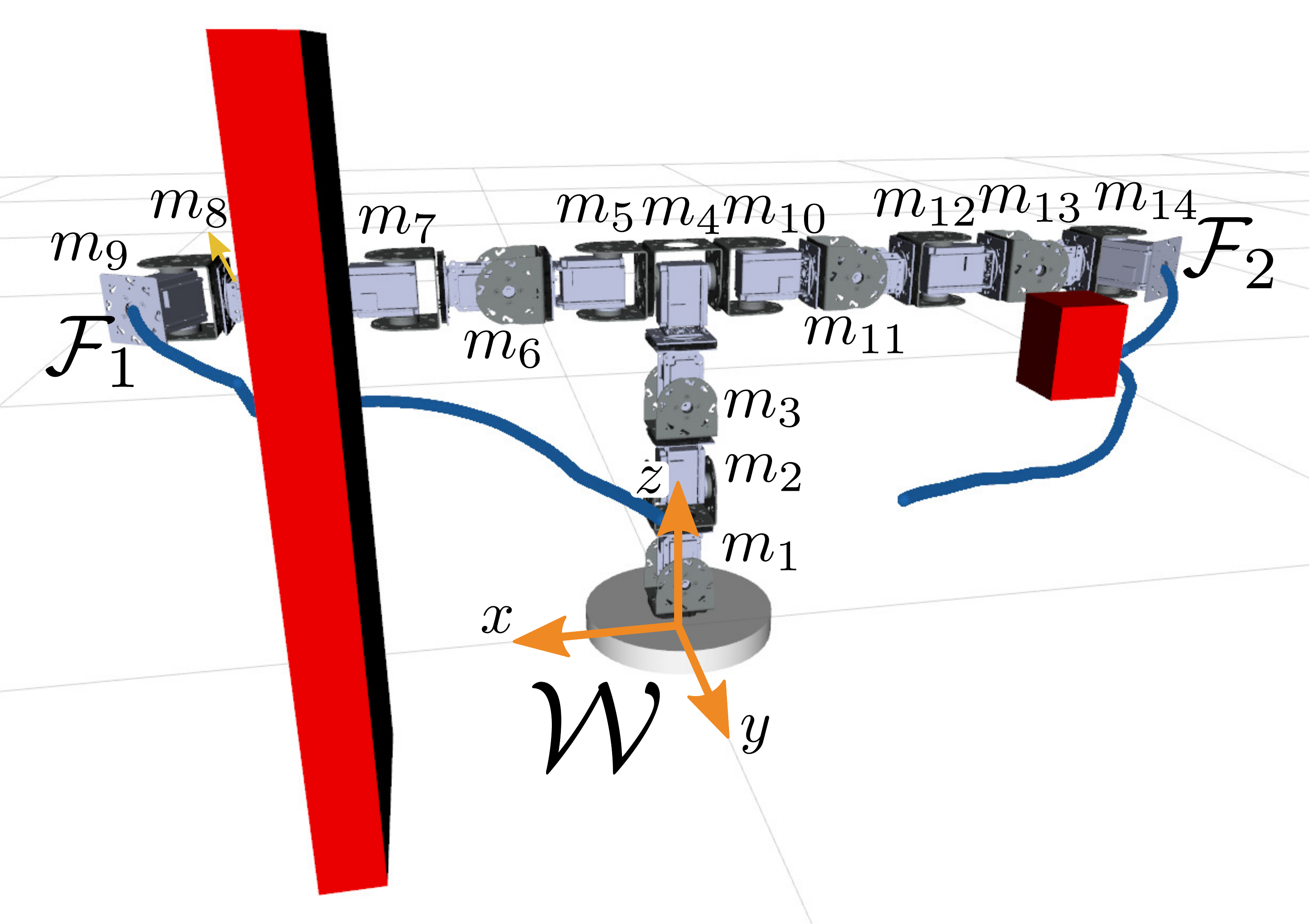}\label{fig:branch-obs-init}}
  \hfil
  \subfloat[]{\includegraphics[width=0.3\textwidth]{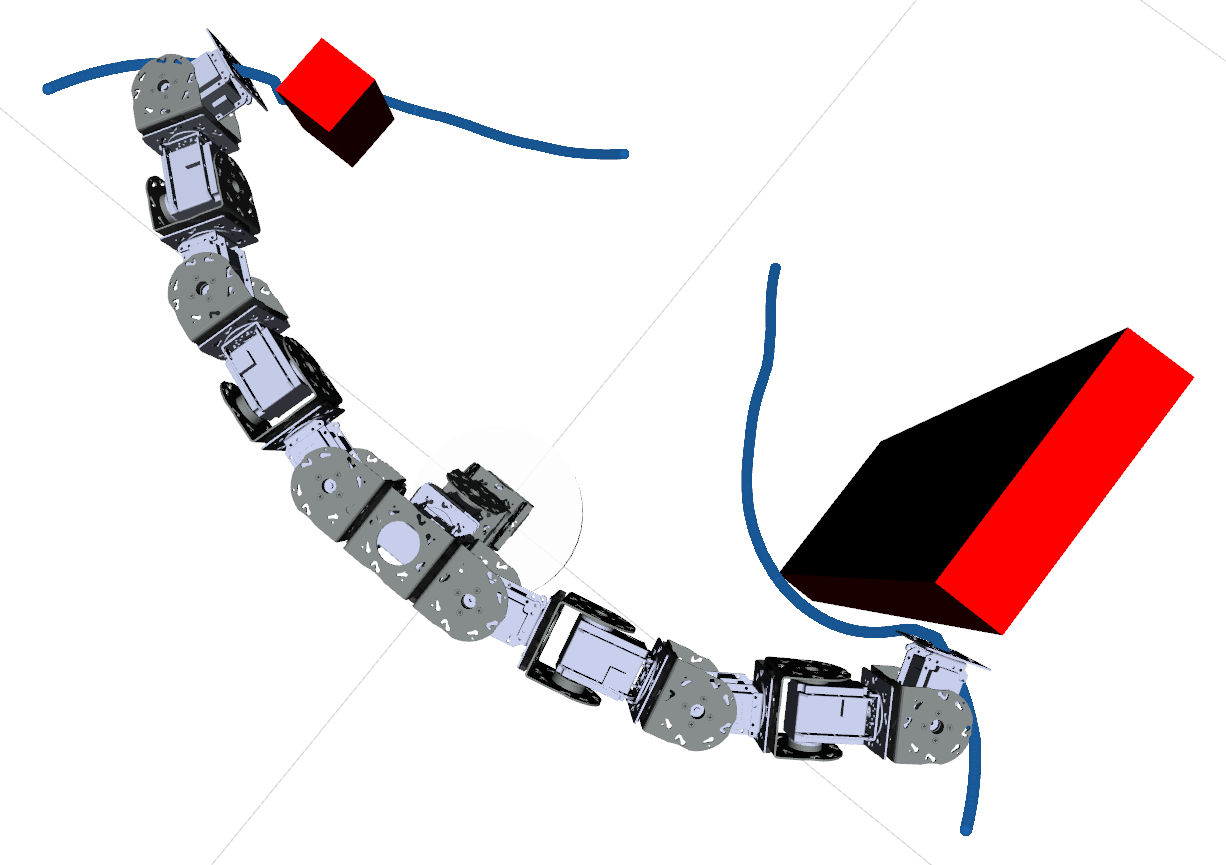}\label{fig:branch-obs-2}}
  \hfil
  \subfloat[]{\includegraphics[width=0.3\textwidth]{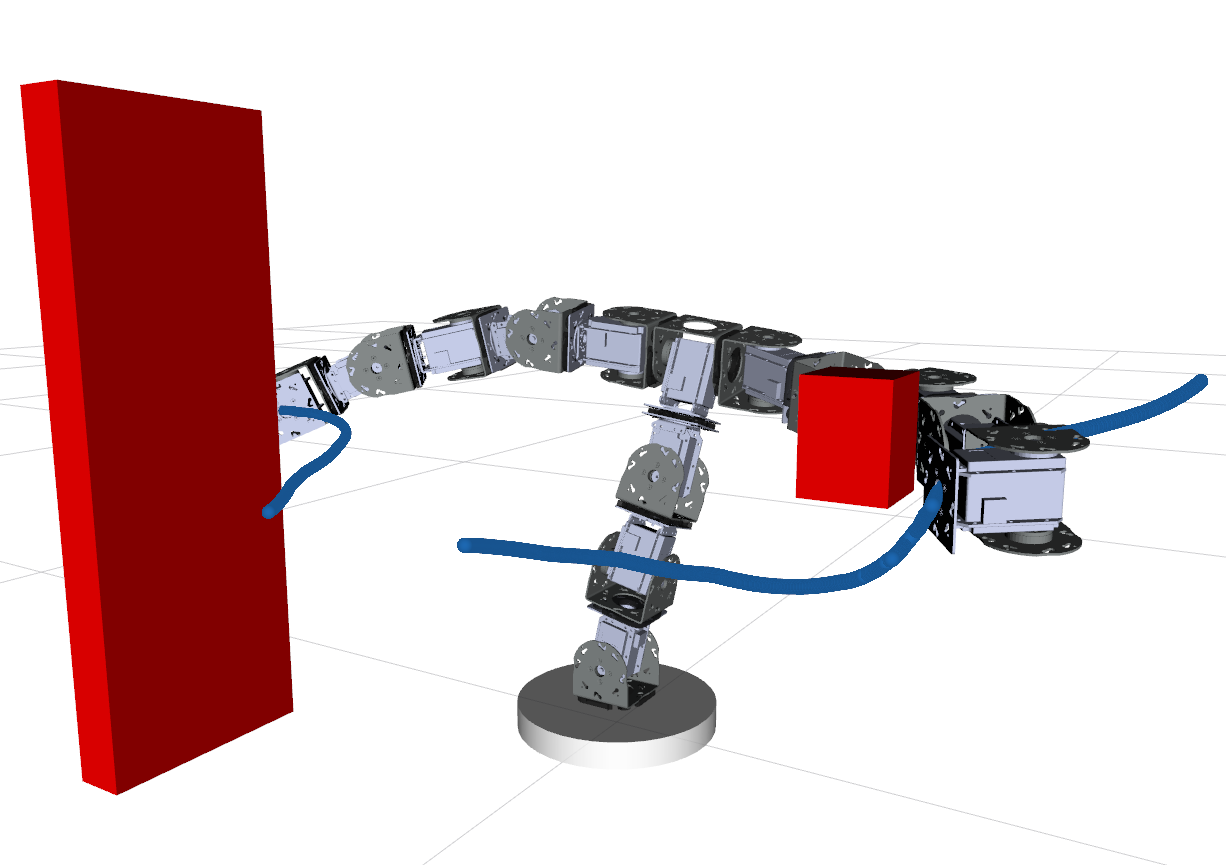}\label{fig:branch-obs-3}}
  \hfil
  \subfloat[]{\includegraphics[width=0.3\textwidth]{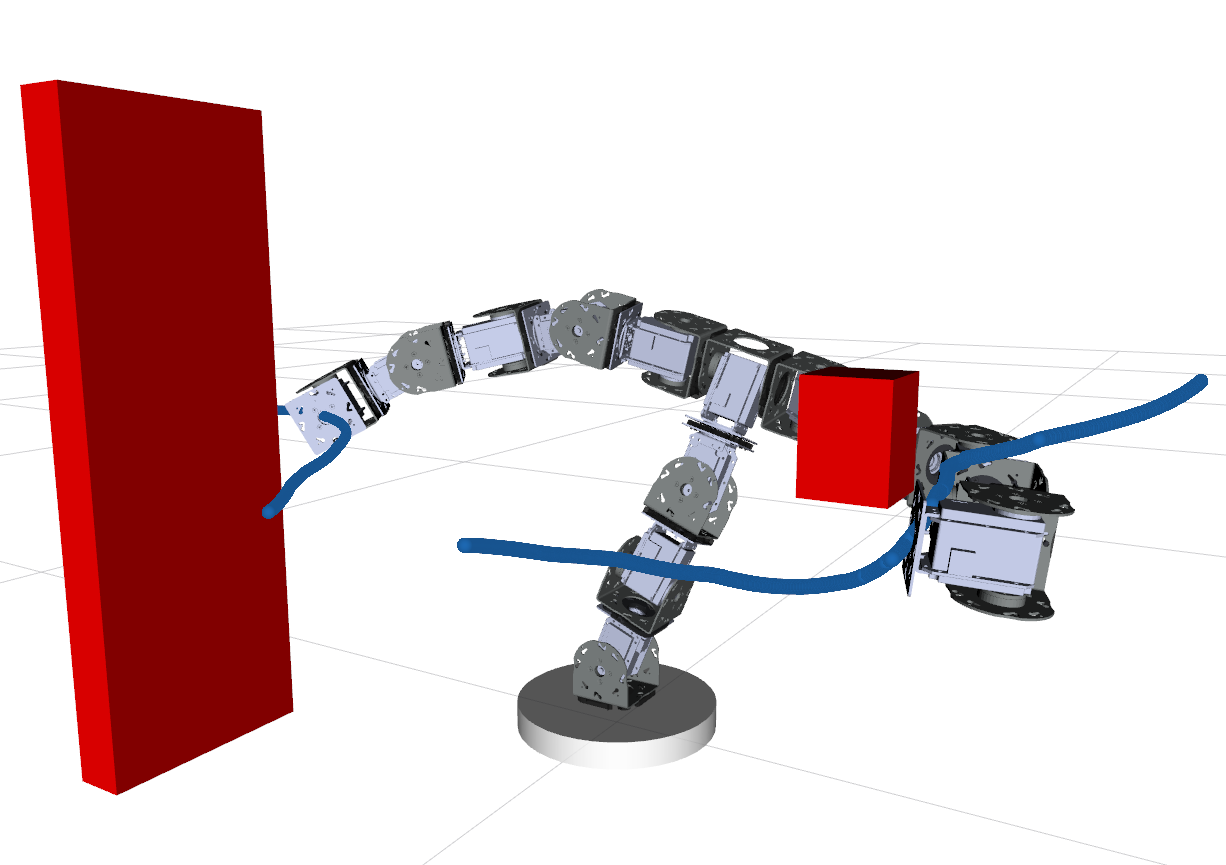}\label{fig:branch-obs-4}}
  \hfil
  \subfloat[]{\includegraphics[width=0.3\textwidth]{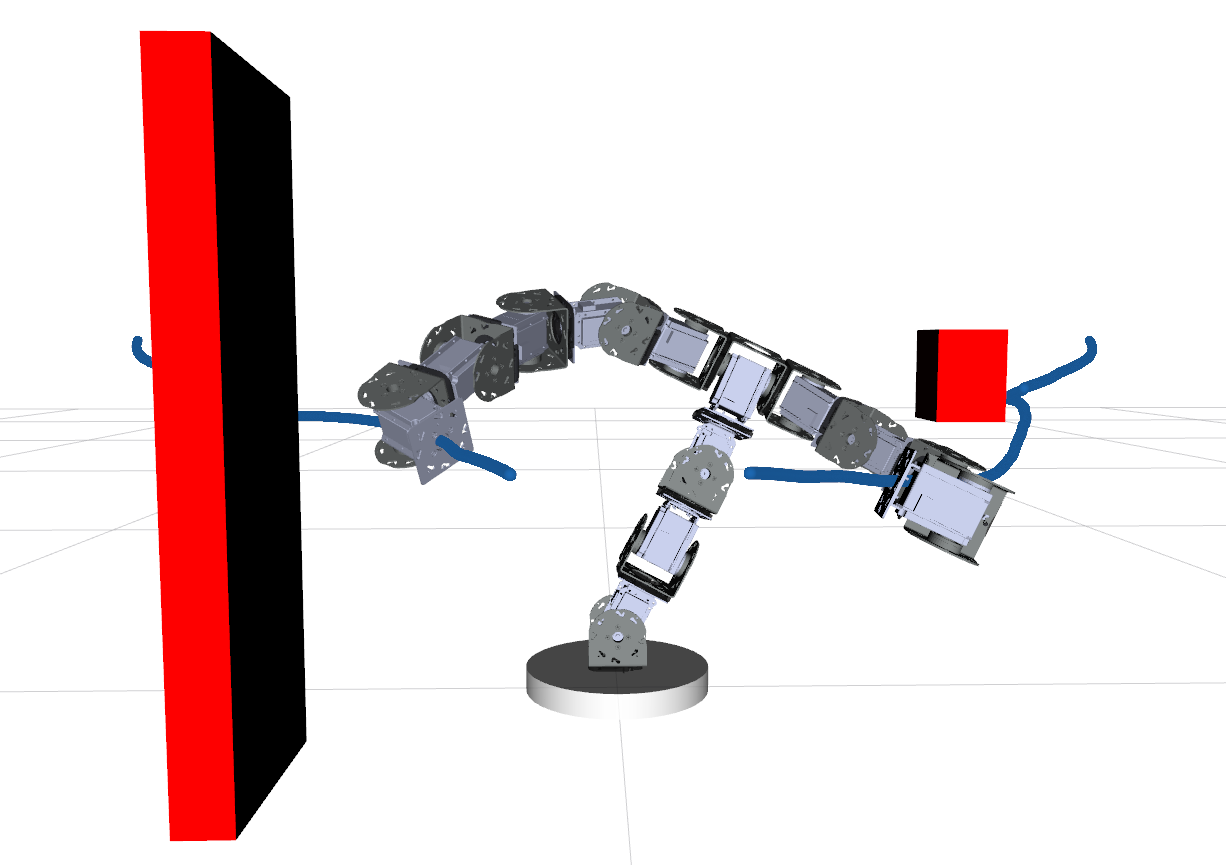}\label{fig:branch-obs-5}}
  \hfil
  \subfloat[]{\includegraphics[width=0.3\textwidth]{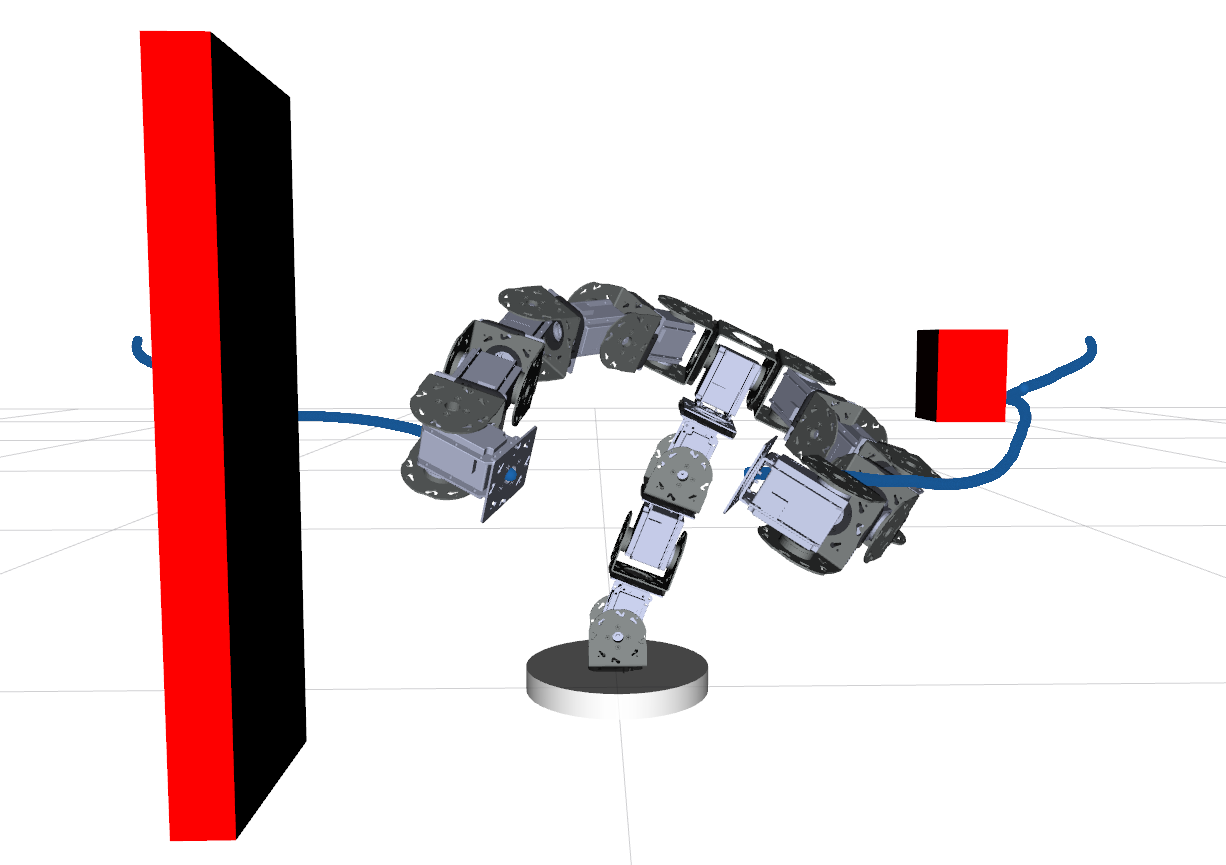}\label{fig:branch-obs-goal}}
  \caption{Control $p_{\mathcal{F}_1}$ and $p_{\mathcal{F}_2}$ from
    the initial pose (a) to new locations between the obstacles
    (f). The body composed by module $m_1$, $m_2$, and $m_3$ first
    moves backward a little bit (b) and then moves to one side in
    order to help $\mathcal{F}_1$ and $\mathcal{F}_2$ to go around
    obstacles (c) --- (e). After going around obstacles, both frames can
    navigate quickly to their destinations. The planned trajectories
    are shown as blue lines.}
  \label{fig:whole-body-manipulation}
\end{figure}

\begin{figure}[t!]
  \centering
  \subfloat[]{\includegraphics[width=0.25\textwidth]{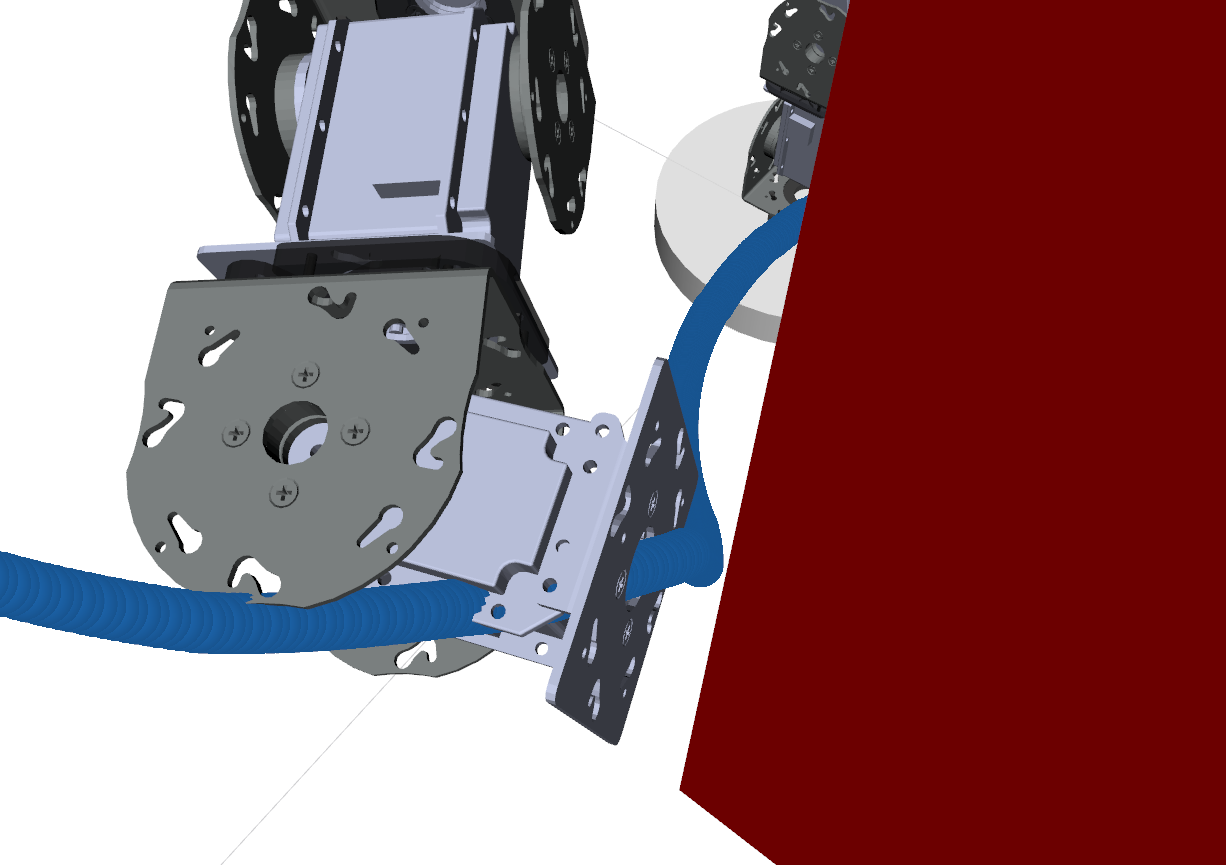}\label{fig:branch-obs-close-1}}
  \hfil
  \subfloat[]{\includegraphics[width=0.25\textwidth]{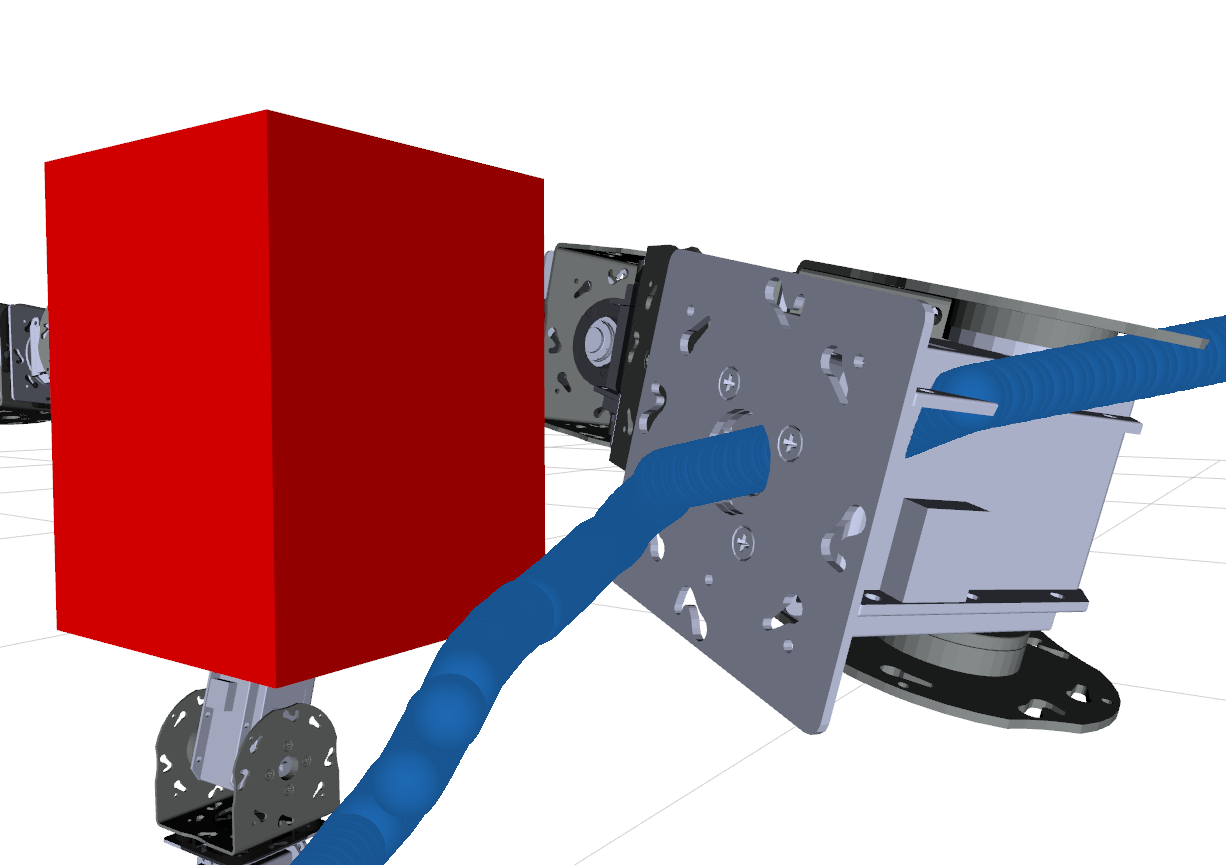}\label{fig:branch-obs-close-2}}
  \caption{(a) Module $m_9$ approaches an obstacle. (b) Module
    $m_{14}$ approaches an obstacle.}
  \label{fig:close-view}
\end{figure}

\subsection{Whole-Body Manipulation}
\label{sec:whole-body}

A configuration with fourteen CKBot UBar modules is constructed in a
simulation environment shown in Fig.~\ref{fig:branch-obs-init}. The
base module $\bar{m} = m_1$ is fixed to the world frame
$\mathcal{W}$. There are two obstacles in the workspace which are
close to the robot. The sphere-tree construction outputs 126 obstacle
spheres in total to approximate these two obstacles. Frame
$\mathcal{F}_1$ and $\mathcal{F}_2$ are attached to connector
$\mathcal{T}$ of module $m_9$ and module $m_{14}$
respectively. Similarly, Chain
$G_K:\mathcal{W}\rightsquigarrow \mathcal{F}_1$ and
$G_K:\mathcal{W}\rightsquigarrow \mathcal{F}_2$ share four
modules. The task is to control $p_{\mathcal{F}_1}$ and
$p_{\mathcal{F}_2}$ to new locations between these two obstacles. The
control loop runs at \SI{20}{Hz} and the gain is
$\mathrm{diag}(0.1, 0.1, 0.1)$ for both motion goals. In this complex
scenario, the quadratic program can be constructed and solved by
Gurobi~\cite{gurobi} in \SI{6.3}{ms} in average with standard
deviation of \SI{2.3}{ms} and a maximum time of \SI{15.5}{ms} on a
laptop computer (Intel Core i7-8750H CPU, 16GB RAM).

\begin{figure}[t!]
  \centering
  \subfloat[]{\includegraphics[width=0.35\textwidth]{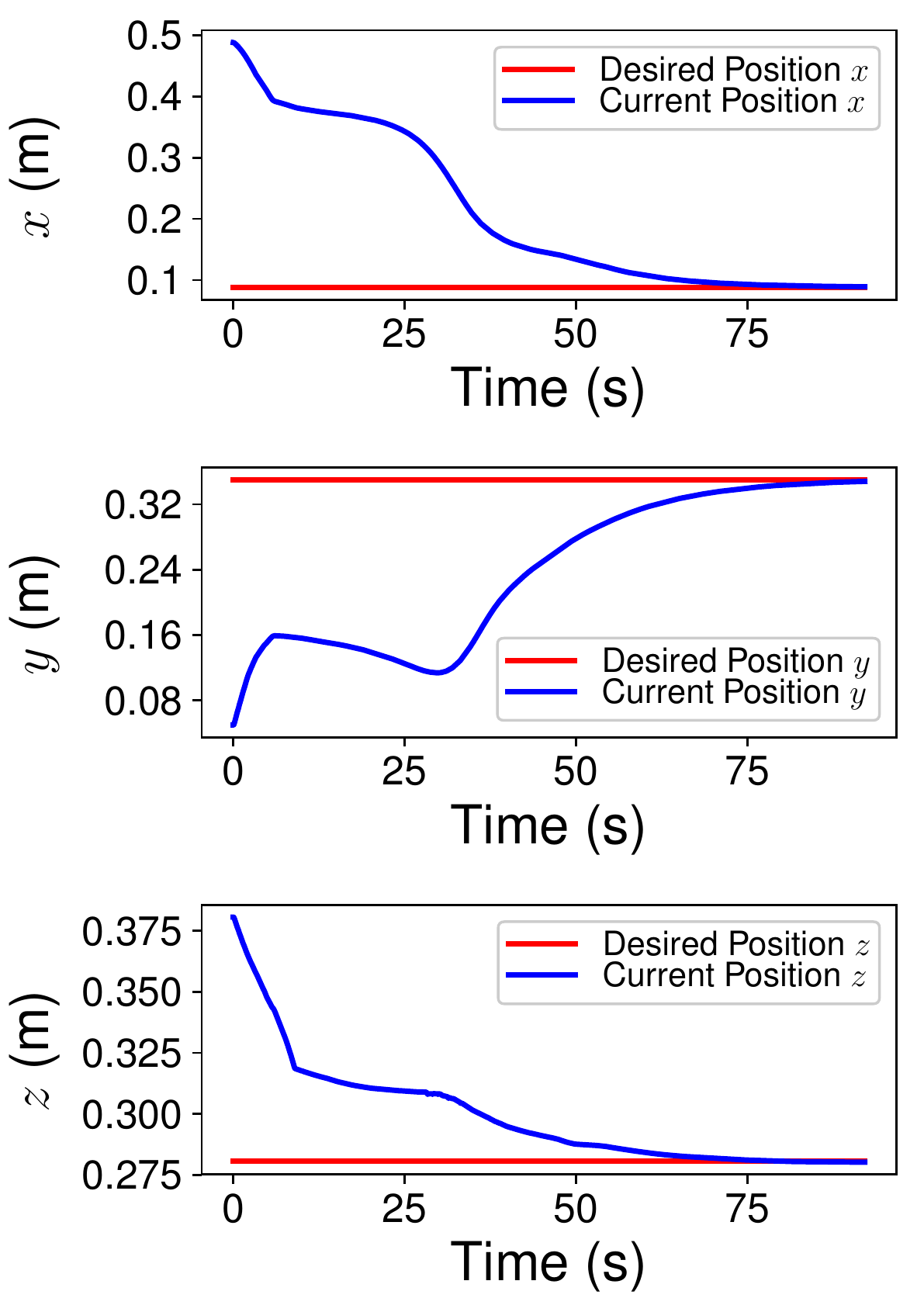}\label{fig:branch-navigation-1}}
  \hfil
  \subfloat[]{\includegraphics[width=0.35\textwidth]{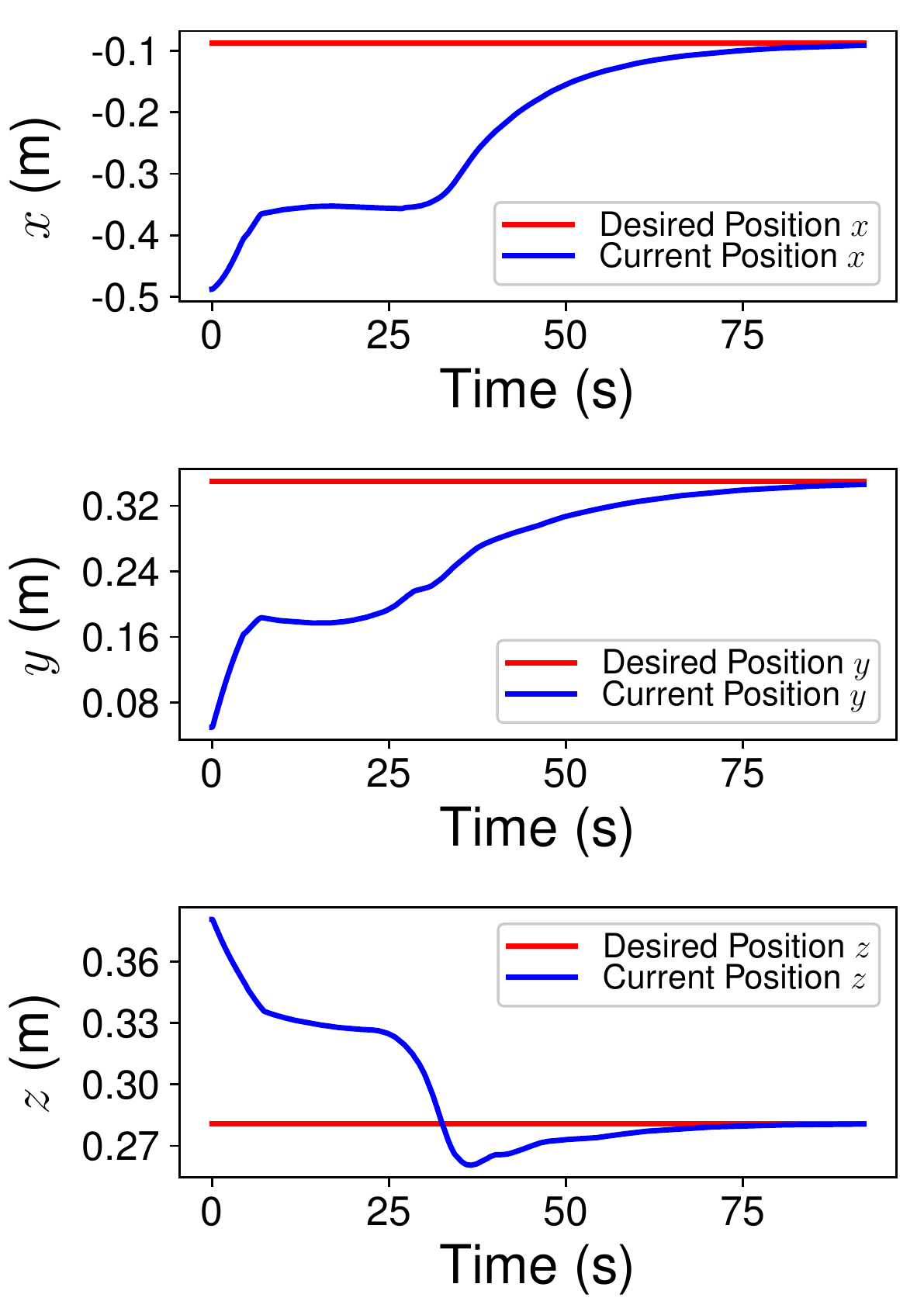}\label{fig:branch-navigation-2}}
  \caption{(a) The motion of $p_{\mathcal{F}_1}$. (b) The motion of
    $p_{\mathcal{F}_2}$.}
  \label{fig:branch-navigation-pos}
\end{figure}

Initially the motion of $p_{\mathcal{F}_1}$ and $p_{\mathcal{F}_2}$
are symmetric because modules are all not very close to obstacles. At
this stage, the main body composed by module $m_1$, $m_2$, and $m_3$
moves backward slightly. Frame $\mathcal{F}_1$ approaches one of the
obstacle first (Fig.~\ref{fig:branch-obs-2} and
Fig.~\ref{fig:branch-obs-close-1}), and the objective function is
updated to penalize the motion of $m_9$ which has to move along the
obstacle. Then module $m_{14}$ approaches the other obstacle
(Fig.~\ref{fig:branch-obs-3} and Fig.~\ref{fig:branch-obs-close-2}),
and a penalty term for this module is also added to the objective
function. Both frames move slowly during this phase
(Fig.~\ref{fig:branch-obs-3} and Fig.~\ref{fig:branch-obs-4}).  We can
see from Fig.~\ref{fig:branch-navigation-pos} that $p_{\mathcal{F}_1}$
and $p_{\mathcal{F}_2}$ change slowly. Repulsive motion constraints
are added to the optimization function when some module nearly contact
obstacles. The main body leans to one side to help both frames to go
around obstacles. After moving around obstacles slowly, both frames
can quickly navigate to their destinations
(Fig.~\ref{fig:branch-obs-5} and Fig.~\ref{fig:branch-obs-goal}). The
final planned trajectory takes \SI{92.15}{s}.

\section{Conclusion}
\label{sec:conclusion}

We present a new approach to online manipulation motion planning well
suited for reconfigurable modular robot systems. This approach
formulates the motion planning problem as a sequential quadratic
program. We propose a novel way to approximate obstacles in the
environment considering both accuracy and simplicity so that the
obstacle-avoidance requirement can be modeled as a small number of
linear constraints. The objective function and constraints are updated
according to the current scenario in order to handle a larger range of
tasks. All motion constraints are linear that allows it to be applied
to real-time control. Multiple strongly coupled motion tasks can be
handled easily which is particularly useful for modular robots.




\bibliographystyle{splncs04}
\bibliography{biblio}

\end{document}

%% file: preamble.tex
\usepackage[T1]{fontenc}
\usepackage{amsmath}
\usepackage{amsfonts}
\usepackage{amssymb}
\usepackage{hyperref}
\usepackage{graphicx}
\usepackage{epstopdf}
\AppendGraphicsExtensions{.gif}
\graphicspath{{figures/pdf/}{figures/png/}{figures/jpg/}{figures/eps/}{figures/gif/}}
\DeclareGraphicsExtensions{.pdf,.png,.jpg,.eps,.gif}
\usepackage[font=footnotesize]{subfig}
\usepackage{multirow}
\usepackage{romannum}
\usepackage{booktabs} 
\usepackage[all]{nowidow}
\usepackage[utf8]{inputenc}
\usepackage{gensymb}
\usepackage{float}

\usepackage{algorithmic}
\usepackage[ruled,norelsize,linesnumbered]{algorithm2e}
\usepackage{siunitx}
\usepackage{bm}
\usepackage{hyperref}
\usepackage[strings]{underscore}

\usepackage[mathlines]{lineno}
\usepackage{etoolbox} 
\newcommand*\linenomathpatchAMS[1]{%
  \expandafter\pretocmd\csname #1\endcsname {\linenomathAMS}{}{}%
  \expandafter\pretocmd\csname #1*\endcsname{\linenomathAMS}{}{}%
  \expandafter\apptocmd\csname end#1\endcsname {\endlinenomath}{}{}%
  \expandafter\apptocmd\csname end#1*\endcsname{\endlinenomath}{}{}%
}

\expandafter\ifx\linenomath\linenomathWithnumbers
  \let\linenomathAMS\linenomathWithnumbers
  
  \patchcmd\linenomathAMS{\advance\postdisplaypenalty\linenopenalty}{}{}{}
\else
  \let\linenomathAMS\linenomathNonumbers
\fi

\linenomathpatchAMS{gather}
\linenomathpatchAMS{multline}
\linenomathpatchAMS{align}
\linenomathpatchAMS{alignat}
\linenomathpatchAMS{flalign}

\makeatletter
\renewcommand\section{\@startsection{section}{1}{\z@}%
                       {-8\p@ \@plus -4\p@ \@minus -4\p@}%
                       {6\p@ \@plus 4\p@ \@minus 4\p@}%
                       {\normalfont\large\bfseries\boldmath
                        \rightskip=\z@ \@plus 8em\pretolerance=10000 }}
\renewcommand\subsection{\@startsection{subsection}{2}{\z@}%
                       {-8\p@ \@plus -4\p@ \@minus -4\p@}%
                       {6\p@ \@plus 4\p@ \@minus 4\p@}%
                       {\normalfont\normalsize\bfseries\boldmath
                        \rightskip=\z@ \@plus 8em\pretolerance=10000 }}
\renewcommand\subsubsection{\@startsection{subsubsection}{3}{\z@}%
                       {-4\p@ \@plus -4\p@ \@minus -4\p@}%
                       {-1.5em \@plus -0.22em \@minus -0.1em}%
                       {\normalfont\normalsize\bfseries\boldmath}}
\makeatother